\documentclass[11pt]{article}

\usepackage[final]{acl}

\usepackage{fontspec}
\usepackage[bidi=default]{babel}
\babelprovide[main, import]{english}
\babelprovide[import]{urdu}

\usepackage{times}
\usepackage{latexsym}
\usepackage{graphicx}
\usepackage{multirow}
\usepackage{subcaption}
\usepackage[T1]{fontenc}
\usepackage[utf8]{inputenc}
\usepackage{microtype}
\usepackage{inconsolata}
\usepackage[most]{tcolorbox}
\usepackage{xcolor}
\usepackage{hyperref}
\usepackage{fontawesome5}
\usepackage{caption}
\usepackage{colortbl}
\usepackage{rotating}
\definecolor{barblue}{HTML}{4C78A8}
\definecolor{bargray}{HTML}{E6E6E6}
\definecolor{softgray}{HTML}{EBE0D2}

\usepackage{booktabs}
\usepackage{tabularx}
\usepackage{array}
\usepackage[colorinlistoftodos]{todonotes}
\usepackage{enumitem}
\usepackage{xcolor}
\definecolor{darkgreen}{rgb}{0.0,0.55,0.0}
\definecolor{darkred}{rgb}{0.75,0.0,0.0}

\babelfont[urdu]{rm}[
  Path=fonts/,
  UprightFont = *,
  Script=Arabic,
  Language=Urdu,
  Scale=0.85
]{NotoNastaliqUrdu-Regular.ttf}

\newlength{\distbarwidth}
\newcommand{\distbar}[1]{%
  \begingroup
  \setlength{\fboxsep}{0pt}%
  \colorbox{bargray}{%
    \makebox[\distbarwidth][l]{%
      \colorbox{barblue}{\rule{#1\dimexpr\distbarwidth/100\relax}{0.9ex}}%
    }%
  }%
  \endgroup
}

\newcommand{\distrowtwo}[4]{%
  #1 & #2 & #3 & #4\% & \distbar{#4} \\
}

\newcounter{invex}[section]
\renewcommand{\theinvex}{\thesection.\arabic{invex}}

\usepackage{xspace}

\newcommand{\dataset}[1]{\textsc{#1}\xspace}

\newcommand{\ds}{\dataset{UrduMMLU}}

\title{\ds{}: A Massive Multitask Benchmark\\ for Urdu Language Understanding}

\author{
\textbf{Ahmer Tabassum}\thanks{Equal contribution.}\textsuperscript{1} \
\textbf{Sarfraz Ahmad}$^*$\textsuperscript{1} \
\textbf{Hasan Iqbal}$^*$\textsuperscript{1} \\
\textbf{Owais Aijaz}\textsuperscript{1} \
\textbf{Momina Ahsan}\textsuperscript{1}
\textbf{Preslav Nakov}\textsuperscript{1} \\
\textsuperscript{1}Mohamed bin Zayed University of Artificial Intelligence (MBZUAI) \\
\parbox{\linewidth}{\centering
\texttt{\{ahmer.tabassum, sarfraz.ahmad, hasan.iqbal\}@mbzuai.ac.ae} \\
[0.3em]
\faGlobe\ \href{https://mbzuai-nlp.github.io/UrduMMLU/}{Project}
\quad
\faDatabase\ \href{https://huggingface.co/datasets/MBZUAI/UrduMMLU}{\ds{}}
\quad
\faGithub\ \href{https://github.com/mbzuai-nlp/urdu-mmlu}{Code}
\quad
\faTrophy\ \href{https://mbzuai-nlp.github.io/UrduMMLU/leaderboard.html}{Leaderboard}
}
}

\begin{document}
\maketitle

\begin{abstract}
Meaningful multilingual evaluation must test models in the target language and educational context. Urdu, spoken by more than 230 million people, lacks a broad MMLU-style benchmark built from native educational sources. We introduce \ds{}, a benchmark of 26{,}389 Urdu MCQs across 26 subjects and five domains, collected from native Urdu MCQ banks and public examination PDFs. Unlike translation-based benchmarks, \ds{} combines academic subjects with content specific to Urdu and regional education. We label the exam-derived portion through dual human annotation with strict consensus filtering. We evaluate 30 LLMs under English and Urdu prompts, yielding 60 zero-shot evaluations, and further evaluate four open-source LLMs under multiple few-shot settings across both prompt languages. Gemini-3.5-Flash performs best, reaching 90.23\% and 90.45\% accuracy, while no other model exceeds 85\%. The strongest open-source model trails by 7.78 and 9.12 points, and many models lose 25 to 40 points on Urdu-centered Humanities subjects compared with STEM. Few-shot prompting yields only modest gains. Results on \ds{} show that current LLMs have uneven Urdu knowledge, particularly for content grounded in the regional context.
\end{abstract}
\section{Introduction}
\label{sec:introduction}

Evaluating the knowledge and reasoning abilities of Large Language Models (LLMs) has become central to Natural Language Processing (NLP). Benchmarks such as \dataset{MMLU}~\citep{hendrycks2021measuring} and \dataset{MMLU-Pro}~\citep{wang2024mmlupro} are widely used for this purpose, but they are in English and largely reflect English-language educational and cultural contexts. This limits their ability to test whether model competence transfers across language, script, and regional knowledge. As a result, these benchmarks provide only a partial view of model performance in multilingual and culturally diverse settings.

\begin{figure*}[t]
    \centering
    \includegraphics[
        width=0.93\linewidth,
        trim={2cm 3cm 3cm 3.5cm},
        clip]{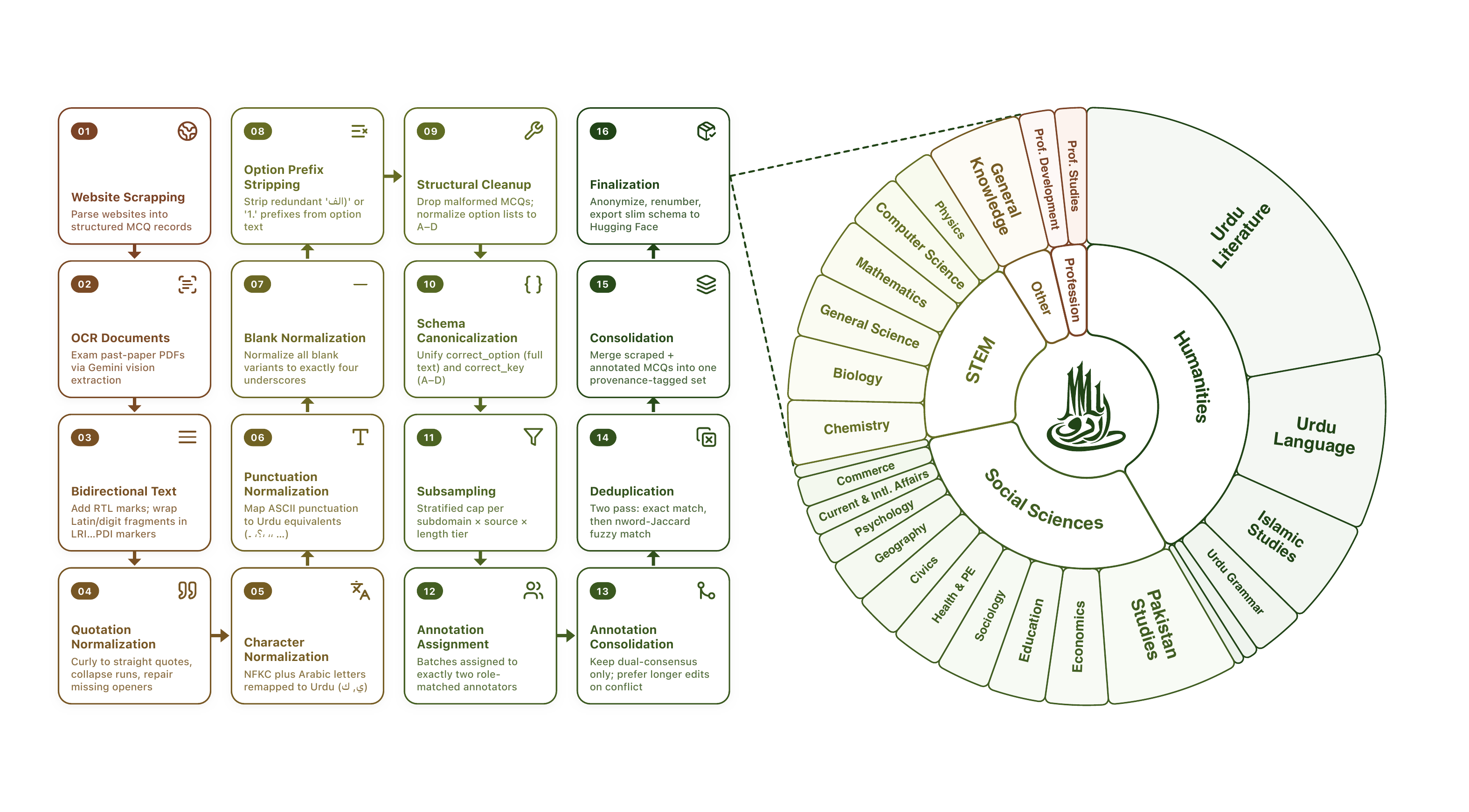}
    \caption{\textbf{UrduMMLU construction and composition.} The 16-stage pipeline (left) produces a 26,389-MCQ benchmark across 5 domains and 26 subdomains (right), with wedge size proportional to MCQ count.}
    \label{fig:data_collection_pipeline}
\end{figure*}

This gap is especially important for Urdu, which is spoken by over 230 million people and has a long literary and educational tradition but few broad-coverage evaluation resources. Existing benchmarks focus mainly on reading comprehension, syntactic diagnostics, task-level NLP, or translated reasoning \citep{urduquad,kazi-etal-2025-crossing,urblimp,tahir-etal-2025-benchmarking,urdubench, 10.1145/3748308, 10.1145/3622939}. Multilingual benchmarks covering Urdu, including \dataset{MMLU-ProX}~\citep{xuan-etal-2025-mmlu}, \dataset{Global-MMLU}~\citep{singh-etal-2025-global}, and \dataset{IndicMMLU-Pro}~\citep{kj2025indicmmlu}, also rely largely on translated questions. They capture only part of the knowledge in Urdu-medium education, literature, local history, religious studies, and civic curricula.

Language-specific benchmarks such as \dataset{ArabicMMLU}~\citep{koto-etal-2024-arabicmmlu}, \dataset{CMMLU}~\citep{li-etal-2024-cmmlu}, \dataset{IndoMMLU}~\citep{koto-etal-2023-large}, \dataset{KMMLU}~\citep{son-etal-2025-kmmlu}, and \dataset{KazMMLU}~\citep{togmanov-etal-2025-kazmmlu} show the value of evaluation grounded in local educational material. We introduce \ds{}, which to our knowledge is the first broad-coverage MMLU-style benchmark for Urdu, built primarily from questions originally written in Urdu. It contains 26{,}389 MCQs across 26 subjects and five domains, collected from Urdu MCQ banks and public SSC/HSSC examination PDFs. It combines answer-labeled items and exam-derived questions retained through dual annotation and strict consensus filtering, covering academic subjects and Urdu- and region-specific content. Figure~\ref{fig:data_collection_pipeline} summarizes the subject distribution.

We run 60 zero-shot evaluations of 30 open-weight and proprietary LLMs under English and Urdu prompts, plus one-, three-, and five-shot evaluations of four open-weight models. Accuracy covers all items, with unparsable outputs counted as incorrect. Gemini-3.5-Flash~\citep{google2026gemini35flash} leads with 90.23\% and 90.45\% accuracy, while the strongest open-weight model trails by 7.78 and 9.12 points.

Across models, performance remains substantially higher on STEM subjects than on Urdu-centered Humanities, with many systems losing 25 to 40 points on Urdu literature, Urdu language, and Islamic studies. Within \ds{}, then, high-capacity models remain uneven across domains and prompt languages, and that unevenness concentrates in Urdu-specific content. These results highlight the need for benchmarks that better capture linguistic and cultural diversity beyond English.

The main contributions of this work are:
\begin{itemize}[itemsep=1pt, topsep=1pt]
\item We introduce \ds{}, a natively written Urdu MMLU-style benchmark with 26{,}389 MCQs across 26 subjects and five domains, covering both standard academic subjects along with Urdu- and region-specific knowledge.
\item We produce human-annotated gold answers for the exam-derived portion of the benchmark using dual annotation and strict consensus filtering.
\item We conduct 60 zero-shot evaluations across 30 open-weight and proprietary LLMs under English and Urdu prompt settings, and 24 additional few-shot evaluations across four open-weight LLMs.
\item We release the dataset and evaluation code to support future work on Urdu-capable language models.
\end{itemize}

\section{Related Work}
\label{sec:background}

\paragraph{Urdu evaluation resources:} Existing Urdu resources cover reading comprehension, cross-lingual question answering , syntax, and task-level NLP. \dataset{UQuAD+}~\citep{urduquad} provides annotated Urdu reading comprehension, while~\citet{kazi-etal-2025-crossing} study Urdu-English QA with \dataset{UQuAD1.0}~\citep{kazi2021uquad} and \dataset{SQuAD2.0}~\citep{rajpurkar-etal-2018-know}. \dataset{UrBLiMP}~\citep{urblimp} evaluates Urdu syntax via minimal pairs, and \citet{tahir-etal-2025-benchmarking} benchmark models across Urdu NLP tasks. For reasoning, \dataset{UrduBench}~\citep{urdubench} translates \dataset{Mgsm}~\citep{shi2023language}, \dataset{CommonsenseQA}~\citep{talmor-etal-2019-commonsenseqa}, \dataset{OpenBookQA}~\citep{mihaylov-etal-2018-suit}, and \dataset{Math-500}~\citep{lightman2024lets} into Urdu, and \dataset{UrduFactCheck}~\citep{ahmad-etal-2025-urdufactcheck} targets factual QA. These resources remain task-specific, diagnostic, or translation-derived. In contrast, \ds{} evaluates broad educational knowledge using questions originally written for Urdu-speaking educational settings.

\paragraph{Multilingual benchmarks:} \dataset{MMLU}~\citep{hendrycks2021measuring} and \dataset{MMLU-Pro}~\citep{wang2024mmlupro} are widely used for evaluating general knowledge and reasoning. Several multilingual extensions adapt these benchmarks through translation. \dataset{MMLU-ProX}~\citep{xuan-etal-2025-mmlu} extends \dataset{MMLU-Pro} to 29 languages using LLM-based translation and expert review, while \dataset{Global-MMLU}~\citep{singh-etal-2025-global} studies cultural and linguistic bias in multilingual evaluation. 

\dataset{IndicMMLU-Pro}~\citep{kj2025indicmmlu} adapts \dataset{MMLU-Pro} to nine Indic languages, including Urdu. Other multilingual exam-based resources, such as \dataset{Exams}~\citep{hardalov-etal-2020-exams}, \dataset{Include}~\citep{romanou2025include}, and \dataset{Milu}~\citep{verma-etal-2025-milu}, collect examination questions across multiple languages and regions. However, Urdu still appears primarily in translated or cross-lingual settings rather than through a dedicated native benchmark, limiting fair knowledge assessment in cultural context. 

\paragraph{Localized MMLU-style benchmarks:} Recent work increasingly builds \dataset{MMLU}-style benchmarks from local educational material instead of translating English benchmarks. \dataset{ArabicMMLU}~\citep{koto-etal-2024-arabicmmlu}, \dataset{CMMLU}~\citep{li-etal-2024-cmmlu}, \dataset{IndoMMLU}~\citep{koto-etal-2023-large}, \dataset{KMMLU}~\citep{son-etal-2025-kmmlu}, and \dataset{KazMMLU}~\citep{togmanov-etal-2025-kazmmlu} show that language-specific curricula and regional cultural knowledge remain important for evaluating LLMs beyond English. \ds{} follows this direction for Urdu by combining regional SSC/HSSC examination material, native Urdu MCQ banks, human annotation for exam-derived questions, and broad coverage of both standard academic subjects and Urdu- and Pakistan-specific knowledge.
\section{\ds{}}
\label{sec:dataset}

\ds{} is a broad-coverage benchmark for evaluating knowledge and reasoning in Urdu. Unlike translation-based multilingual benchmarks, \ds{} draws its questions directly from Urdu educational and examination material. The benchmark contains 26{,}389 MCQs across 26 subdomains and five domains, covering both standard academic subjects and Urdu- and region-specific content such as Urdu literature, Urdu language, Islamic studies, and Pakistan studies. Appendix~\ref{app:source-domain-distributions} and Figure~\ref{fig:subject-bars} provide detailed benchmark statistics and subdomain distributions. We collect questions from Urdu MCQ banks and public SSC/HSSC examination PDFs, and produce gold answers for exam-derived questions through dual human annotation with strict consensus filtering. We design \ds{} around broad subject coverage, faithful representation of Urdu educational material, and reliable multiple-choice evaluation through clean text extraction, normalized metadata, and verified gold labels. Figure~\ref{fig:data_collection_pipeline} summarizes the overall construction pipeline.

\subsection{Data Sources}
\label{sec:source-collection}

We collect candidate questions from two source families. The first consists of public SSC and HSSC examination PDFs from Pakistan covering school- and high school-level subjects such as mathematics, physics, chemistry, biology, computer science, Urdu, Islamic studies, Pakistan studies, and economics. The second consists of native Urdu MCQ websites that publish answer-labeled questions for examination preparation. Together, these sources allow \ds{} to cover both globally shared academic subjects and region-specific educational content taught in Urdu-medium curricula. We treat all collected items as candidates and include them in the final benchmark only after cleaning, answer annotation or verification, deduplication, and release packaging.

\subsection{Raw MCQ Extraction}
\label{sec:raw-extraction}

For PDF-based sources, we use a multi-stage extraction pipeline to recover Urdu MCQs from heterogeneous examination layouts. We first convert each PDF into page images and classify each page with Claude Sonnet 4.6~\citep{anthropic2026sonnet46}, filtering out English-only pages, non-MCQ pages, answer keys, and unrelated material. For the remaining pages, we extract question stems, answer options, source metadata, and page-level provenance with Gemini 3 Flash under a vision-language OCR prompt. We design the extraction prompt specifically for Urdu examination documents. The prompt preserves Urdu question text, answer options, poetry, quotations, and other context required to answer the question correctly. In bilingual material, we ignore English text unless it forms a structural part of the Urdu question, and we discard unreadable questions rather than reconstructing missing content. For web-based sources, we directly scrape question stems, answer options, category labels, and answer keys when available.

\subsection{Metadata and Schema Normalization}
\label{sec:metadata-schema-normalization}

The collected sources use heterogeneous category names, grade labels, and answer formats, so we normalize all examples into a unified representation. We map source-specific labels to a controlled set of subdomains. For example, we map variants such as \texttt{Everyday Science} and \texttt{General Science} to \texttt{general science}, and mathematics-related labels such as \texttt{maths}, \texttt{General Mathematics}, and \texttt{riazi} to \texttt{mathematics}. 

For curriculum-derived material, we normalize grade labels into regional examination levels: Grade 9 to \texttt{SSC-I}, Grade 10 to \texttt{SSC-II}, Grade 11 to \texttt{HSSC-I}, and Grade 12 to \texttt{HSSC-II}. Table~\ref{tab:urdummlu-domains-levels} in Appendix~\ref{app:acronyms} summarizes the final domain hierarchy, subdomains, acronyms, and examination levels covered in \ds{}. We also canonicalize the MCQ schema to support consistent evaluation. Each released item stores a question, four answer options, normalized domain and subdomain labels, academic level, source metadata, and answer annotations. We remove ambiguous index-based answer fields because different sources follow different option-ordering conventions.

\subsection{Cleaning and Quality Control}
\label{sec:quality-control}

We apply several cleaning and validation steps to reduce noise from OCR, web scraping, and heterogeneous source formatting. First, we normalize Urdu text representation through right-to-left display normalization, punctuation and quote normalization, standardization of fill-in-the-blank markers, and Unicode normalization for visually similar Arabic and Urdu codepoints. We then enforce structural validity by removing items with missing, empty, duplicate, or malformed answer options, discarding examples with invalid option counts, and standardizing option fields into a consistent schema format. Next, we deduplicate the candidate pool. We merge exact duplicates with consistent answers while preserving source provenance and discard duplicate groups with conflicting labels. To handle OCR and wording variations, we additionally apply conservative near-duplicate filtering based on high question-token overlap together with answer-option overlap. Finally, we remove residual non-Urdu artifacts, including a small number of English OCR artifacts that survived earlier filtering stages. We use the resulting cleaned pool for annotation, answer verification, and final benchmark construction.

\subsection{Human Annotation}
\label{sec:human-annotation}

The exam-derived portion of \ds{} did not include answer keys, so we produced gold labels through annotation. We organized annotation batches by subdomain and assigned each item to two annotators with relevant subject familiarity. Annotators selected the correct answer, marked questions as unsure, flagged problematic items, and could suggest light corrections to question text, answer options, and subdomain labels.

\begin{table}[!t]
\centering
\small
\renewcommand{\arraystretch}{1}

\begin{tabular*}{\columnwidth}{@{\extracolsep{\fill}}lrr}
\toprule
\multirow{2}{*}{\textbf{Provenance}} &
\multicolumn{2}{c}{\textbf{Composition}} \\
\cmidrule(lr){2-3}
& \textbf{Count}
& \textbf{Share (\%)} \\
\midrule

Answer-labeled web MCQs
& 13{,}630
& 51.7 \\

Exam-extracted annotated MCQs
& 12{,}759
& 48.3 \\

\midrule
\textbf{Total}
& \textbf{26{,}389}
& \textbf{100.0} \\

\bottomrule
\end{tabular*}

\caption{\textbf{Benchmark composition by provenance}. Web-derived questions use published human-validated answers, while exam-derived questions use dual annotation with consensus filtering.}
\label{tab:benchmark-provenance}
\end{table}

\begin{table}[!h]
\centering
\small
\renewcommand{\arraystretch}{1}

\begin{tabular*}{\columnwidth}{@{\extracolsep{\fill}}lrr}
\toprule
\multirow{2}{*}{\textbf{Domain}} &
\multicolumn{2}{c}{\textbf{Benchmark Composition}} \\
\cmidrule(lr){2-3}
& \textbf{Questions}
& \textbf{Subdomains} \\
\midrule

STEM
& 5{,}113
& 6 \\

Humanities
& 10{,}982
& 6 \\

Social Sciences
& 7{,}959
& 11 \\

Profession
& 973
& 2 \\

Other
& 1{,}362
& 1 \\

\midrule
\textbf{Total}
& \textbf{26{,}389}
& \textbf{26} \\

\bottomrule
\end{tabular*}

\caption{\textbf{Domain composition of \ds{}}. Question and subdomain counts across the five domains.}
\label{tab:urdummlu-main-stats}
\end{table}

Seventeen annotators participated in the process; 94.1\% identified Urdu as their native language, and most held either a bachelor's degree (47.1\%) or a master's degree (41.2\%). Appendix~\ref{app:annotation-details} provides full annotation details. Items were retained only when both annotators selected the same valid answer without flags or unsure labels. This process helped ensure high annotation quality and label reliability. In total, 17{,}565 exam-extracted MCQs entered annotation, and 14{,}459 satisfied the consensus criteria. Exclusions comprised 1{,}611 answer disagreements, 1{,}247 flagged items, 243 unsure selections, and five single annotations. Annotators also corrected 141 subdomain labels. After excluding flagged, unsure, and single-annotation items, 16{,}070 items remained answer-eligible. Of these, annotators agreed on 14{,}459 and disagreed on 1{,}611. Let $p_o=14{,}459/16{,}070=0.8998$ denote observed agreement. Using the fixed four-option chance baseline of 0.25, fixed-chance-adjusted agreement is $(p_o-0.25)/(1-0.25)=0.8663$. We report this measure rather than Cohen's $\kappa$, which estimates expected agreement from the annotators' observed marginals. Following verification, deduplication, and release packaging, 12{,}759 of the 14{,}459 exam-derived items were retained, a reduction of 1{,}700 items (11.8\%) through duplicate filtering, as shown in Table~\ref{tab:construction-ledger}. Annotators also verified answer labels for web-derived MCQs.

\begin{table}[!t]
\centering
\small
\renewcommand{\arraystretch}{1}

\begin{tabular*}{\columnwidth}{@{\extracolsep{\fill}}lrr}
\toprule
\multirow{2}{*}{\textbf{Group}} &
\multicolumn{2}{c}{\textbf{Mean Characters}} \\
\cmidrule(lr){2-3}
& \textbf{Question}
& \textbf{Answer} \\
\midrule

\rowcolor{softgray}
\multicolumn{3}{c}{\textbf{Domain}} \\
\midrule

Humanities
& 41.7
& 10.2 \\

Social Sciences
& 42.9
& 8.6 \\

STEM
& 46.0
& 7.9 \\

Other
& 48.7
& 9.7 \\

Profession
& 41.3
& 8.2 \\

\midrule
\rowcolor{softgray}
\multicolumn{3}{c}{\textbf{Academic Level}} \\
\midrule

SSC-I
& 43.5
& 9.3 \\

SSC-II
& 43.9
& 9.0 \\

HSSC-I
& 41.7
& 9.2 \\

HSSC-II
& 42.7
& 9.1 \\

\bottomrule
\end{tabular*}

\caption{\textbf{Text length in \ds{}.} Mean character counts for question stems and correct answers by domain and academic level. Stems exclude options, while Appendix~\ref{sec:dataset-format} reports serialized lengths including them.}
\label{tab:urdummlu-length-stats}
\end{table}

\subsection{Final Benchmark}
\label{sec:benchmark-composition}

After cleaning, annotation, answer verification, deduplication, and packaging, \ds{} contains 26{,}389 Urdu MCQs. Web sources contribute 13{,}630 answer-labeled questions, while exam sources contribute 12{,}759 questions retained through dual annotation and strict consensus filtering (Table~\ref{tab:benchmark-provenance}). Appendix~\ref{app:candidate-pool-analysis} describes the larger candidate pool before annotation and selection. The benchmark covers 26 subdomains across STEM, Humanities, Social Sciences, Profession, and Other. Humanities and Social Sciences form the largest shares, with broad coverage of Urdu language, literature, Islamic studies, Pakistan studies, and related content. Table~\ref{tab:urdummlu-main-stats} summarizes the domain-level composition, while Table~\ref{tab:urdummlu-domains-levels} lists the corresponding subdomains and academic levels. Table~\ref{tab:urdummlu-length-stats} reports average question and answer lengths, and Appendix~\ref{sec:dataset-format} describes the dataset schema.

\section{Experiments}
\label{sec:experiments}

We evaluate \ds{} with generation-based protocols that require each model to select an answer option for an Urdu MCQ. We run a large zero-shot evaluation across 30 open-weight and proprietary LLMs using both English and Urdu instruction prompts. We also run a focused few-shot study on four open-weight LLMs using one-, three-, and five-shot settings in both prompt languages. All evaluations use the same format and accuracy metric, enabling direct comparison across models, prompt languages, and few-shot settings.

\subsection{Models, Prompting, and Decoding}
\label{sec:models-prompting}

We evaluate 30 LLMs spanning different model sizes, access regimes, and training backgrounds, including proprietary API systems, open-weight multilingual models, compact models, mixture-of-experts architectures, reasoning-oriented variants, and Urdu- or regionally specialized models. Table~\ref{tab:language-models} in Appendix~\ref{app:model-roster} lists the full model roster. We evaluate each model with English and Urdu prompt templates while keeping the Urdu question stem, answer options, and response format fixed. For 29 models, the inference configuration is also matched across prompt languages. Qwen3.6-35B-A3B is the exception because reasoning was enabled only for its Urdu-prompt run. We therefore report both of its scores but exclude their difference from prompt-language conclusions. Appendix~\ref{app:prompt-templates} provides the templates in Figures~\ref{fig:eng_mcq_prompt} and~\ref{fig:urdu_mcq_prompt}. We use temperature 0 whenever deterministic decoding is available and otherwise follow provider-specific settings. We set the maximum output length to 4096 tokens, batch API requests with a concurrency of 10, and use greedy decoding for locally evaluated Hugging Face models.

\begin{table*}[t]
\centering
\resizebox{\textwidth}{!}{%
\begin{tabular}{l *{7}{r} c *{7}{r}}
\toprule
\multirow{2}{*}{\textbf{Model}} &
\multicolumn{7}{c}{\textbf{English Prompt (Accuracy \% $\uparrow$)}} &&
\multicolumn{7}{c}{\textbf{Urdu Prompt (Accuracy \% $\uparrow$)}} \\
\cmidrule(lr){2-8}\cmidrule(lr){10-16}
& \textbf{STEM} & \textbf{H} & \textbf{SS} & \textbf{P} & \textbf{O} & \textbf{Overall} & \textbf{Inv\%}\,$\downarrow$ &&
  \textbf{STEM} & \textbf{H} & \textbf{SS} & \textbf{P} & \textbf{O} & \textbf{Overall} & \textbf{Inv\%}\,$\downarrow$ \\
\midrule
\rowcolor{softgray}
\multicolumn{16}{c}{\textbf{Open-weight Models: $>$ 25B Parameters}} \\
\midrule
DeepSeek-V4-Flash
  & \textbf{97.38} & \textbf{68.79} & \textbf{90.09} & \textbf{90.03} & \textbf{86.56} & \textbf{82.45} & 0.1 &&
    \textbf{97.36} & \textbf{67.30} & \textbf{88.78} & \textbf{89.62} & \textbf{84.88} & \textbf{81.33} & 0.3 \\
Gemma-4-26B-A4B-IT
  & 85.92 & 57.58 & 73.92 & 75.64 & 70.70 & 69.34 & $<$0.1 &&
    87.19 & 57.87 & 75.46 & 77.90 & 71.95 & 70.32 & $<$0.1 \\
Gemma-4-31B-IT
  & 93.33 & 63.77 & 81.79 & 83.86 & 79.66 & 76.49 & $<$0.1 &&
    93.86 & 63.38 & 82.12 & 84.28 & 78.56 & 76.49 & $<$0.1 \\
LLaMA-3.3-70B
  & 81.32 & 56.45 & 75.54 & 76.67 & 73.94 & 68.67 & 0 &&
    78.39 & 56.24 & 73.51 & 71.43 & 71.81 & 67.10 & $<$0.1 \\
Qwen3.6-27B
  & 90.46 & 56.09 & 73.28 & 74.92 & 69.75 & 69.33 & 0 &&
    91.12 & 55.86 & 74.63 & 74.61 & 70.70 & 69.81 & 0 \\
Qwen3.6-35B-A3B$^{\dagger}$
  & 88.77 & 56.96 & 73.60 & 75.64 & 69.97 & 69.50 & 0 &&
    96.15 & 57.98 & 84.19 & 84.07 & 77.53 & 75.25 & 0.4 \\
LLaMA-4-Scout-17B-16E
  & 84.14 & 57.18 & 71.28 & 74.31 & 69.68 & 67.93 & 0 &&
    85.39 & 56.69 & 72.57 & 72.76 & 69.16 & 68.28 & $<$0.1 \\
LLaMA-4-Maverick-17B-128E
  & 91.96 & 63.43 & 80.20 & 81.81 & 80.84 & 75.59 & $<$0.1 &&
    92.24 & 63.41 & 81.38 & 79.75 & 80.98 & 75.93 & $<$0.1 \\
\midrule
\rowcolor{softgray}
\multicolumn{16}{c}{\textbf{Open-weight Models: $\leq$ 25B Parameters}} \\
\midrule
BLOOMZ-1.1B
  & 23.51 & 27.01 & 24.39 & 26.00 & 25.55 & 25.43 & 0.5 &&
    24.53 & 25.89 & 25.37 & 23.54 & 24.52 & 25.31 & $<$0.1 \\
BLOOMZ-1.7B
  & 27.52 & 27.14 & 33.16 & 35.15 & 30.25 & 29.48 & 2.5 &&
    17.15 & 20.78 & 28.91 & 27.34 & 25.92 & 23.04 & 24.7 \\
BLOOMZ-3B
  & 27.44 & 26.89 & 32.15 & 29.09 & 27.97 & 28.72 & 6.5 &&
    9.74 & 3.60 & 11.22 & 6.47 & 7.56 & 7.40 & 74.2 \\
BLOOMZ-7B
  & 24.98 & 23.93 & 30.38 & 29.09 & 27.53 & 26.45 & 11.1 &&
    22.75 & 17.75 & 22.73 & 20.04 & 22.54 & 20.55 & 34.5 \\
Gemma-2-9B-IT
  & 67.08 & \textbf{47.33} & 58.46 & \textbf{60.12} & \textbf{54.70} & \textbf{55.37} & 0 &&
    69.02 & \textbf{48.21} & 60.60 & \textbf{62.28} & 55.95 & 56.89 & 0 \\
Gemma-3-4B-IT
  & 49.97 & 37.96 & 47.73 & 47.89 & 45.67 & 44.00 & 0 &&
    51.79 & 38.37 & 48.75 & 50.26 & 46.33 & 44.95 & 0 \\
LLaMA-3.2-3B
  & 37.12 & 26.85 & 38.11 & 36.07 & 35.68 & 33.03 & 0 &&
    37.24 & 29.39 & 40.02 & 38.44 & 38.11 & 34.90 & 0 \\
LLaMA-3.1-8B
  & 46.96 & 36.47 & 49.72 & 48.61 & 44.64 & 43.37 & 0 &&
    46.49 & 37.71 & 49.91 & 49.64 & 45.08 & 43.91 & 0 \\
Ministral-3-3B
  & 55.04 & 43.80 & 49.04 & 49.13 & 48.02 & 47.97 & $<$0.1 &&
    57.25 & 43.17 & 52.13 & 52.31 & 48.97 & 49.24 & $<$0.1 \\
Ministral-3-8B
  & 67.81 & 45.38 & \textbf{59.05} & 59.71 & 54.55 & 54.85 & 0 &&
    71.37 & 45.86 & \textbf{62.81} & 61.66 & \textbf{57.12} & \textbf{57.08} & 0 \\
Phi-4-mini
  & 37.02 & 28.92 & 38.13 & 40.49 & 32.75 & 33.89 & $<$0.1 &&
    37.08 & 28.77 & 39.01 & 38.75 & 35.90 & 34.21 & $<$0.1 \\
Phi-3.5-mini
  & 33.76 & 22.95 & 30.93 & 32.27 & 28.34 & 28.07 & 0 &&
    33.54 & 27.22 & 31.70 & 34.22 & 30.62 & 30.23 & 0.4 \\
Qwen3-4B-Instruct-2507
  & 68.61 & 42.44 & 51.54 & 52.72 & 47.94 & 50.92 & 0 &&
    68.75 & 43.11 & 53.36 & 53.34 & 47.80 & 51.79 & 0 \\
Qwen3-8B
  & \textbf{70.70} & 39.31 & 54.05 & 56.73 & 50.29 & 51.05 & 0 &&
    \textbf{74.05} & 30.77 & 56.39 & 56.83 & 49.49 & 48.81 & 0.5 \\
\midrule
\rowcolor{softgray}
\multicolumn{16}{c}{\textbf{Proprietary Models}} \\
\midrule
Claude-Haiku-4.5
  & 90.44 & 58.67 & 75.80 & 76.05 & 74.30 & 71.44 & 0.1 &&
    91.92 & 59.42 & 77.06 & 78.42 & 74.45 & 72.52 & $<$0.1 \\
Claude-Sonnet-4.6
  & 96.34 & 72.87 & 87.46 & 87.36 & 86.20 & 83.04 & 0 &&
    96.26 & 72.87 & 87.62 & 87.36 & 86.05 & 83.07 & 0 \\
Gemini-3.1-Flash-Lite
  & 96.85 & 74.39 & 90.11 & 90.44 & 86.27 & 84.69 & $<$0.1 &&
    97.09 & 74.57 & 90.20 & 90.54 & 85.76 & 84.81 & $<$0.1 \\
Gemini-3.5-Flash
  & \fbox{\textbf{97.56}} & \fbox{\textbf{85.16}} & \fbox{\textbf{92.16}} & \fbox{\textbf{92.29}} & \fbox{\textbf{90.82}} & \fbox{\textbf{90.23}} & 0.1 &&
    \fbox{\textbf{97.75}} & \fbox{\textbf{85.49}} & \fbox{\textbf{92.25}} & \fbox{\textbf{91.57}} & \fbox{\textbf{91.85}} & \fbox{\textbf{90.45}} & $<$0.1 \\
GPT-5.4-mini
  & 88.34 & 62.98 & 77.61 & 79.24 & 75.40 & 73.55 & 0 &&
    88.25 & 62.51 & 78.54 & 79.75 & 75.26 & 73.63 & 0 \\
GPT-5.4
  & 95.13 & 69.47 & 86.44 & 86.02 & 84.29 & 80.94 & 0 &&
    97.40 & 74.29 & 89.47 & 87.26 & 83.92 & 84.32 & 0.4 \\
\midrule
\rowcolor{softgray}
\multicolumn{16}{c}{\textbf{Urdu Models}} \\
\midrule
Qalb-1.0-8B
  & 38.18 & \textbf{30.07} & 37.72 & \textbf{40.39} & 39.65 & \textbf{34.82} & 0 &&
    \textbf{30.59} & \textbf{31.39} & \textbf{31.12} & \textbf{35.46} & \textbf{36.05} & \textbf{31.54} & 11.3 \\
Alif-1.0-8B
  & \textbf{40.62} & 25.75 & \textbf{41.11} & 39.26 & \textbf{41.04} & 34.55 & 0.6 &&
    29.04 & 27.98 & 28.76 & 30.94 & 29.66 & 28.62 & 12.6 \\
\bottomrule
\end{tabular}%
}

\caption{\textbf{Model performance on \ds{}}. Accuracy (\%) under English and Urdu prompts across five domains and all $26{,}389$ items. H, SS, P, and O denote Humanities, Social Sciences, Profession, and Other. Invalid and error outputs count as incorrect, with their rate reported as Inv\%. Boxes mark column-best scores, and bold marks the best within each model group. $^{\dagger}$Qwen3.6-35B-A3B uses reasoning only under the Urdu prompt, so its cross-prompt difference is excluded from prompt-language comparisons.}

\label{tab:modelperformance}
\end{table*}

\subsection{Evaluation Protocols}
\label{sec:evaluation-protocols}

\paragraph{Zero-shot evaluation:} We use a generation-based zero-shot protocol in which each input contains the domain, subdomain, academic level, Urdu question, and labeled answer options. We evaluate all 30 models under both English and Urdu prompt templates, resulting in 60 zero-shot runs. Since the question and answer options remain unchanged across settings, this protocol isolates the effect of instruction language on the same Urdu MCQs.

\paragraph{Few-shot evaluation:} We conduct a controlled few-shot study on four open-weight LLMs under one-, three-, and five-shot settings with both English and Urdu prompts, yielding 24 runs. We reserve 130 validated MCQs as a demonstration pool and run these evaluations with \texttt{lm-evaluation-harness}~\citep{eval-harness}, which we use only for prompt construction, demonstration sampling, and execution management. Models generate the answer key and answer text, and the harness scores the generated continuation against the gold label by exact match. We exclude the 130 demonstration items from scoring, so all shot settings in Table~\ref{tab:fewshot}, including the zero-shot baseline, are compared on the same 26{,}259 items.

\subsection{Evaluation Measure}
\label{sec:metrics}

Our primary measure is all-item exact-match accuracy. We parse an answer key from each generated response, compare it with the gold label, and score any response that is unparsable, malformed, or returned as an error as incorrect. The denominator is therefore every benchmark item, which keeps accuracy comparable across models that differ widely in how reliably they follow the response format. We report the invalid-output rate separately, defined as the percentage of responses falling into those three failure categories. Accuracy restricted to parseable responses is a strictly different quantity, conditional on the model having produced a readable answer, so wherever we report it we label it as such and pair it with the invalid rate. We further analyze results by domain, subdomain, academic level, prompt language, and model category to examine performance differences across standard academic and Urdu- or region-specific subjects.

\section{Results}
\label{sec:results}

We analyze overall zero-shot performance, then compare domains, prompt languages, model scales, and few-shot settings. Table~\ref{tab:modelperformance} summarizes accuracy under English and Urdu prompts with invalid-output rates. We focus on four main findings: a small set of models performs strongly, STEM transfers much better than Urdu-centered Humanities, prompt language has limited effect for most models, and few-shot prompting gives modest but insufficient gains. Appendix~\ref{app:detailed-results} provides detailed per-subdomain results.

\subsection{Overall Model Performance}
\label{sec:aggregate-accuracy}

Gemini-3.5-Flash leads with $90.23\%$ accuracy under the English prompt and $90.45\%$ under the Urdu prompt, while all others remain below $85\%$. Gemini-3.1-Flash-Lite~\citep{google2026gemini31flashlite}, GPT-5.4~\citep{singh2026openaigpt5card}, Claude-Sonnet-4.6~\citep{anthropic2026sonnet46}, and DeepSeek-V4-Flash~\citep{deepseekai2026deepseekv4} rank next.

DeepSeek-V4-Flash is the strongest open-weight model, scoring $82.45\%$ under the English prompt and $81.33\%$ under the Urdu prompt, but trails Gemini-3.5-Flash by $7.78$ and $9.12$ points. Performance falls sharply below this tier. Among open-weight models with $\leq 25$B parameters, Gemma-2-9B-IT~\citep{gemma2024gemma2} and Ministral-3-8B~\citep{liu2026ministral3} lead at about $55$--$57\%$, while Qwen3-4B and Qwen3-8B~\citep{qwen2025qwen3} remain near $50\%$. BLOOMZ~\citep{muennighoff-etal-2023-crosslingual} stays near or below the $25\%$ random baseline and drops further under the Urdu prompt because many outputs are unparsable (Section~\ref{sec:output-validity}). Qalb-1.0-8B~\citep{hassan2026qalb} and Alif-1.0-8B~\citep{shafique-etal-2025-alif} remain below $35\%$, showing Urdu-focused tuning alone does not ensure broad Urdu knowledge.

\subsection{Domain-Level Performance}
\label{sec:domain-breakdown}

\begin{figure}[!t]
    \centering
    \includegraphics[width=\linewidth]{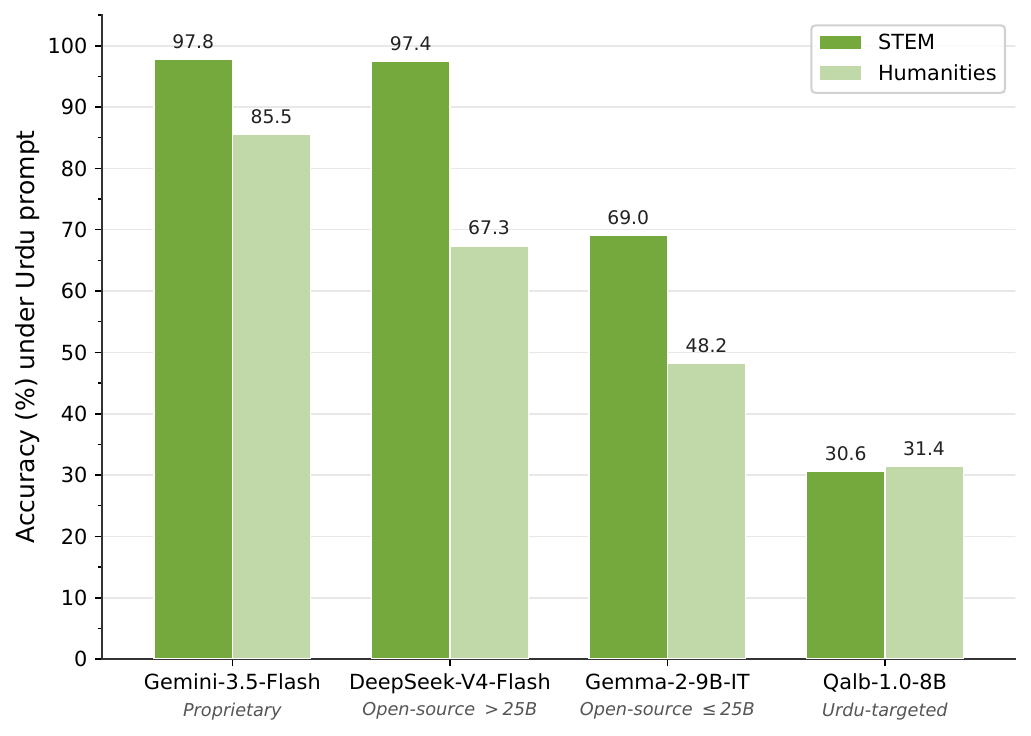}
    \caption{\textbf{Accuracy across domains.} Under Urdu prompting, the first three models favor STEM, while Qalb performs similarly across domains, with Humanities 0.80 points higher.}
    \label{fig:stem-hum-gap}
\end{figure}

Domain results show a consistent STEM--Humanities gap. Nearly every above-chance model scores highest on STEM and lowest on Humanities. Near-chance BLOOMZ checkpoints and Qalb-1.0-8B are exceptions because their STEM accuracy is already low. Under the Urdu prompt, Gemini-3.5-Flash drops from $97.75\%$ on STEM to $85.49\%$ on Humanities, while DeepSeek-V4-Flash falls from $97.36\%$ to $67.30\%$. GPT-5.4 and Claude-Sonnet-4.6 lose over $23$ points, and several Qwen models lose over $35$. Figure~\ref{fig:stem-hum-gap} shows this pattern across model groups. Humanities covers Urdu literature, language, Islamic studies, ethics, and related cultural content, while STEM overlaps more with English-language training data. Models often answer science questions in Urdu but struggle with literary, linguistic, and religious content. This gap may reflect training exposure, source mix, item difficulty, or label quality, which our design cannot isolate. Social Sciences falls between the two, combining global and region-specific knowledge.

\begin{figure}[!h]
    \centering
    \includegraphics[width=\linewidth]{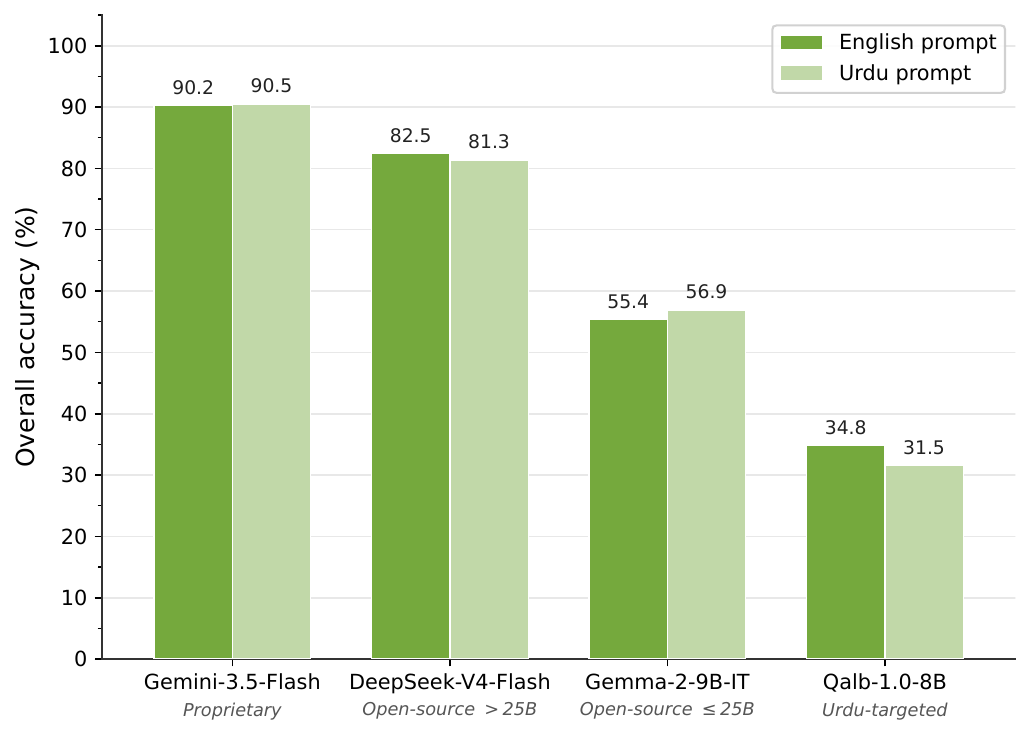}
    \caption{\textbf{Accuracy across prompt languages.} Results for representative models from each group show little effect of prompt language.}
    \label{fig:overall-en-vs-ur}
\end{figure}

\begin{table*}[!t]
\centering
\small
\renewcommand{\arraystretch}{1}

\resizebox{\textwidth}{!}{%
\begin{tabular}{@{}l *{4}{c} c *{4}{c}@{}}
\toprule
\multirow{2}{*}{\textbf{Model}} &
\multicolumn{4}{c}{\textbf{English Prompt (Accuracy \% $\uparrow$)}} &&
\multicolumn{4}{c}{\textbf{Urdu Prompt (Accuracy \% $\uparrow$)}} \\
\cmidrule(lr){2-5}
\cmidrule(lr){7-10}
& \textbf{0-shot}
& \textbf{1-shot}
& \textbf{3-shot}
& \textbf{5-shot}
&& \textbf{0-shot}
& \textbf{1-shot}
& \textbf{3-shot}
& \textbf{5-shot} \\
\midrule

\rowcolor{softgray}
\multicolumn{10}{c}{\textbf{Evaluated Open-weight Models}} \\
\midrule

LLaMA-3.1-8B
& 43.35
& 45.00\,\textcolor{darkgreen}{\scriptsize($+$1.66)}
& 46.16\,\textcolor{darkgreen}{\scriptsize($+$2.81)}
& 46.66\,\textcolor{darkgreen}{\scriptsize($+$3.32)}
&& 43.89
& 45.15\,\textcolor{darkgreen}{\scriptsize($+$1.25)}
& 46.38\,\textcolor{darkgreen}{\scriptsize($+$2.49)}
& 46.17\,\textcolor{darkgreen}{\scriptsize($+$2.28)} \\

Gemma-3-4B-IT
& 43.98
& 45.97\,\textcolor{darkgreen}{\scriptsize($+$1.99)}
& 46.13\,\textcolor{darkgreen}{\scriptsize($+$2.14)}
& 46.31\,\textcolor{darkgreen}{\scriptsize($+$2.33)}
&& 44.91
& 46.45\,\textcolor{darkgreen}{\scriptsize($+$1.54)}
& 46.40\,\textcolor{darkgreen}{\scriptsize($+$1.49)}
& 46.04\,\textcolor{darkgreen}{\scriptsize($+$1.13)} \\

Qwen3-8B
& 51.02
& 50.30\,\textcolor{darkred}{\scriptsize($-$0.72)}
& 52.33\,\textcolor{darkgreen}{\scriptsize($+$1.31)}
& 53.27\,\textcolor{darkgreen}{\scriptsize($+$2.25)}
&& 48.73
& 51.91\,\textcolor{darkgreen}{\scriptsize($+$3.19)}
& 53.31\,\textcolor{darkgreen}{\scriptsize($+$4.58)}
& 53.56\,\textcolor{darkgreen}{\scriptsize($+$4.83)} \\

Qwen3-4B-Instruct-2507
& 50.92
& 52.64\,\textcolor{darkgreen}{\scriptsize($+$1.71)}
& 53.49\,\textcolor{darkgreen}{\scriptsize($+$2.57)}
& 53.72\,\textcolor{darkgreen}{\scriptsize($+$2.80)}
&& 51.80
& 52.14\,\textcolor{darkgreen}{\scriptsize($+$0.34)}
& 52.97\,\textcolor{darkgreen}{\scriptsize($+$1.17)}
& 53.00\,\textcolor{darkgreen}{\scriptsize($+$1.20)} \\

\midrule
\textbf{Mean}
& \textbf{47.32}
& \textbf{48.48\,\textcolor{darkgreen}{\scriptsize($+$1.16)}}
& \textbf{49.53\,\textcolor{darkgreen}{\scriptsize($+$2.21)}}
& \textbf{49.99\,\textcolor{darkgreen}{\scriptsize($+$2.67)}}
&& \textbf{47.33}
& \textbf{48.91\,\textcolor{darkgreen}{\scriptsize($+$1.58)}}
& \textbf{49.77\,\textcolor{darkgreen}{\scriptsize($+$2.43)}}
& \textbf{49.69\,\textcolor{darkgreen}{\scriptsize($+$2.36)}} \\

\bottomrule
\end{tabular}%
}

\caption{\textbf{Few-shot performance on \ds{}}. Accuracy (\%) at $0$-, $1$-, $3$-, and $5$-shot under English and Urdu prompts. All settings use the same $26{,}259$ items after excluding 130 held-out demonstrations. Zero-shot values differ by at most $0.08$ points from Table~\ref{tab:modelperformance}, which uses all $26{,}389$ items. Deltas are relative to zero-shot, with \textcolor{darkgreen}{green} indicating gains and \textcolor{darkred}{red} indicating losses. Means cover the four models.}

\label{tab:fewshot}
\end{table*}

\subsection{Prompt-Language Effects}
\label{sec:prompt-language}

Under matched inference settings, prompt language has little effect on overall accuracy. Figure~\ref{fig:overall-en-vs-ur} shows small changes for models with reliable formatting, including $+0.22$ points for Gemini-3.5-Flash, $-1.12$ for DeepSeek-V4-Flash, and $+1.52$ for Gemma-2-9B-IT. Qalb-1.0-8B loses $3.28$ points as its invalid-output rate rises to $11.3\%$ under the Urdu prompt, indicating that formatting contributes to the decline.

Among models with invalid-output rates below $1\%$ in both settings, most change by less than one point. Larger matched differences include gains of $3.38$ points for GPT-5.4 and losses of $2.24$ for Qwen3-8B and $5.93$ for Alif-1.0-8B. We exclude Qwen3.6-35B-A3B~\citep{qwen36_35b_a3b} from prompt-language conclusions because reasoning was enabled only under its Urdu prompt. Although its scores differ by $5.75$ points, prompt language is not the only change. Under matched settings, prompt-language differences remain smaller than the STEM--Humanities gaps in \ds{}.

\begin{figure}[!t]
    \centering
    \includegraphics[width=\linewidth]{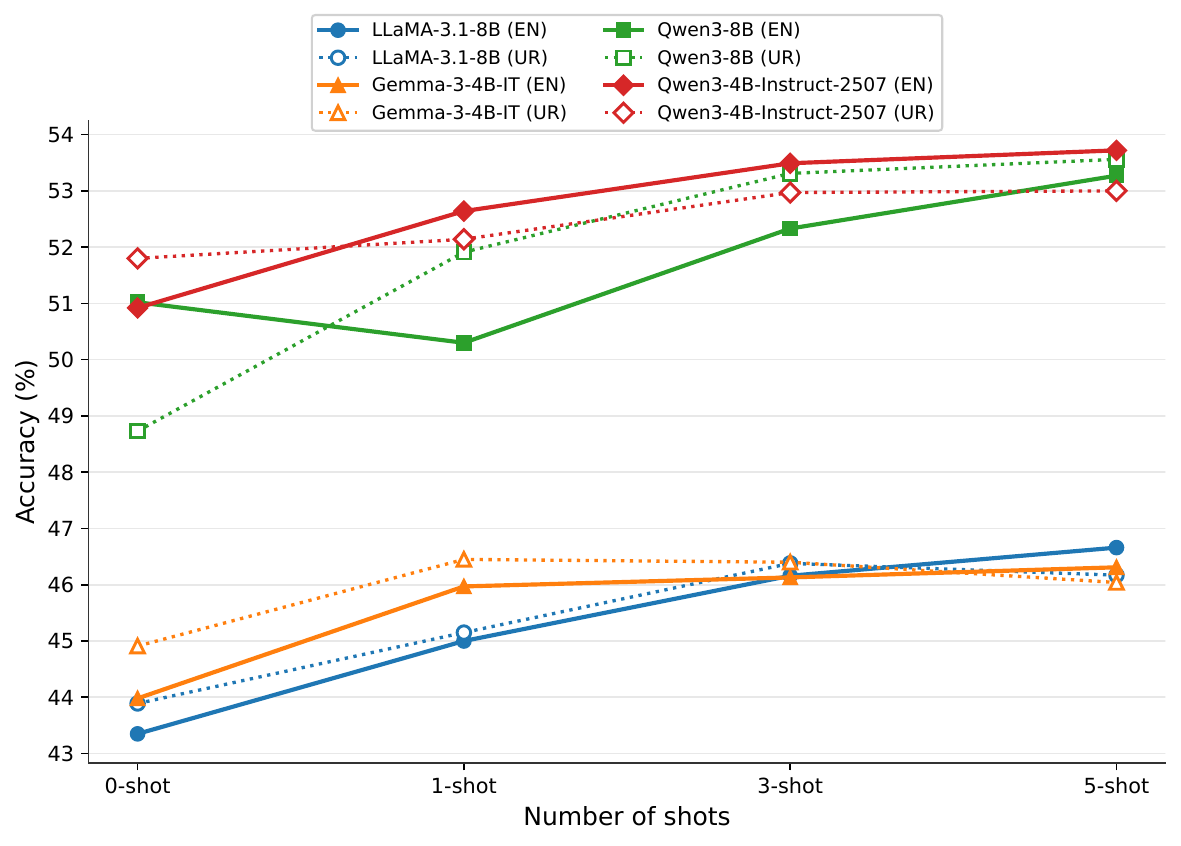}
    \caption{\textbf{Few-shot accuracy for four open-weight models under English (solid) and Urdu (dotted) prompts.} Accuracy generally improves from zero-shot to five-shot, with modest gains in both languages.}
    \label{fig:fewshot}
\end{figure}

\subsection{Invalid-Output Rates}
\label{sec:output-validity}

Invalid-output rates distinguish incorrect answers from failures to follow the response format. Most proprietary and larger open-weight models remain below $0.1\%$, while weaker models fail more often under the Urdu prompt. 

Rates reach $74.2\%$ for BLOOMZ-3B, $34.5\%$ for BLOOMZ-7B, $24.7\%$ for BLOOMZ-1.7B, $11.3\%$ for Qalb-1.0-8B, and $12.6\%$ for Alif-1.0-8B. We therefore score all items, including invalid outputs. BLOOMZ-3B scores $28.66\%$ on parseable Urdu responses but only $7.40\%$ overall, showing how parseable-only accuracy can hide format failure. Reporting both measures captures answer accuracy and instruction following. Appendix~\ref{app:invalid-outputs} provides examples.

\subsection{Few-Shot Performance}
\label{sec:fewshot-results}

Table~\ref{tab:fewshot} and Figure~\ref{fig:fewshot} report results for four open-weight models under English and Urdu prompts. Demonstrations use a released 130-item pool excluded from every setting, including zero-shot, so all columns evaluate $26{,}259$ items. Few-shot prompting improves $23$ of $24$ configurations. Mean gains at one-, three-, and five-shot are $+1.16$, $+2.21$, and $+2.67$ points in English and $+1.58$, $+2.43$, and $+2.36$ in Urdu.

Qwen3-8B shows the largest gain, rising from $48.73\%$ to $53.56\%$ under the Urdu prompt. At five-shot, all prompt-language gaps are within $0.72$ points. Few-shot prompting changes some local rankings but does not close the gap with larger models or replace missing Urdu-specific knowledge.

\subsection{Comparison with a Translated Urdu Benchmark}
\label{sec:native-vs-translated}

\begin{table*}[!t]
\centering
\resizebox{\textwidth}{!}{%
\begin{tabular}{l c *{5}{rrr}}
\toprule
\multirow{2}{*}{\textbf{Subject}} &
\multirow{2}{*}{\textbf{\begin{tabular}[c]{@{}c@{}}$n$\\ (U\,/\,P)\end{tabular}}} &
\multicolumn{3}{c}{\textbf{Gemma-4-31B-IT}} &
\multicolumn{3}{c}{\textbf{Qwen3.6-35B-A3B}$^{\dagger}$} &
\multicolumn{3}{c}{\textbf{Qwen3.6-27B}} &
\multicolumn{3}{c}{\textbf{LLaMA-3.1-8B}} &
\multicolumn{3}{c}{\textbf{LLaMA-3.2-3B}} \\
\cmidrule(lr){3-5}
\cmidrule(lr){6-8}
\cmidrule(lr){9-11}
\cmidrule(lr){12-14}
\cmidrule(lr){15-17}
& & \textbf{U} & \textbf{P} & \textbf{$\Delta$}
  & \textbf{U} & \textbf{P} & \textbf{$\Delta$}
  & \textbf{U} & \textbf{P} & \textbf{$\Delta$}
  & \textbf{U} & \textbf{P} & \textbf{$\Delta$}
  & \textbf{U} & \textbf{P} & \textbf{$\Delta$} \\
\midrule

\rowcolor{softgray}
\multicolumn{17}{c}{\textbf{STEM}} \\
\midrule

Chemistry & 936\,/\,115
& $92.5$ & $80.7$ & \textcolor{darkgreen}{$+11.8^{*}$}
& $97.1$ & $94.8$ & \textcolor{darkgreen}{$+2.3$}
& $92.9$ & $81.7$ & \textcolor{darkgreen}{$+11.2^{*}$}
& $42.2$ & $40.9$ & \textcolor{darkgreen}{$+1.3$}
& $32.8$ & $35.7$ & \textcolor{darkred}{$-2.9$} \\

Biology & 931\,/\,139
& $94.0$ & $91.7$ & \textcolor{darkgreen}{$+2.3$}
& $95.9$ & $93.4$ & \textcolor{darkgreen}{$+2.5$}
& $88.8$ & $92.4$ & \textcolor{darkred}{$-3.5$}
& $46.0$ & $48.1$ & \textcolor{darkred}{$-2.1$}
& $36.0$ & $37.3$ & \textcolor{darkred}{$-1.3$} \\

Mathematics & 876\,/\,242
& $93.6$ & $68.3$ & \textcolor{darkgreen}{$+25.3^{*}$}
& $96.9$ & $96.9$ & \textcolor{darkgreen}{$+0.1$}
& $88.9$ & $61.2$ & \textcolor{darkgreen}{$+27.8^{*}$}
& $43.4$ & $31.4$ & \textcolor{darkgreen}{$+12.0^{*}$}
& $37.6$ & $30.7$ & \textcolor{darkgreen}{$+6.9^{*}$} \\

Computer Science & 823\,/\,62
& $94.4$ & $93.9$ & \textcolor{darkgreen}{$+0.5$}
& $96.5$ & $94.8$ & \textcolor{darkgreen}{$+1.6$}
& $93.1$ & $91.3$ & \textcolor{darkgreen}{$+1.8$}
& $51.8$ & $45.8$ & \textcolor{darkgreen}{$+6.0$}
& $41.3$ & $33.2$ & \textcolor{darkgreen}{$+8.1$} \\

Physics & 621\,/\,127
& $95.7$ & $80.0$ & \textcolor{darkgreen}{$+15.7^{*}$}
& $96.9$ & $89.0$ & \textcolor{darkgreen}{$+8.0^{*}$}
& $94.8$ & $79.2$ & \textcolor{darkgreen}{$+15.6^{*}$}
& $43.6$ & $36.2$ & \textcolor{darkgreen}{$+7.4$}
& $36.6$ & $33.2$ & \textcolor{darkgreen}{$+3.3$} \\

\midrule

\rowcolor{softgray}
\multicolumn{17}{c}{\textbf{Social Sciences}} \\
\midrule

Economics & 907\,/\,358
& $84.7$ & $86.3$ & \textcolor{darkred}{$-1.6$}
& $89.7$ & $92.3$ & \textcolor{darkred}{$-2.5$}
& $81.5$ & $85.9$ & \textcolor{darkred}{$-4.4$}
& $49.3$ & $40.3$ & \textcolor{darkgreen}{$+8.9^{*}$}
& $36.5$ & $32.3$ & \textcolor{darkgreen}{$+4.2$} \\

Sociology & 778\,/\,52
& $91.1$ & $82.7$ & \textcolor{darkgreen}{$+8.4$}
& $91.4$ & $83.5$ & \textcolor{darkgreen}{$+7.9$}
& $89.3$ & $79.2$ & \textcolor{darkgreen}{$+10.1$}
& $68.9$ & $43.8$ & \textcolor{darkgreen}{$+25.0^{*}$}
& $53.3$ & $34.6$ & \textcolor{darkgreen}{$+18.7^{*}$} \\

Geography & 630\,/\,65
& $87.0$ & $82.8$ & \textcolor{darkgreen}{$+4.2$}
& $92.9$ & $85.5$ & \textcolor{darkgreen}{$+7.3$}
& $84.3$ & $82.5$ & \textcolor{darkgreen}{$+1.8$}
& $53.3$ & $39.4$ & \textcolor{darkgreen}{$+13.9^{*}$}
& $42.5$ & $41.5$ & \textcolor{darkgreen}{$+1.0$} \\

Psychology & 460\,/\,169
& $90.4$ & $85.2$ & \textcolor{darkgreen}{$+5.2$}
& $91.3$ & $88.0$ & \textcolor{darkgreen}{$+3.3$}
& $85.0$ & $83.3$ & \textcolor{darkgreen}{$+1.7$}
& $44.3$ & $41.4$ & \textcolor{darkgreen}{$+2.9$}
& $34.6$ & $36.0$ & \textcolor{darkred}{$-1.4$} \\

\midrule
\textbf{Macro avg} & \textbf{6{,}962\,/\,1{,}329}
& $\mathbf{91.5}$ & $\mathbf{83.5}$ & \textcolor{darkgreen}{$\mathbf{+8.0}$}
& $\mathbf{94.3}$ & $\mathbf{90.9}$ & \textcolor{darkgreen}{$\mathbf{+3.4}$}
& $\mathbf{88.7}$ & $\mathbf{81.8}$ & \textcolor{darkgreen}{$\mathbf{+6.9}$}
& $\mathbf{49.2}$ & $\mathbf{40.8}$ & \textcolor{darkgreen}{$\mathbf{+8.4}$}
& $\mathbf{39.0}$ & $\mathbf{34.9}$ & \textcolor{darkgreen}{$\mathbf{+4.1}$} \\

\bottomrule
\end{tabular}%
}

\caption{\textbf{Cross-benchmark gaps on shared subjects.} Zero-shot accuracy (\%) on nine subjects shared by \ds{} (U) and MMLU-ProX Urdu (P), with $\Delta=\mathrm{U}-\mathrm{P}$. Using all \ds{} curriculum levels gives 6{,}962 U and 1{,}329 P items. MMLU-ProX items are reduced to four options and averaged over five distractor samples. Green and red denote positive and negative gaps. $^{*}$ marks gaps whose 95\% confidence intervals exclude zero. Because items are not matched, the gaps are descriptive rather than causal effects of language, translation, or format. Macro averages weight subjects equally. $^{\dagger}$Qwen3.6-35B-A3B uses reasoning for both arms, so its gap is comparable within the model but its absolute accuracy is not comparable with the others.}

\label{tab:native-vs-prox}
\end{table*}

\dataset{MMLU-ProX}~\citep{xuan-etal-2025-mmlu} provides a translated Urdu reference for \ds{}. Nine subjects overlap, yielding $6{,}962$ \ds{} questions across all curriculum levels and $1{,}329$ \dataset{MMLU-ProX} questions. We reduce each \dataset{MMLU-ProX} item to four options by retaining the answer and sampling three distractors, then average over five randomized samples. Both benchmarks are evaluated under Urdu prompts with a $25\%$ guessing floor. Because they contain different questions and unequal subject-level samples, the results are descriptive cross-benchmark gaps rather than causal effects of language, translation, or format. Macro averages weight subjects equally. Models score higher on \ds{}, with gaps from $+3.4$ to $+8.4$ points. This does not rank benchmark quality but shows translated questions remain difficult. Qwen3.6-35B-A3B uses reasoning for both arms, making its gap valid but its absolute accuracy incomparable with other models. The aggregate gaps hide clear subject-level variation. Mathematics and Physics show the largest differences, reaching $+27.8$ and $+15.6$ points for Qwen3.6-27B and $+25.3$ and $+15.7$ for Gemma-4-31B-IT. Conversely, the three larger models score $1.6$ to $4.4$ points higher on \dataset{MMLU-ProX} Economics. Biology shows small gaps. Macro averages mask subject variation rather than a consistent benchmark advantage.

The main contribution of \ds{} is its coverage, which extends Urdu evaluation in two ways:

\begin{itemize}
    \item \textbf{Broader shared-subject coverage:} \ds{} contains over five times as many questions across the nine shared subjects, supporting stable subdomain (Appendix~\ref{app:detailed-results}) and education-level analysis.

    \item \textbf{Urdu-specific coverage:} \ds{} includes Urdu language, grammar, literature, Islamic studies, Pakistan studies, and other local curricular topics absent from the English-derived \dataset{MMLU-ProX}.
\end{itemize}

Unlike \dataset{MMLU-ProX}, \ds{} evaluates models on questions drawn from Urdu educational sources. Its broader coverage of linguistic, curricular, and cultural content tests knowledge taught and assessed in Urdu rather than translated question answering alone.
\section{Conclusion and Future Work}
\label{sec:conclusion}

We introduced \ds{}, a broad-coverage, natively written \dataset{MMLU}-style benchmark for Urdu with 26{,}389 MCQs across 26 subjects and five domains, collected from Urdu MCQ banks and public SSC/HSSC examination PDFs. The benchmark combines standard academic subjects with Urdu- and region-specific content and uses dual human annotation with strict consensus filtering for exam-derived questions. 

Evaluating 30 open-weight and proprietary LLMs under English and Urdu prompts, scoring every item and counting unparsable outputs as incorrect, reveals a clear gap in current model capability. Gemini-3.5-Flash performs best at 90.23\% and 90.45\% accuracy, while the strongest open-weight model trails by 7.78 and 9.12 points. Models perform substantially better on STEM than on Urdu-centered Humanities, often losing 25 to 40 points on Urdu literature, Urdu language, and Islamic studies. Prompt language has limited effect for models that follow the response format reliably, and few-shot prompting yields only modest gains. Overall, \ds{} shows that capable models remain uneven across Urdu educational and cultural content, and provides a stronger foundation for evaluating Urdu-capable LLMs.

Future work can extend \ds{} beyond MCQ-based evaluation through open-ended generation, summarization, and translation tasks. Expanding the benchmark to include Indian Urdu curricula, undergraduate material, professional examinations, and dialectal content would further broaden its scope. Psychometrics also remains difficult for all evaluated models, motivating future Urdu reasoning benchmarks focused on analogies, logical patterns, and aptitude-style tasks. Finally, weak Urdu-targeted model performance calls for continued pretraining and instruction tuning on native educational and literary Urdu material. These efforts should cover different educational levels and regional Urdu varieties.
\section*{Limitations}
\label{sec:limitations}

\paragraph{Curriculum and source scope:} \ds{} focuses on the Pakistani SSC/HSSC curriculum and a limited set of Urdu MCQ websites targeting the same educational setting. Strong performance therefore reflects competence on Pakistani secondary-school material rather than Urdu in its full linguistic diversity. The benchmark does not cover undergraduate content, Indian Urdu curricula, dialectal variation, or Urdu--English code-switching. A small subset of Islamic Studies questions includes Quranic or Classical Arabic text, reflecting its use in the Pakistani Urdu-medium curriculum; these questions constitute only a small fraction of the overall benchmark and do not materially affect its characterization as Urdu-focused. Although we reduce source skew through deduplication, annotation, and balancing, Ustad~360 still contributes $57.8\%$ of the cleaned candidate pool and $41.9\%$ of the released benchmark.

\paragraph{Format and ceiling effects:} \ds{} uses a four-option multiple-choice format and therefore does not evaluate open-ended writing, summarization, translation quality, long-form reasoning, or conversational ability. The psychometrics subdomain adds reasoning-heavy items, on which no model exceeds $60\%$ accuracy, but reasoning-heavy MCQs do not remove the limitation of the multiple-choice format itself. Future work should extend evaluation toward more open-ended Urdu tasks.

\paragraph{Prompt-language and few-shot effects are limited:} English and Urdu instruction wrappers, together with 1-, 3-, and 5-shot prompting, change accuracy by only a few points and rarely alter model rankings. We also do not evaluate option-order robustness, chain-of-thought prompting, or specialized reasoning modes. Our setup therefore prioritizes consistency and comparability over fully optimized prompting configurations.

\paragraph{Cross-benchmark comparison.}

The comparison with MMLU-ProX in Section~\ref{sec:native-vs-translated} is descriptive rather than causal. The benchmarks use different, unmatched questions, curricula, question-stem lengths (median 39 vs.\ 112 characters, excluding answer options), and difficulty profiles. Reducing MMLU-ProX to four options equalises the guessing floor but not difficulty. We therefore interpret Table~\ref{tab:native-vs-prox} as a comparison of benchmark coverage and measurement scope, not as evidence of the effect of native authorship.

\section*{Ethical Statement \& Broad Impact}

We develop \ds{} to support more inclusive multilingual evaluation for Urdu, a widely spoken but underrepresented language in NLP research. The benchmark draws from publicly available educational and examination material and aims to improve evaluation coverage beyond English-centered benchmarks.

\paragraph{Transparency and Reproducibility:} We release the dataset, evaluation code, and prompting protocols to support reproducible research and transparent model comparison. We document the construction pipeline, annotation procedure, and evaluation setup in detail. The release also records the dataset version, code revision, and checksums needed to identify the evaluated artifacts.

\paragraph{Annotation and Data Quality:} We use dual human annotation with strict consensus filtering for exam-derived questions and additionally verify answer labels for web-derived items. We further apply cleaning, deduplication, and normalization procedures to reduce OCR noise, malformed questions, and metadata inconsistencies. Items with disagreement, uncertainty, flags, or incomplete annotations are excluded rather than assigned a gold label.

\paragraph{Bias and Scope Limitations:}
\ds{} primarily reflects the Pakistani SSC/HSSC curriculum and the educational content available through Urdu MCQ resources. As a result, it may not fully represent other Urdu-speaking communities, dialects, or educational systems. The benchmark also contains culturally and regionally grounded subjects such as Islamic studies and Pakistan studies that reflect the underlying curriculum sources. Results should therefore be interpreted within this curricular context rather than as a complete measure of Urdu knowledge.

\paragraph{Broader Impact:} We hope \ds{} supports the development of stronger Urdu-capable language models and more representative multilingual evaluation. At the same time, benchmark scores should not be interpreted as complete measures of reasoning ability, factual reliability, or cultural understanding beyond the educational scope represented in the dataset. They should not be used alone for high-stakes decisions about model deployment or educational suitability in practical settings.

\newpage

\bibliography{custom}

\appendix
\clearpage

\section{Candidate Pool Analysis}
\label{app:candidate-pool-analysis}

We construct \ds{} in two stages. First, an automatic preprocessing pipeline collects and cleans multiple-choice questions from Pakistani examination boards and Urdu MCQ websites to produce a candidate pool. Second, annotation, verification, deduplication, and balancing transform this pool into the final benchmark used in all evaluations. This appendix analyzes both stages and shows how the dataset composition changes throughout the construction process.

\begin{figure*}[t]
    \centering
    \includegraphics[width=\linewidth]{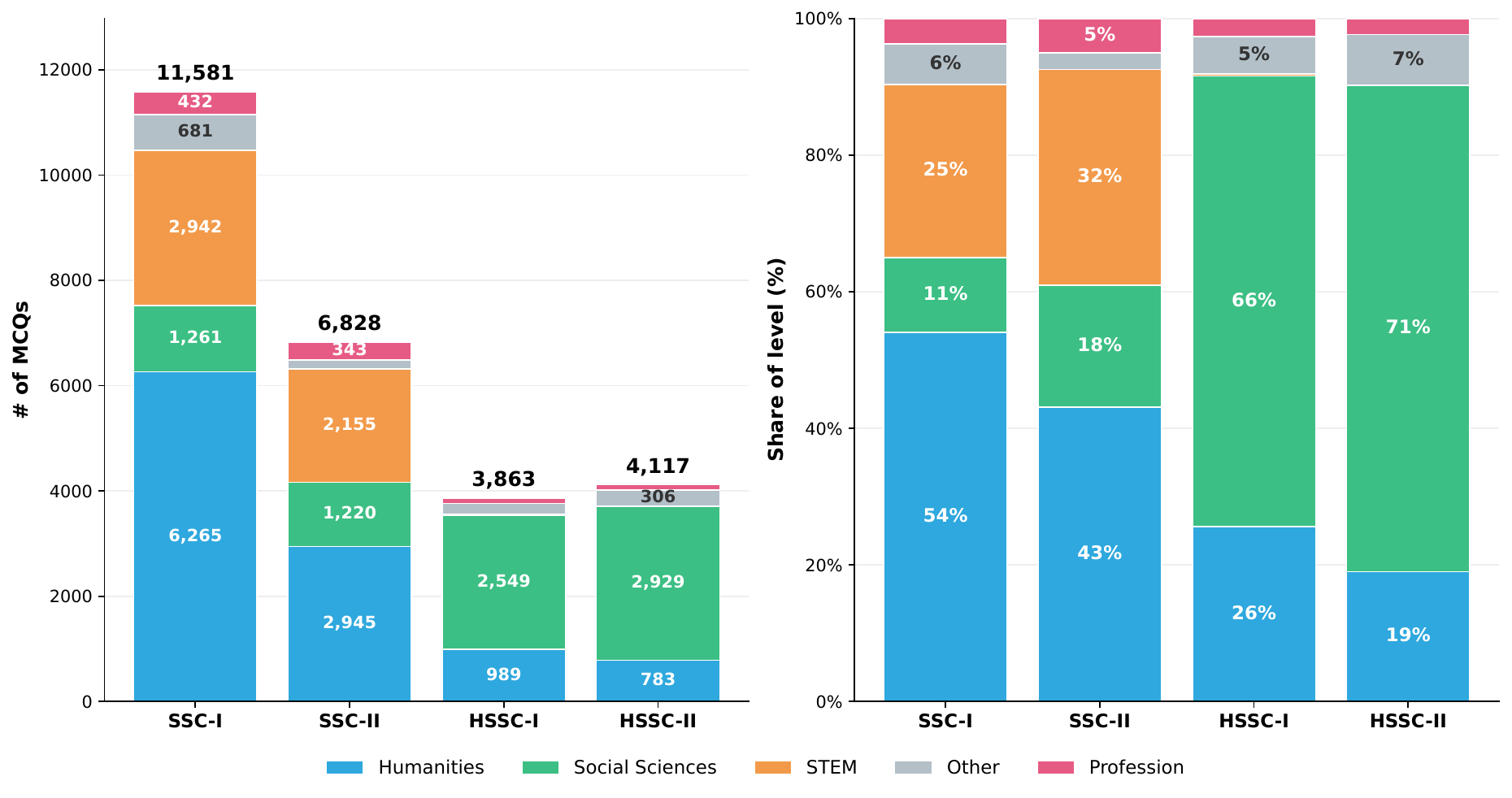}
    \caption{\textbf{Distribution of \ds{} items across Pakistani examination levels by domain.} \textbf{Left:} absolute item counts per level. \textbf{Right:} within-level domain distribution. Humanities dominates SSC levels, Social Sciences dominates HSSC levels, and STEM is concentrated at SSC levels.}
    \label{fig:levels-stacked}
\end{figure*}

Figure~\ref{fig:levels-stacked} shows the distribution of \ds{} items across the four Pakistani examination levels: SSC-I, SSC-II, HSSC-I, and HSSC-II. The left panel reports absolute item counts, while the right panel reports the within-level domain distribution. Figure~\ref{fig:levels-stacked} highlights two trends. First, Humanities dominates the SSC levels, at $54.1\%$ of SSC-I and $43.1\%$ of SSC-II, where language and literature subjects occupy a larger portion of the curriculum. Second, Social Sciences dominates the HSSC levels, at $66.0\%$ of HSSC-I and $71.1\%$ of HSSC-II, where students specialize into commerce and humanities tracks. STEM does not follow this pattern: our examination sources supply STEM papers almost exclusively at the SSC levels, so only 16 of the 5{,}113 STEM items carry an HSSC label. The level distribution therefore reflects both the structure of the Pakistani curriculum and the availability of examination material by level.

\begin{figure*}[!h]
    \centering

    \begin{subfigure}{0.48\linewidth}
        \centering
        \includegraphics[width=\linewidth]{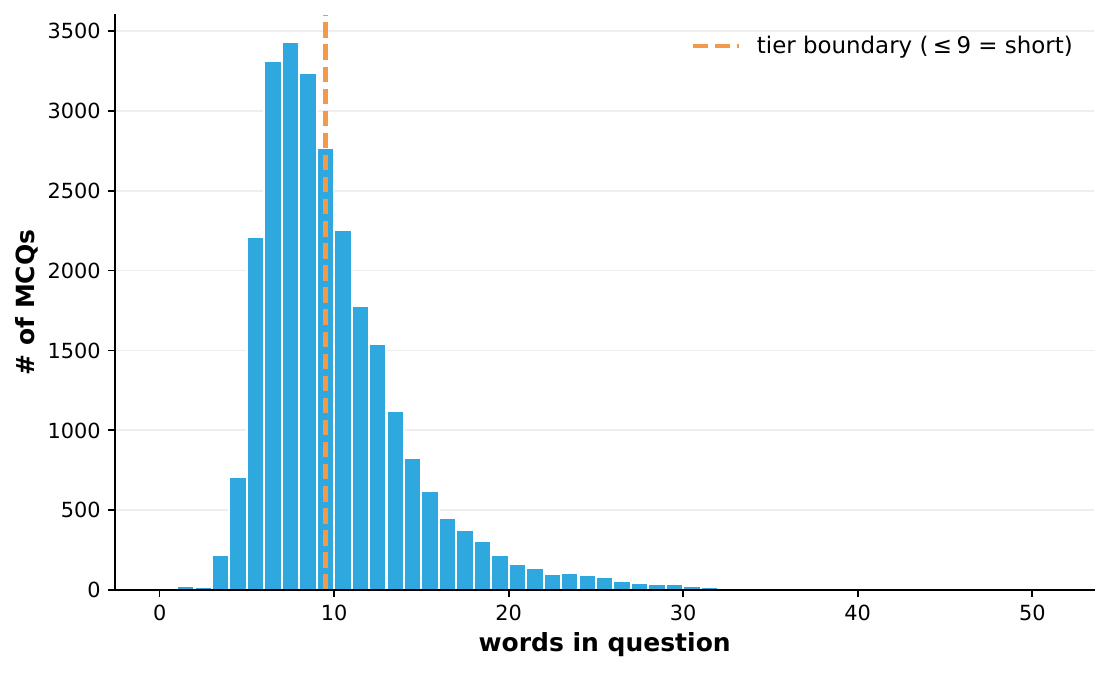}
        \caption{Distribution of question-stem lengths across all released items. The dashed line marks the nine-word boundary.}
        \label{fig:length-hist}
    \end{subfigure}
    \hfill
    \begin{subfigure}{0.48\linewidth}
        \centering
        \includegraphics[width=\linewidth]{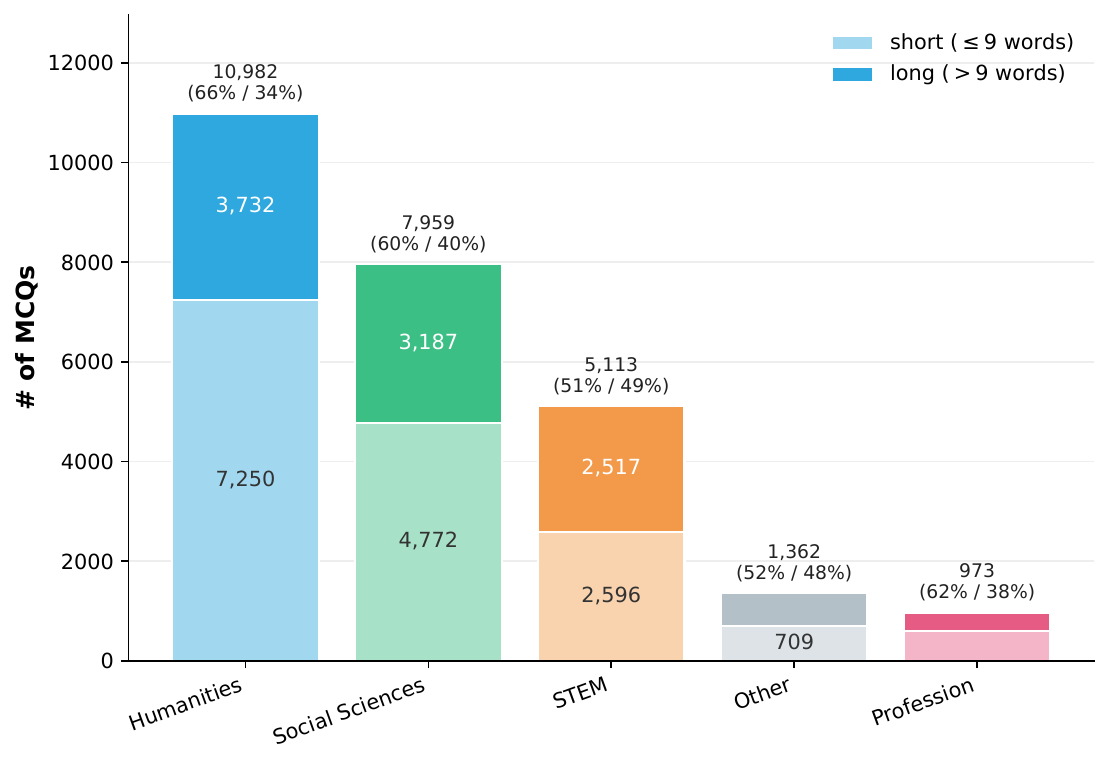}
        \caption{Domain-level counts of short and long stems, with within-domain shares in parentheses.}
        \label{fig:length-tier-bar}
    \end{subfigure}

    \caption{\textbf{Question-length analysis for \ds{}.} Word counts use normalized stems from all $26{,}389$ items. Short stems contain at most nine words. This split is independent of exam-only tier metadata. STEM is nearly balanced, while Humanities and Profession skew shorter.}
    
    \label{fig:length-tier}
\end{figure*}

We also analyze question-stem length because it can influence model performance and vary across domains. For Figure~\ref{fig:length-tier}, we compute word counts directly from the normalized question field for all $26{,}389$ released items, including $12{,}759$ exam-derived and $13{,}630$ web-derived questions. We classify stems with at most nine words as short and those with more than nine words as long. This analytical classification is separate from the tier labels assigned during exam extraction, which are available only for exam-derived items and are not used in this figure. Most stems are short, while STEM has the most balanced split and Humanities and Profession contain larger shares of short stems.

\subsection{Source and Domain Distributions}
\label{app:source-domain-distributions}

Tables~\ref{tab:appendix-source-analysis} and~\ref{tab:appendix-domain-analysis} compare the cleaned candidate pool (\textbf{Raw}) and released benchmark (\textbf{Final}), distinguishing the initial collection distribution from the curated evaluation benchmark.

\begin{table}[t]
\centering
\small
\setlength{\tabcolsep}{4pt}
\renewcommand{\arraystretch}{1.05}
\setlength{\distbarwidth}{1.6cm}
\begin{tabularx}{\linewidth}{lrrrX}
\toprule
\textbf{Source} & \textbf{Raw} & \textbf{Final} & \textbf{\%} & \textbf{Share} \\
\midrule
\distrowtwo{Ustad 360}{23,668}{11,068}{41.9}
\distrowtwo{MCQ Times}{6,099}{5,918}{22.4}
\distrowtwo{TestPoint PK}{3,624}{3,484}{13.2}
\distrowtwo{eTest}{3,096}{2,783}{10.5}
\distrowtwo{FBISE}{2,404}{1,459}{5.5}
\distrowtwo{ExamAunty}{613}{530}{2.0}
\distrowtwo{GoTest}{551}{501}{1.9}
\distrowtwo{PakMCQs}{433}{414}{1.6}
\distrowtwo{BISE Multan 2025}{439}{232}{0.9}
\midrule
\textbf{Total} & \textbf{40,927} & \textbf{26,389} & \textbf{100.0} & \\
\bottomrule
\end{tabularx}
\caption{\textbf{Source distribution of the cleaned candidate pool (Candidate) and the released \ds{} benchmark (Final).} A small number of items are attested by more than one source; each item is attributed to its primary source so that both columns partition their totals. Percentages and share bars correspond to the final benchmark distribution and are rounded independently.}
\label{tab:appendix-source-analysis}
\end{table}

\begin{table}[!h]
\centering
\small
\setlength{\tabcolsep}{4pt}
\renewcommand{\arraystretch}{1.05}
\setlength{\distbarwidth}{1.6cm}
\begin{tabularx}{\linewidth}{lrrrX}
\toprule
\textbf{Domain} & \textbf{Raw} & \textbf{Final} & \textbf{\%} & \textbf{Share} \\
\midrule
\distrowtwo{Humanities}{11,678}{10,982}{41.6}
\distrowtwo{Social Sciences}{14,985}{7,959}{30.2}
\distrowtwo{STEM}{11,592}{5,113}{19.4}
\distrowtwo{Other}{1,460}{1,362}{5.2}
\distrowtwo{Profession}{1,212}{973}{3.7}
\midrule
\textbf{Total} & \textbf{40,927} & \textbf{26,389} & \textbf{100.0} & \\
\bottomrule
\end{tabularx}
\caption{\textbf{Domain distribution in the cleaned candidate pool (Raw) and released \ds{} benchmark}. Both use the final taxonomy, including 570 items reassigned from Other to Profession. Rounded shares sum to $100.0$.}
\label{tab:appendix-domain-analysis}

\end{table}

\paragraph{Source distribution:} The cleaned candidate pool contains 40{,}927 items collected from nine Pakistani examination and MCQ-bank sources, of which 26{,}389 survive into the final benchmark. Table~\ref{tab:appendix-source-analysis} shows that the candidate pool is heavily concentrated in a few large sources. Ustad~360 alone contributes 23{,}668 candidate items, while the four largest sources together account for nearly 90\% of the pool. The final benchmark is substantially less skewed. Annotation, deduplication, and balancing reduce the relative share of the largest sources, while smaller sources such as MCQ~Times, TestPoint~PK, and eTest contribute proportionally more to the released benchmark. Ustad~360 shows the largest proportional reduction, retaining $46.8\%$ of its candidate items, followed by BISE Multan~2025 at $52.8\%$ and FBISE at $60.7\%$; in each case the losses come mainly from the balancing step and from duplicates shared with other examination sources.

Table~\ref{tab:construction-ledger} traces the exam-derived and web-derived halves separately from the candidate pool through to the release, with every removal itemised, so that each subtotal can be checked against the rows above it. Two steps account for most of the reduction. First, balancing withholds 8{,}946 STEM and Social Sciences items from annotation entirely, which is what drives the domain shift in Table~\ref{tab:appendix-domain-analysis}. Second, near-duplicate filtering removes a further 2{,}356 items after the two halves are merged. The remaining removals are comparatively small, and follow stated rules.

\begin{table}[!h]
\centering
\small
\setlength{\tabcolsep}{4pt}
\renewcommand{\arraystretch}{1.05}
\begin{tabularx}{\linewidth}{Xrrr}
\toprule
\textbf{Stage} & \textbf{Exam} & \textbf{Web} & \textbf{Total} \\
\midrule
Cleaned candidate pool & 26{,}511 & 14{,}416 & 40{,}927 \\
\quad withheld from annotation (balancing) & $-8{,}946$ & --- & $-8{,}946$ \\
Entered answer acquisition & 17{,}565 & 14{,}416 & 31{,}981 \\
\midrule
\quad flagged as problematic & $-1{,}247$ & --- & $-1{,}247$ \\
\quad marked unsure & $-243$ & --- & $-243$ \\
\quad single annotation only & $-5$ & --- & $-5$ \\
\quad annotators selected different answers & $-1{,}611$ & --- & $-1{,}611$ \\
\quad no valid answer key & --- & $-41$ & $-41$ \\
\quad duplicate question stem & --- & $-19$ & $-19$ \\
Gold labels established & 14{,}459 & 14{,}356 & 28{,}815 \\
\midrule
\quad exam item matching a web item, counted once & $-28$ & --- & $-28$ \\
Combined pool & 14{,}431 & 14{,}356 & 28{,}787 \\
\quad near-duplicate filtering (Jaccard $\geq 0.85$) & $-1{,}672$ & $-684$ & $-2{,}356$ \\
Near-duplicate filtered & 12{,}759 & 13{,}672 & 26{,}431 \\
\quad gold answer outside A--D & --- & $-42$ & $-42$ \\
\midrule
\textbf{Released benchmark} & \textbf{12{,}759} & \textbf{13{,}630} & \textbf{26{,}389} \\
\bottomrule
\end{tabularx}

\caption{\textbf{Construction ledger from the cleaned candidate pool to release, with each column reconciling independently.} Exam items lack source keys and are dual-annotated; flags, unsure labels, single annotations, and disagreement are mutually exclusive exclusions applied in the listed order. Web items use published keys, with missing keys and duplicate stems removed before merging. Near-duplicate filtering forms 27{,}251 clusters, removes 1{,}536 duplicates and 820 answer-conflicting clusters, leaving 26{,}431 items. We then remove 42 web items with keys outside A--D.}

\label{tab:construction-ledger}
\end{table} 

\begin{figure}[!h]
    \centering
    \includegraphics[width=\linewidth]{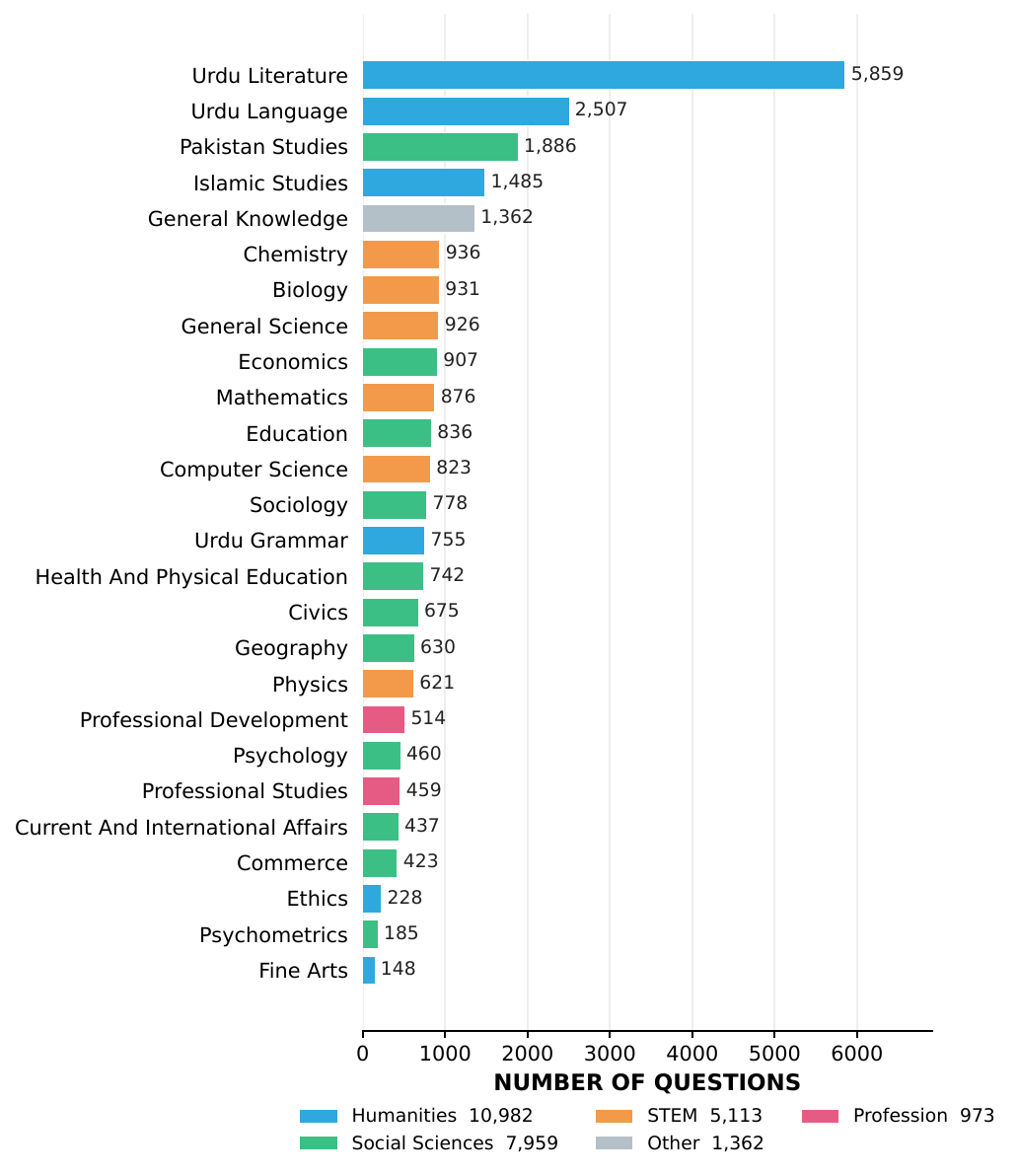}
    \caption{\textbf{Final \ds{} item counts by subdomain, grouped by domain.} Urdu Literature and Urdu Language contribute the largest shares, while Social Sciences and STEM distribute across a larger number of medium-sized subdomains.}
    \label{fig:subject-bars}
\end{figure}

Figure~\ref{fig:subject-bars} expands the statistics from Table~\ref{tab:appendix-domain-analysis} to subdomains. Humanities is dominated by Urdu Literature and Urdu Language, whereas Social Sciences and STEM distribute more evenly across multiple medium-sized subdomains.

\paragraph{Domain distribution:}
Table~\ref{tab:appendix-domain-analysis} compares the cleaned candidate pool and the final benchmark across domains. The candidate pool distributes relatively evenly across Humanities, Social Sciences, and STEM, while Other and Profession remain much smaller. 

The final benchmark shifts toward Humanities, which grows from $28.5\%$ to $41.6\%$, while Social Sciences falls from $36.6\%$ to $30.2\%$ and STEM from $28.3\%$ to $19.4\%$. Every domain decreases in absolute terms when both columns use the final taxonomy. We note this because domain labels changed during normalization: 570 items carrying the \texttt{professional development} and \texttt{federal investigation law} subdomains were reassigned from Other to Profession. Table~\ref{tab:appendix-domain-analysis} applies the final domain taxonomy consistently to both the candidate and released benchmark counts. These changes reflect a deliberate balancing step rather than artifacts of preprocessing or cleaning. We withhold 8{,}946 overrepresented STEM and Social Sciences items from annotation to better reflect the Pakistani SSC and HSSC curriculum shown in Figure~\ref{fig:levels-stacked}. This process improves domain coverage while maintaining curriculum alignment, resulting in a benchmark providing balanced representation of subjects encountered in Pakistani education.

\section{Annotation Details}
\label{app:annotation-details}

The exam-derived portion of \ds{} came from Pakistani examination boards and MCQ sources that did not provide answer keys. To produce reliable gold labels, we recruited 17 Urdu-fluent annotators and ran a dual-annotator consensus process supported by a custom dashboard and written guidelines. This appendix documents the annotator pool, the annotation guidelines and dashboard, the inclusion and edit-resolution rules, and the resulting agreement statistics.

\subsection{Annotator Demographics and Feedback}
\label{app:annotator-demographics}

The annotation pool consisted of 17 annotators recruited for native Urdu fluency and familiarity with the Pakistani school curriculum. Table~\ref{tab:annotator-demographics} summarizes the demographic profile of the pool. The annotators were approximately gender-balanced (52.9\% female, 47.1\% male), predominantly native Urdu speakers (94.1\%), and concentrated in the 18--34 age range. Educationally, 88.2\% held at least a bachelor's degree and 41.2\% held a master's degree, which is important for a benchmark that targets SSC- and HSSC-level subject content. Most annotators also reported between one and six years of professional experience.

Recruitment focused on Urdu language ability and familiarity with the educational context covered by the benchmark. The final pool included 16 native Urdu speakers and one non-native speaker who was fluent in Urdu. All annotators were between 18 and 34 years old. Fifteen annotators had completed at least a bachelor's degree, while the remaining two had completed high school or some college or vocational education. The pool also represented different levels of professional experience, ranging from less than one year to nine years. Gender representation was nearly balanced, with nine women and eight men. After completing their assigned batches, all annotators filled out a post-task satisfaction survey. Table~\ref{tab:annotator-satisfaction} summarizes the responses. Feedback was consistently positive, and no annotator selected \emph{Disagree} or \emph{Strongly disagree} for any statement. Instruction clarity, compensation fairness, guideline usefulness, and overall satisfaction received entirely positive responses. Task enjoyment received a smaller number of neutral responses (23.5\%), suggesting that annotators found the process clear and manageable even if not inherently engaging. Annotators were compensated for completing their assigned annotation batches. 

\begin{table}[!t]
\centering
\small
\renewcommand{\arraystretch}{1}

\begin{tabular*}{\columnwidth}{@{\extracolsep{\fill}}lrr}
\toprule
\multirow{2}{*}{\textbf{Attribute}} &
\multicolumn{2}{c}{\textbf{Annotators}} \\
\cmidrule(lr){2-3}
& \textbf{Count}
& \textbf{Share (\%)} \\
\midrule

\rowcolor{softgray}
\multicolumn{3}{c}{\textbf{Gender}} \\
\midrule

Female
& 9
& 52.9 \\

Male
& 8
& 47.1 \\

\midrule
\rowcolor{softgray}
\multicolumn{3}{c}{\textbf{Native Urdu Speaker}} \\
\midrule

Yes
& 16
& 94.1 \\

No
& 1
& 5.9 \\

\midrule
\rowcolor{softgray}
\multicolumn{3}{c}{\textbf{Age Range}} \\
\midrule

18--24
& 7
& 41.2 \\

25--34
& 10
& 58.8 \\

\midrule
\rowcolor{softgray}
\multicolumn{3}{c}{\textbf{Highest Completed Education}} \\
\midrule

High school diploma
& 1
& 5.9 \\

Some college / vocational
& 1
& 5.9 \\

Bachelor's degree
& 8
& 47.1 \\

Master's degree
& 7
& 41.2 \\

\midrule
\rowcolor{softgray}
\multicolumn{3}{c}{\textbf{Professional Work Experience}} \\
\midrule

Less than 1 year
& 5
& 29.4 \\

1--3 years
& 6
& 35.3 \\

4--6 years
& 4
& 23.5 \\

7--9 years
& 2
& 11.8 \\

\midrule
\textbf{Total Annotators}
& \textbf{17}
& \textbf{100.0} \\

\bottomrule
\end{tabular*}

\caption{\textbf{Annotator demographics.} Profile of the anonymised UrduMMLU annotator pool ($n=17$).}
\label{tab:annotator-demographics}
\end{table}

\begin{table}[!h]
\centering
\small
\setlength{\tabcolsep}{3.5pt}
\renewcommand{\arraystretch}{1}

\begin{tabularx}{\columnwidth}{@{}>{\raggedright\arraybackslash}Xrrrr@{}}
\toprule
\multirow{2}{*}{\textbf{Statement}} &
\multicolumn{4}{c}{\textbf{Response Share (\%)}} \\
\cmidrule(lr){2-5}
& \textbf{SA}
& \textbf{A}
& \textbf{N}
& \textbf{Pos.} \\
\midrule

Instructions clearly explained the task
& 82.4
& 17.6
& 0.0
& 100.0 \\

Task was enjoyable and engaging
& 47.1
& 29.4
& 23.5
& 76.5 \\

Paid fairly for the task
& 76.5
& 23.5
& 0.0
& 100.0 \\

Examples / guidelines were helpful
& 76.5
& 23.5
& 0.0
& 100.0 \\

Overall glad to have completed the task
& 76.5
& 23.5
& 0.0
& 100.0 \\

\bottomrule
\end{tabularx}

\caption{\textbf{Annotator satisfaction survey ($n=17$).} SA, A, and N denote strongly agree, agree, and neutral; Pos.\ combines SA and A. Disagree responses are omitted because none were recorded.}
\label{tab:annotator-satisfaction}
\end{table}

In the post-task survey, 13 annotators strongly agreed that compensation was fair, and the remaining four agreed. No neutral or negative responses were recorded.

\subsection{Annotation Guidelines}
\label{app:annotation-guidelines}

Before annotation began, we held a live online onboarding session in which we walked through the task end-to-end, demonstrated each flag and edit category on real items, and answered annotator questions. Written guidelines remained available in the annotation dashboard, and administrators could be contacted by email for unresolved cases. Annotators were encouraged to review the guidelines when uncertain to maintain consistency across batches. These procedures helped ensure consistent annotation decisions.

\paragraph{Task overview:} Annotators verified answers by selecting the \emph{single best} option from \texttt{A}/\texttt{B}/\texttt{C}/\texttt{D}. When multiple options looked plausible, they selected the most precise or relevant answer rather than guessing.

\paragraph{Look-up and abstention policy:} Annotators were encouraged to consult Google or Wikipedia for fact-based questions (dates, authors, capitals, scientific terms, historical events) rather than relying on memory, with a target pace of 15--30 seconds per question including verification. Annotators were asked to mark an item as \emph{unsure / skip} rather than submit a confident guess in any of the following cases: (i) the answer could not be resolved within roughly a minute of search, (ii) two or more options remained equally plausible after verification, or (iii) the question required specialist context (e.g., niche fiqh details or obscure regional history) that they could not quickly acquire.

\paragraph{Flag vs.\ edit:} Annotators were given a single rule of thumb to choose between the two actions: \emph{edit} when the issue could be fixed in-place by changing text (a typo, missing space, duplicated word, wrong subdomain label), and \emph{flag} when the issue required admin review and could not be repaired by text correction (no correct answer, multiple correct answers, ambiguity, missing visual, out-of-scope content). 

Annotators were explicitly instructed not to rewrite question semantics, not to ``fix'' wrong distractors into correct ones, and not to flag a question solely because they had edited a typo in it. 

\begin{figure*}[t]
    \centering
    \begin{subfigure}[t]{0.48\linewidth}
        \centering
        \includegraphics[width=\linewidth]{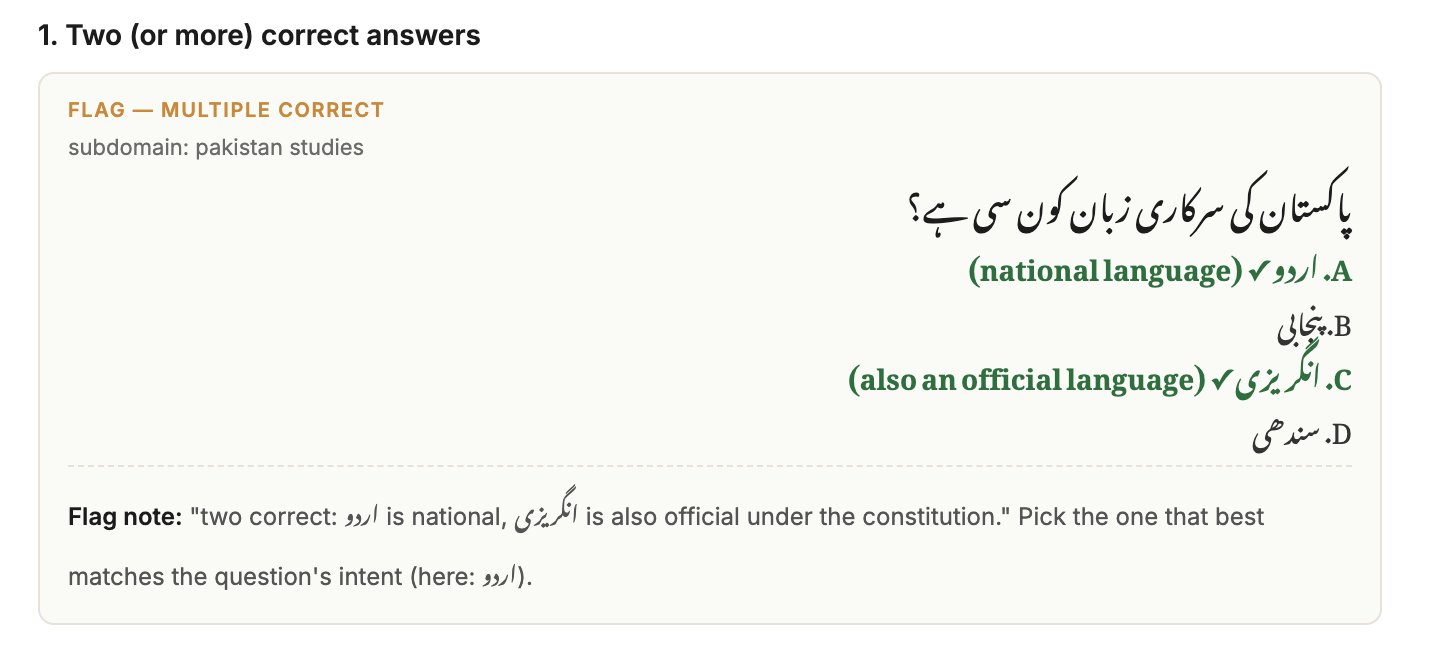}
        \caption{Two or more options are simultaneously correct.}
        \label{fig:guidelines-flag-multiple-correct}
    \end{subfigure}
    \hfill
    \begin{subfigure}[t]{0.48\linewidth}
        \centering
        \includegraphics[width=\linewidth]{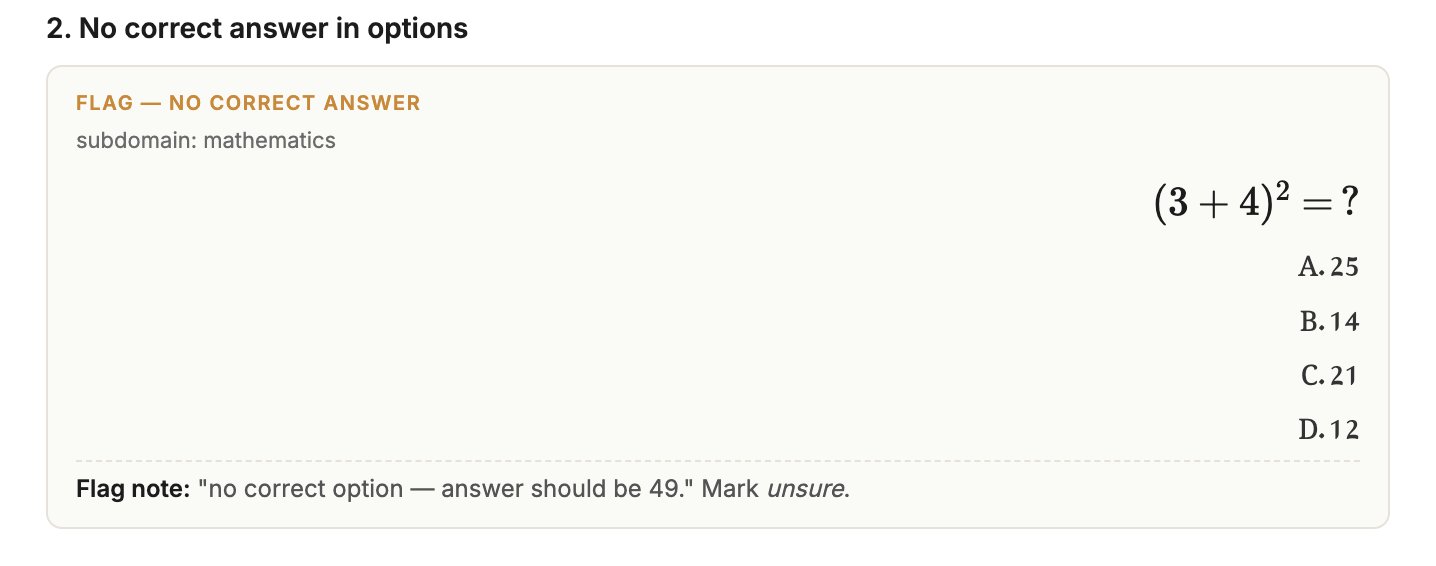}
        \caption{No option in the list is the correct answer.}
        \label{fig:guidelines-flag-no-correct}
    \end{subfigure}

    \vspace{0.6em}

    \begin{subfigure}[t]{0.48\linewidth}
        \centering
        \includegraphics[width=\linewidth]{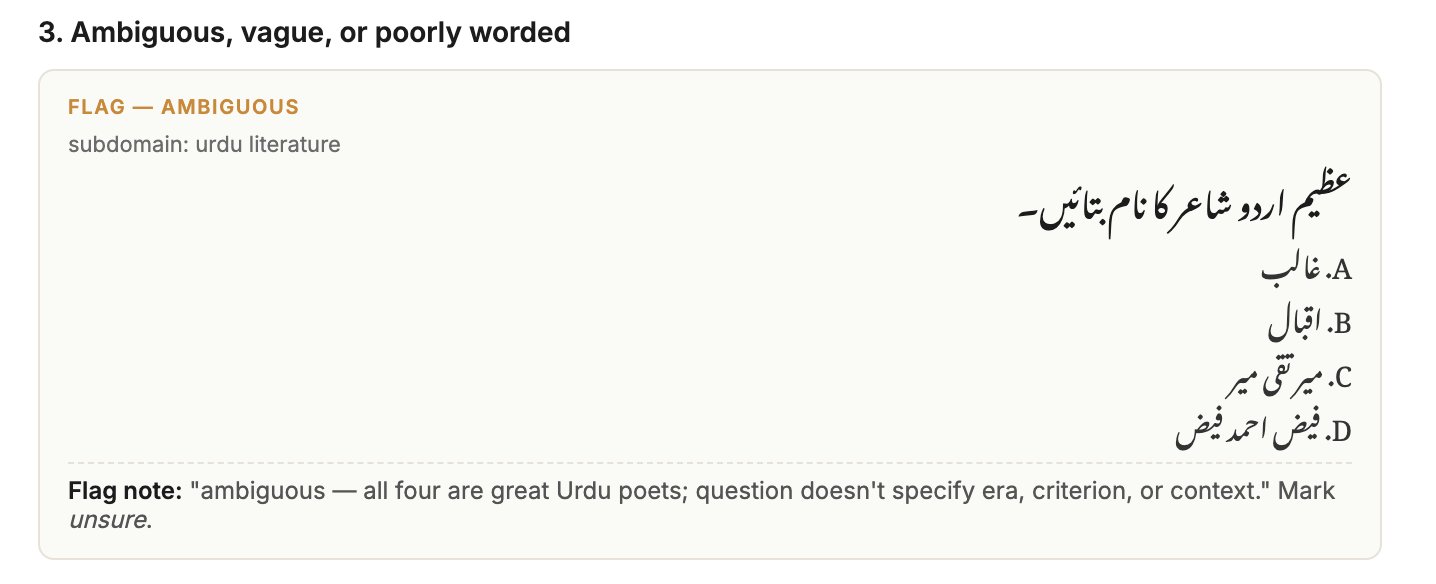}
        \caption{Question is ambiguous, vague, or under-specified.}
        \label{fig:guidelines-flag-ambiguous}
    \end{subfigure}
    \hfill
    \begin{subfigure}[t]{0.48\linewidth}
        \centering
        \includegraphics[width=\linewidth]{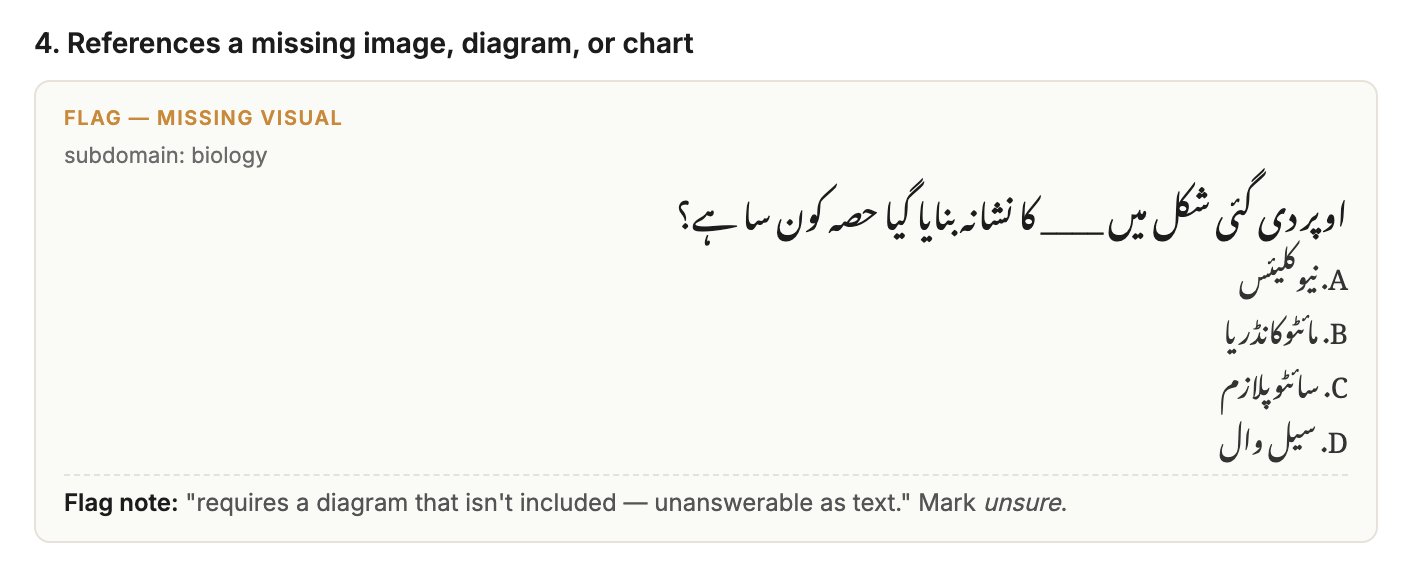}
        \caption{Question references a missing image, diagram, or chart.}
        \label{fig:guidelines-flag-missing-visual}
    \end{subfigure}

    \vspace{0.6em}

    \begin{subfigure}[t]{0.48\linewidth}
        \centering
        \includegraphics[width=\linewidth]{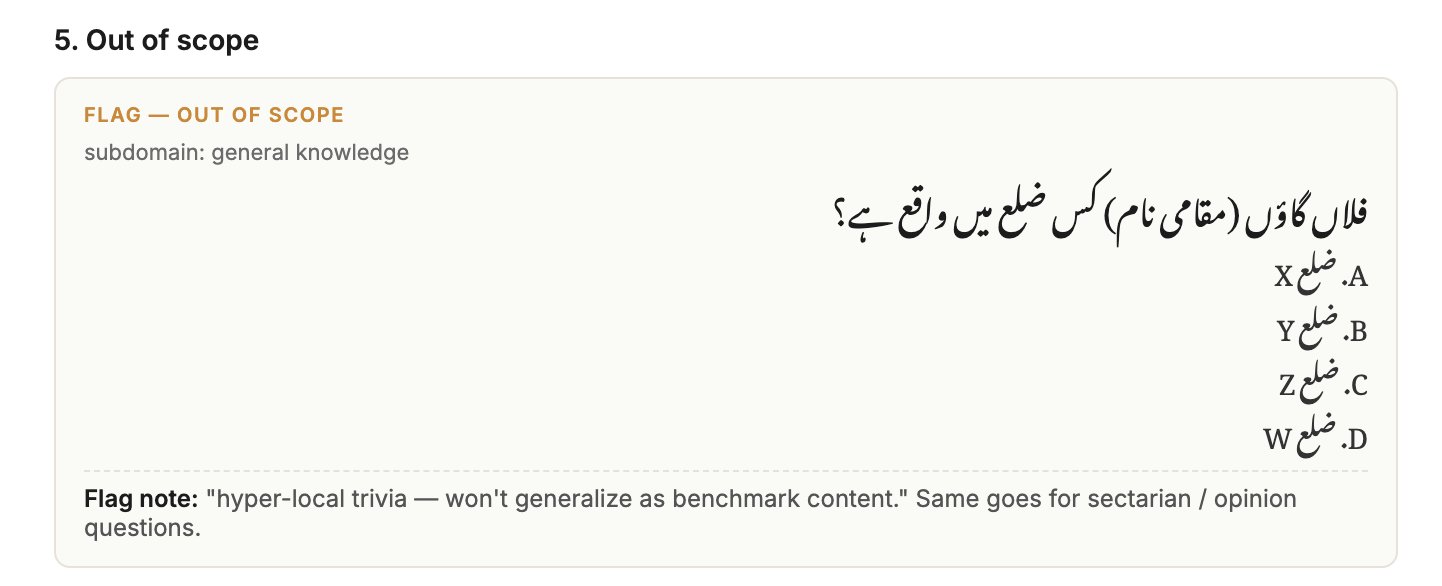}
        \caption{Hyper-local, sectarian, or opinion content that is out of scope for the benchmark.}
        \label{fig:guidelines-flag-out-of-scope}
    \end{subfigure}
    \caption{\textbf{Examples of the five flag categories used in the annotation guidelines.} Annotators were asked to flag the item and attach a short free-text note for each case.}
    \label{fig:guidelines-flag}
\end{figure*}

\paragraph{When to flag:} Figure~\ref{fig:guidelines-flag} illustrates the five flag categories covered in the guidelines. These are: (a) two or more options are simultaneously correct; (b) none of the listed options is the correct answer; (c) the question is ambiguous, vague, or under-specified; (d) the question references a diagram, image, or chart that is not included in the text; and (e) the question is out of scope for the benchmark (hyper-local trivia, sectarian content, opinion questions). For each case, annotators were asked to attach a short free-text note explaining the issue. This information helped reviewers verify and resolve flagged items during quality control. Flagging was independent of answer selection, and annotators could flag with or without picking an option.

\begin{table}[!t]
\centering
\small
\renewcommand{\arraystretch}{1}

\begin{tabularx}{\columnwidth}{@{}
>{\raggedright\arraybackslash}p{0.28\columnwidth}
>{\raggedright\arraybackslash}X
@{}}
\toprule
\rowcolor{softgray}
\multicolumn{2}{c}{\textbf{Annotation Inclusion Rules}} \\
\cmidrule(lr){1-2}
\textbf{Check}
& \textbf{Drop If} \\
\midrule

Dual annotation
& Only one assigned annotator submitted an annotation. \\

Flag
& Either annotator flagged the item. \\

Unsure
& Either annotator selected the unsure/skip option. \\

Valid pick
& Either annotator did not select one of \texttt{A}/\texttt{B}/\texttt{C}/\texttt{D}. \\

Consensus
& The two annotators selected different answer keys. \\

Assignment
& The annotation came from an annotator who was not assigned to that batch. \\

\bottomrule
\end{tabularx}

\caption{\textbf{Annotation inclusion rules.} An item enters the gold-labelled set only if it passes every check.}
\label{tab:annotation-inclusion-rules}
\end{table}

\begin{figure*}[!h]
    \centering
    \begin{subfigure}[t]{0.32\linewidth}
        \centering
        \includegraphics[width=\linewidth]{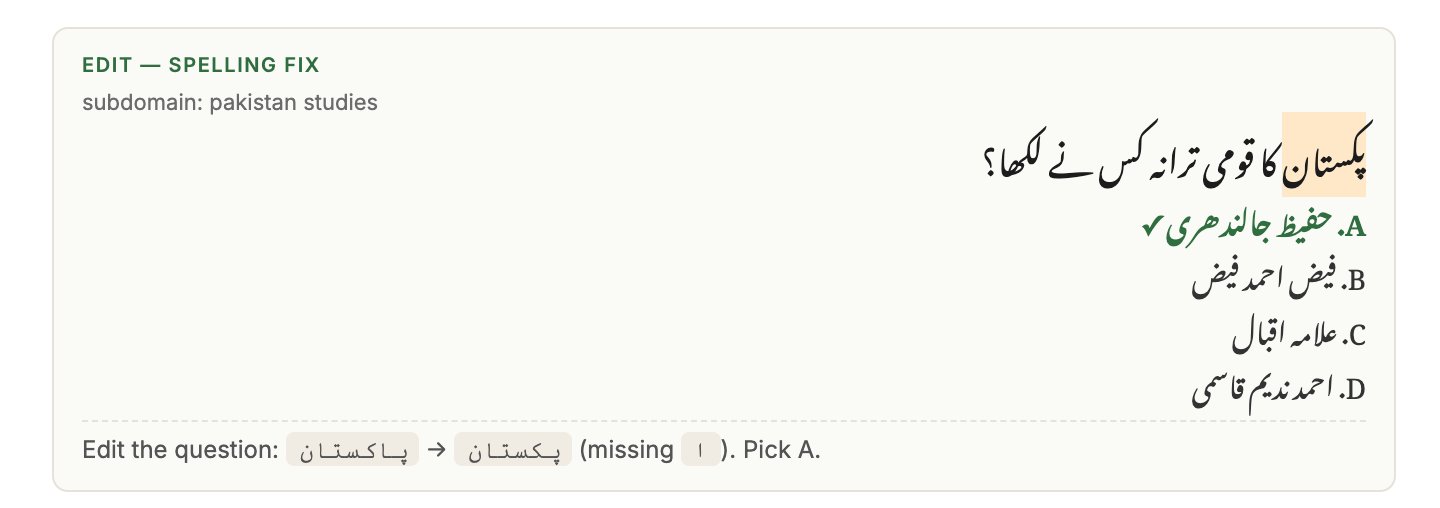}
        \caption{Spelling fix: a missing letter is restored from context.}
        \label{fig:guidelines-edit-spelling}
    \end{subfigure}
    \hfill
    \begin{subfigure}[t]{0.32\linewidth}
        \centering
        \includegraphics[width=\linewidth]{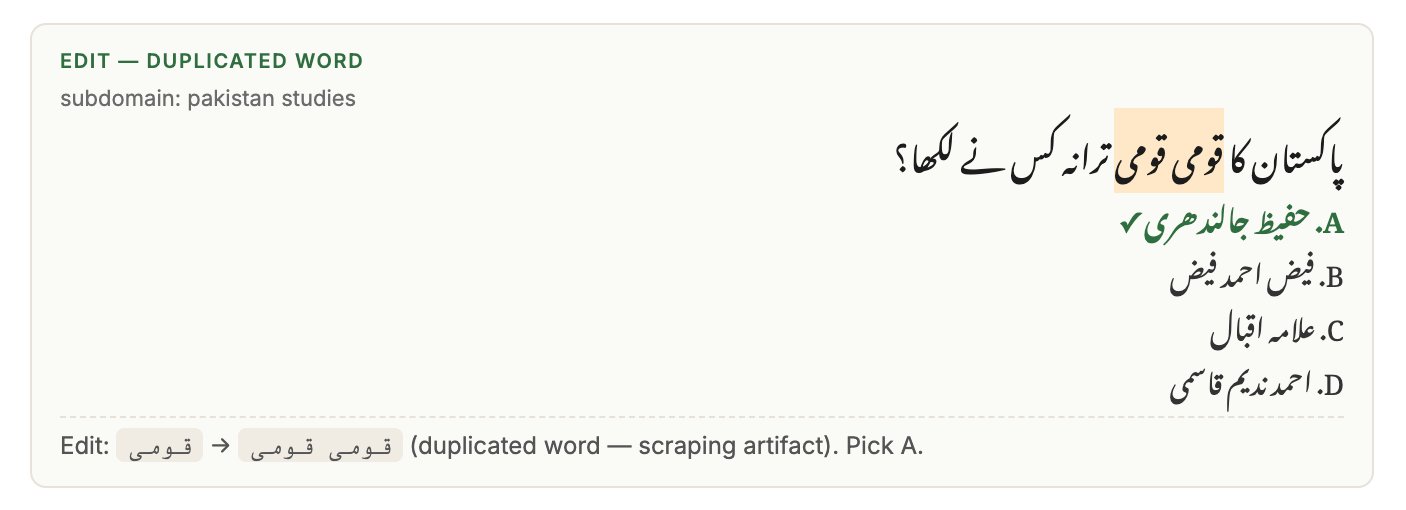}
        \caption{Duplicated word from a scraping artifact is removed.}
        \label{fig:guidelines-edit-duplicated}
    \end{subfigure}
    \hfill
    \begin{subfigure}[t]{0.32\linewidth}
        \centering
        \includegraphics[width=\linewidth]{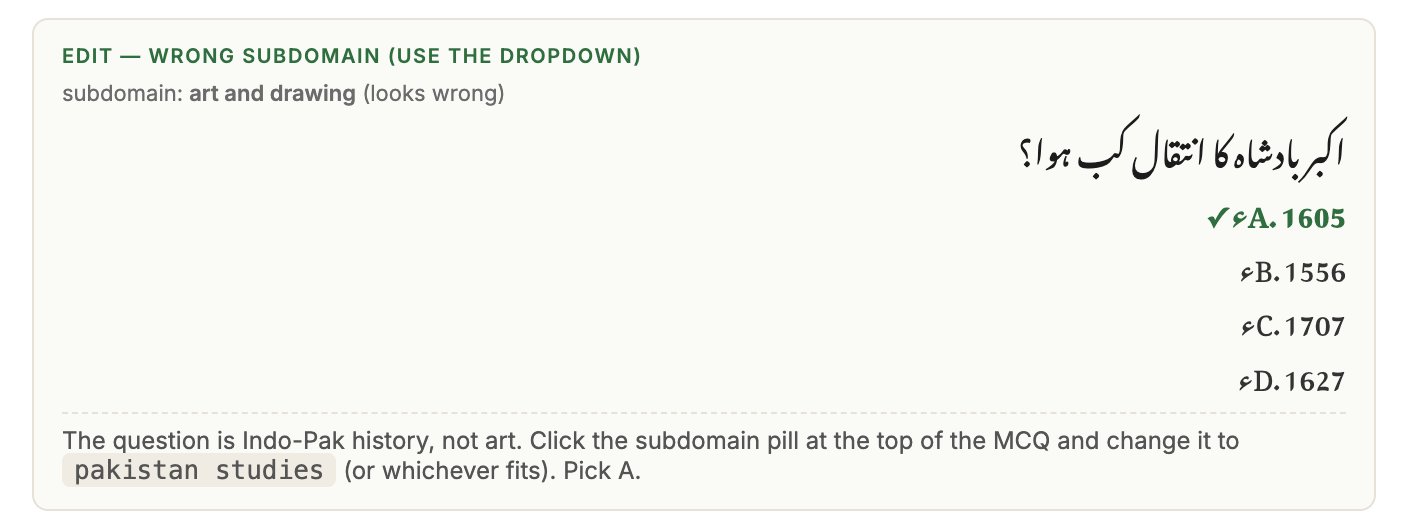}
        \caption{Wrong subdomain (\texttt{art and drawing}) is reassigned via the dropdown to \texttt{pakistan studies}.}
        \label{fig:guidelines-edit-subdomain}
    \end{subfigure}
    \caption{\textbf{Examples of the most common edit categories permitted by the annotation guidelines.} Edits are restricted to OCR, scraping, formatting, and metadata corrections; annotators do not rewrite question semantics or modify answer correctness.}
    \label{fig:guidelines-edit}
\end{figure*}

\paragraph{When to edit:} Figure~\ref{fig:guidelines-edit} illustrates the three most common edit categories: (a) spelling fixes and missing diacritics, where the intended word is clear from context; (b) duplicated words and other scraping artifacts; and (c) subdomain reassignment, where the original subdomain label is clearly inconsistent with the question content. 

Beyond these, the guidelines also permitted spacing corrections, removal of stray punctuation, stripping of redundant in-text option-letter prefixes (e.g., \texttt{A.}, \texttt{B.}, or their Urdu equivalents) already shown by the option badge, and translation of stray English option text when an unambiguous Urdu equivalent existed. Technical English terms, HTML/CSS tags, proper names, and brand names were left in English.

\paragraph{Workflow:} Annotators worked in batches of approximately 50 MCQs, with accuracy prioritized over speed.

\subsection{Inclusion Rules}
\label{app:annotation-inclusion}

We applied a deterministic consensus filter (Table~\ref{tab:annotation-inclusion-rules}): an item was retained only when both assigned annotators independently submitted the same valid answer choice and neither flagged nor abstained. The agreed answer was stored as the final gold label. Two edge cases require clarification. First, when both annotators selected an option such as ``none of these'', we treated it as a valid agreed answer because such options commonly appear in Pakistani MCQ examinations. Second, only the explicit \emph{unsure / skip} action counted as abstention. Missing annotations triggered the incomplete-annotation rule instead, so abstentions always reflected deliberate annotator decisions.

\subsection{Edit Resolution}
\label{app:annotation-edit-resolution}

In addition to selecting answers, annotators could suggest edits to question text, answer options, or subdomain labels. These edits targeted minor extraction and metadata issues such as OCR errors, dropped diacritics, malformed option labels, and incorrect subdomain assignments rather than substantive question rewrites. We resolved all edits deterministically so that the final benchmark could be reconstructed directly from the raw annotations. Table~\ref{tab:annotation-edit-rules} summarizes the resolution rules. 

The resolution policy prioritizes agreed edits when both annotators propose the same correction and otherwise prefers the more conservative or informative revision. For subdomain edits, we recompute the corresponding domain label from the corrected subdomain to preserve consistency between the two fields in the released benchmark. This procedure ensures that metadata corrections remain internally consistent throughout the final dataset.

\begin{table}[!h]
\centering
\small
\renewcommand{\arraystretch}{1.1}

\begin{tabularx}{\columnwidth}{@{}
>{\raggedright\arraybackslash}p{0.38\columnwidth}
>{\raggedright\arraybackslash}X
@{}}
\toprule
\rowcolor{softgray}
\multicolumn{2}{c}{\textbf{Edit-Resolution Rules}} \\
\cmidrule(lr){1-2}
\textbf{Case}
& \textbf{Resolution} \\
\midrule

Only one annotator edited a field
& Use that edit. \\

Both annotators made the same edit
& Use the agreed edit. \\

Both annotators edited the same field differently
& Use the longer edit. \\

No edit was suggested
& Keep the original field. \\

Subdomain was edited
& Recompute the domain from the corrected subdomain. \\

\bottomrule
\end{tabularx}

\caption{\textbf{Edit resolution for annotated MCQs.} Rules are applied per field before evaluating the inclusion criteria in Table~\ref{tab:annotation-inclusion-rules}.}
\label{tab:annotation-edit-rules}
\end{table}

\begin{table}[!h]
\centering
\small
\renewcommand{\arraystretch}{1}

\begin{tabular*}{\columnwidth}{@{\extracolsep{\fill}}lr}
\toprule
\textbf{Outcome}
& \textbf{Count} \\
\midrule

\rowcolor{softgray}
\multicolumn{2}{c}{\textbf{Retention Summary}} \\
\midrule

Input annotated MCQs
& 17,565 \\

Retained after consensus filtering
& 14,459 \\

\midrule
\rowcolor{softgray}
\multicolumn{2}{c}{\textbf{Mutually Exclusive Exclusions}} \\
\midrule

Annotator disagreement
& 1,611 \\

Flagged by annotator
& 1,247 \\

Unsure/skip selected
& 243 \\

Single annotation
& 5 \\

\midrule
\rowcolor{softgray}
\multicolumn{2}{c}{\textbf{Corrections}} \\
\midrule

Subdomain corrections
& 141 \\

\bottomrule
\end{tabular*}

\caption{\textbf{Exam-derived annotation outcomes.} Each excluded item is assigned to one exclusion category.}
\label{tab:annotation-drop-breakdown}
\end{table}

\subsection{Annotation Dashboard}
\label{app:annotation-dashboard}

\begin{figure*}[t]
    \centering
    \begin{subfigure}[t]{0.46\linewidth}
        \centering
        \includegraphics[width=\linewidth]{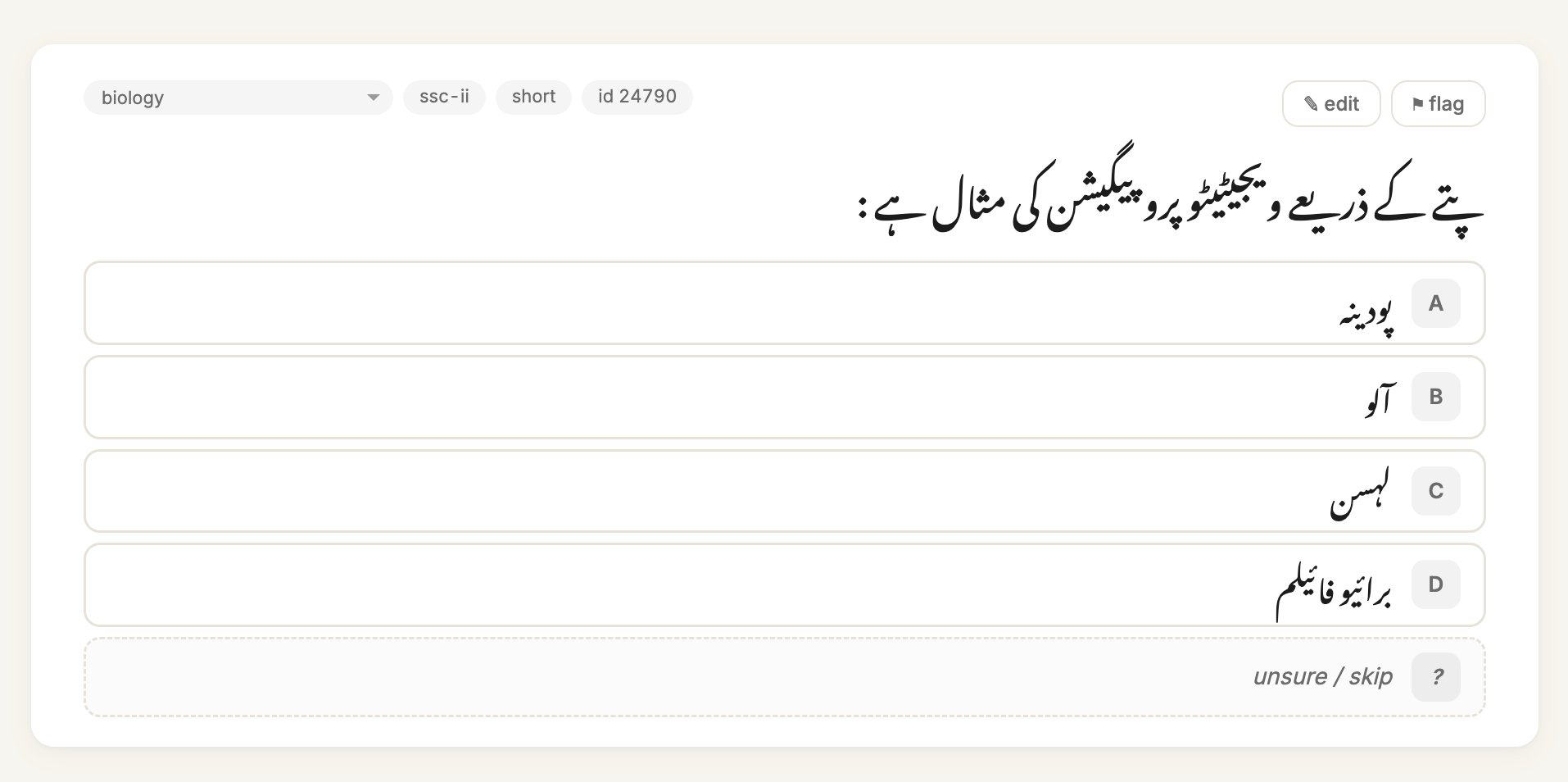}
        \caption{Question view with answer options, metadata, and annotation controls.}
        \label{fig:dash-question-shown}
    \end{subfigure}
    \hfill
    \begin{subfigure}[t]{0.46\linewidth}
        \centering
        \includegraphics[width=\linewidth]{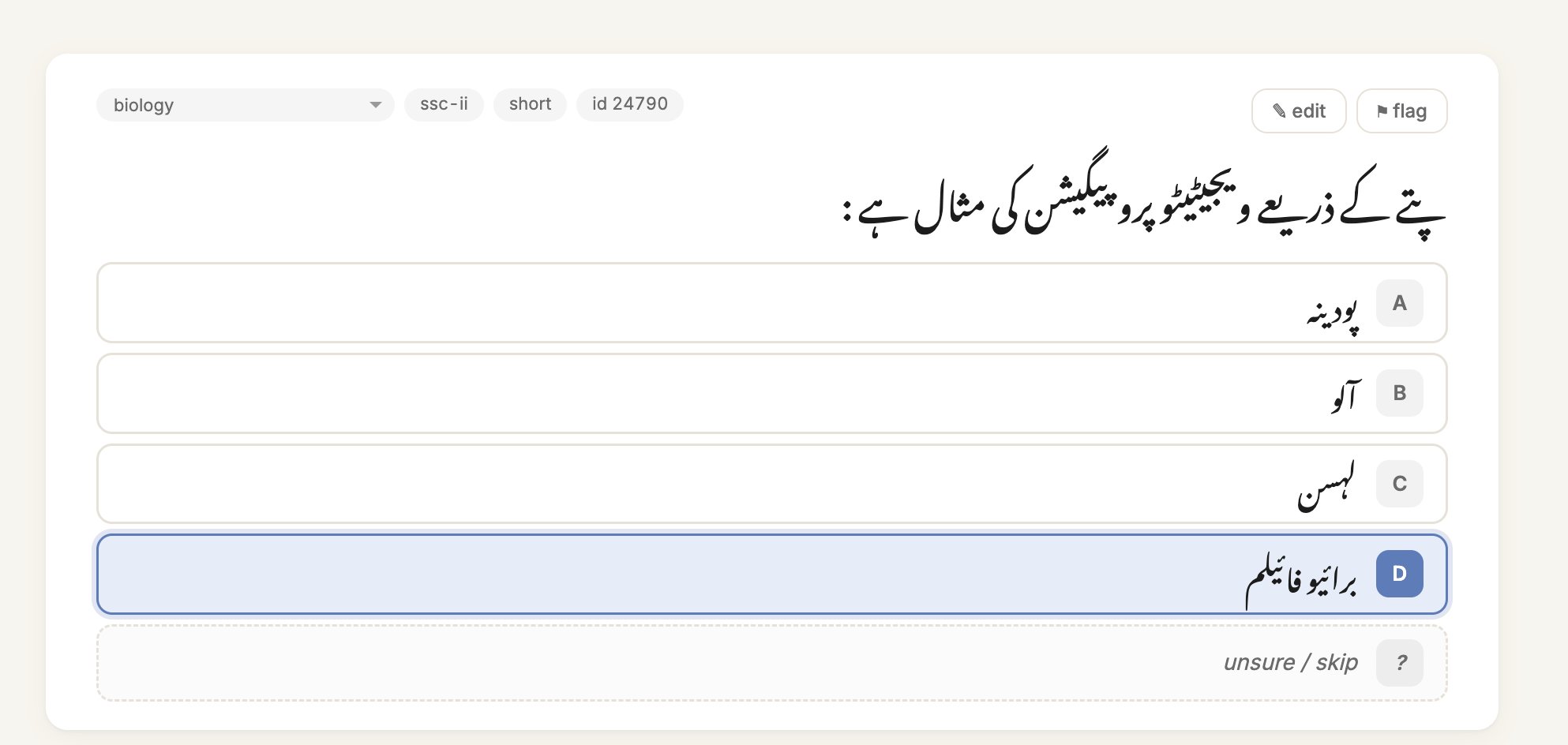}
        \caption{Answer selection interface before advancing to the next item.}
        \label{fig:dash-answer-selected}
    \end{subfigure}
    \caption{\textbf{Annotation dashboard.} Workflow for answer selection.}
    \label{fig:dash-pick}
\end{figure*}

\begin{figure*}[!h]
    \centering
    \begin{subfigure}[t]{0.32\linewidth}
        \centering
        \includegraphics[width=\linewidth]{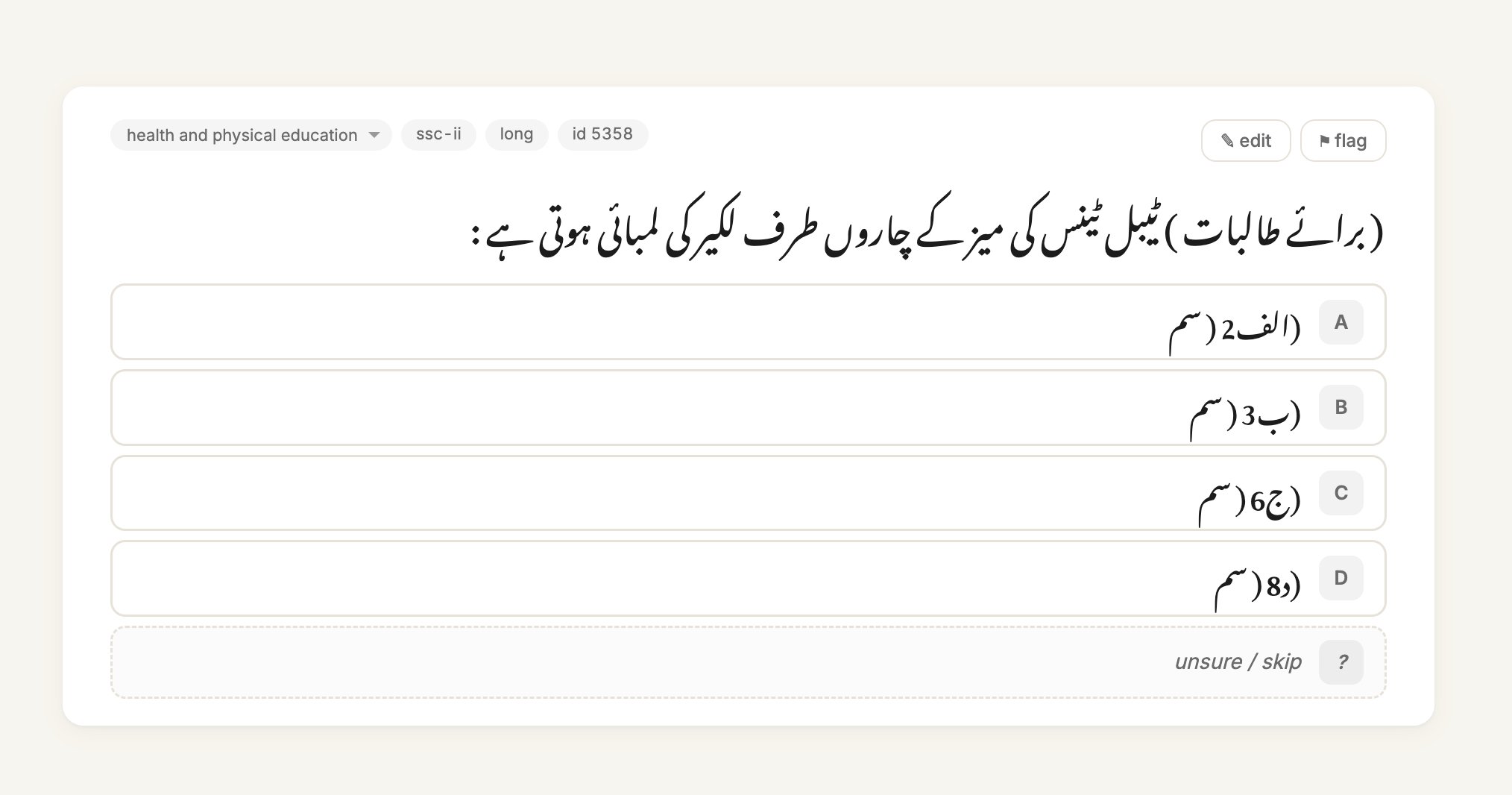}
        \caption{Original OCR-extracted question and options.}
        \label{fig:dash-edit-before}
    \end{subfigure}
    \hfill
    \begin{subfigure}[t]{0.32\linewidth}
        \centering
        \includegraphics[width=\linewidth]{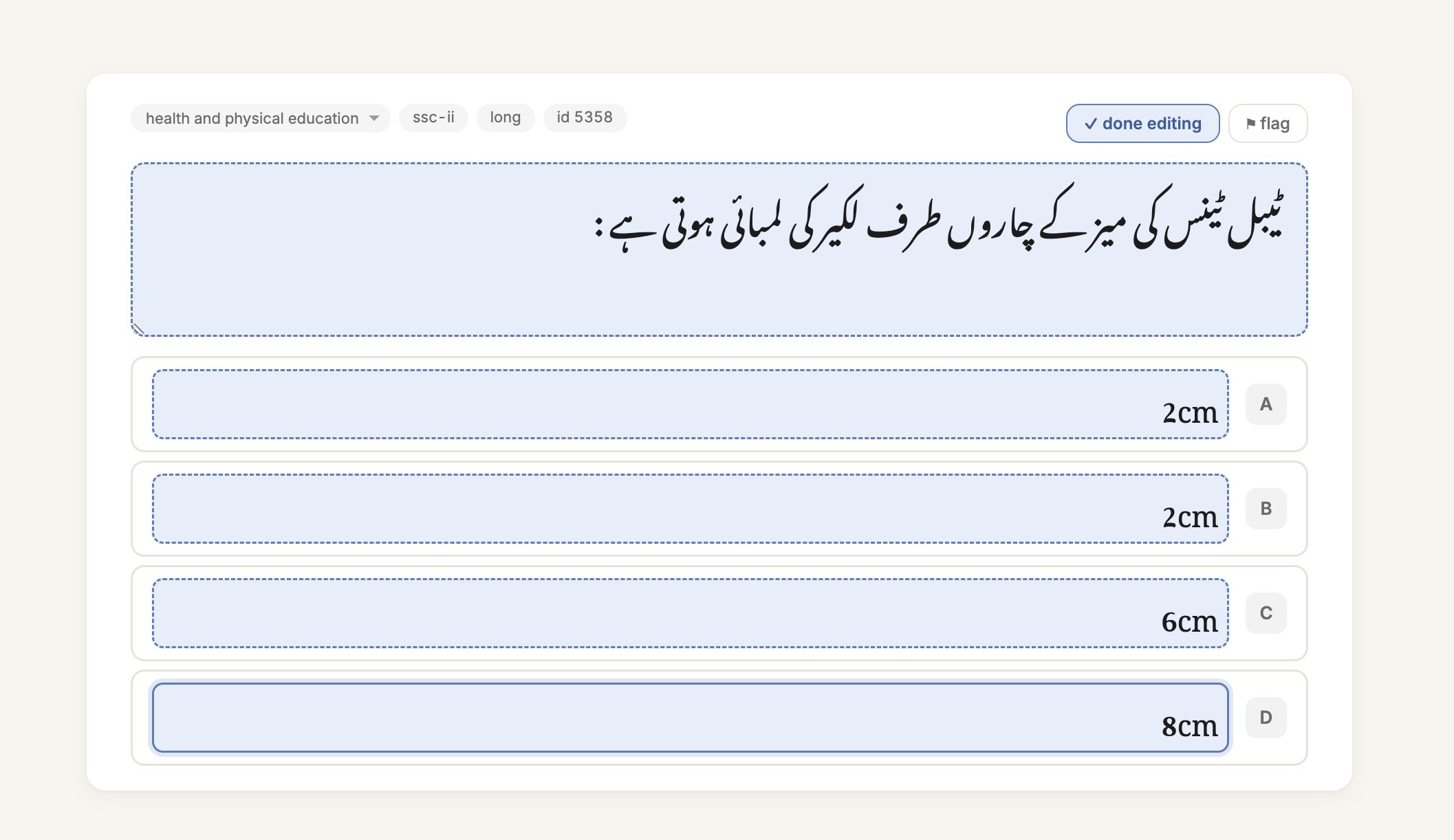}
        \caption{In-place editing of question and option text.}
        \label{fig:dash-edit-in-progress}
    \end{subfigure}
    \hfill
    \begin{subfigure}[t]{0.32\linewidth}
        \centering
        \includegraphics[width=\linewidth]{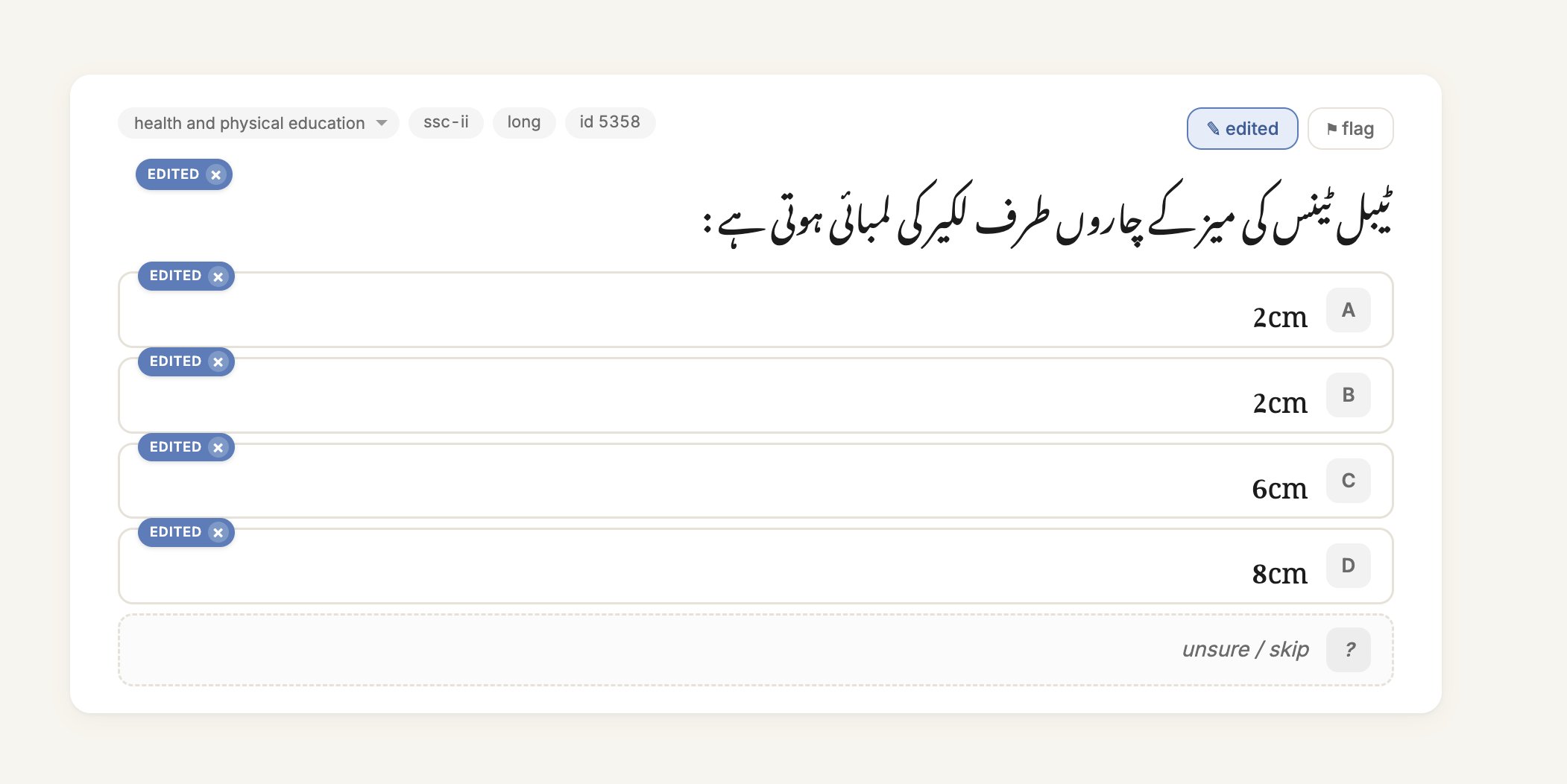}
        \caption{Saved edits with editable revision markers.}
        \label{fig:dash-edit-done}
    \end{subfigure}
    \caption{\textbf{Annotation dashboard.} Workflow for text correction and normalization.}
    \label{fig:dash-edit}
\end{figure*}

\begin{figure*}[!h]
    \centering
    \begin{subfigure}[t]{0.48\linewidth}
        \centering
        \includegraphics[width=\linewidth]{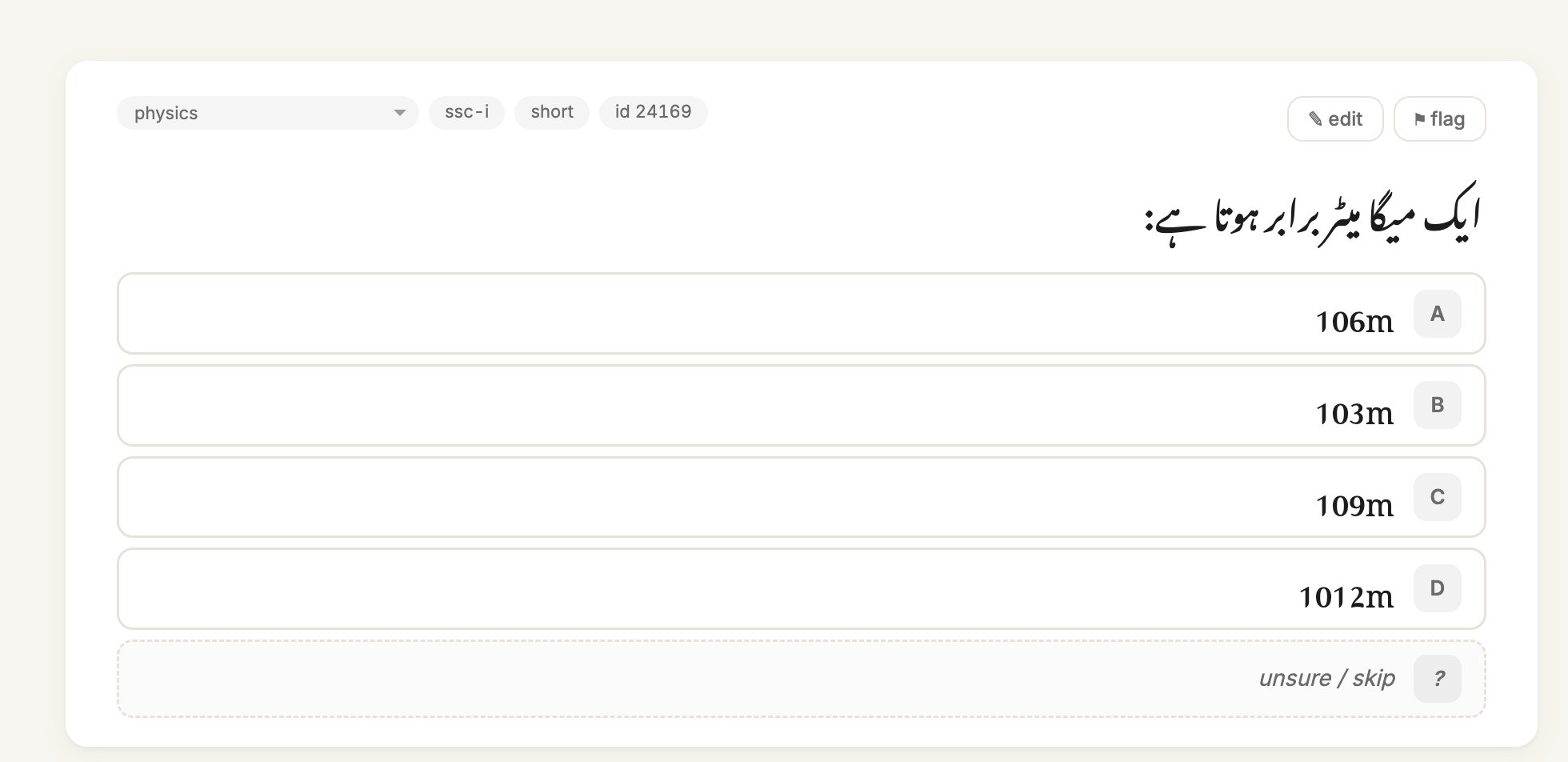}
        \caption{Problematic OCR example marked as \emph{unsure / skip}.}
        \label{fig:dash-flag-before}
    \end{subfigure}
    \hfill
    \begin{subfigure}[t]{0.48\linewidth}
        \centering
        \includegraphics[width=\linewidth]{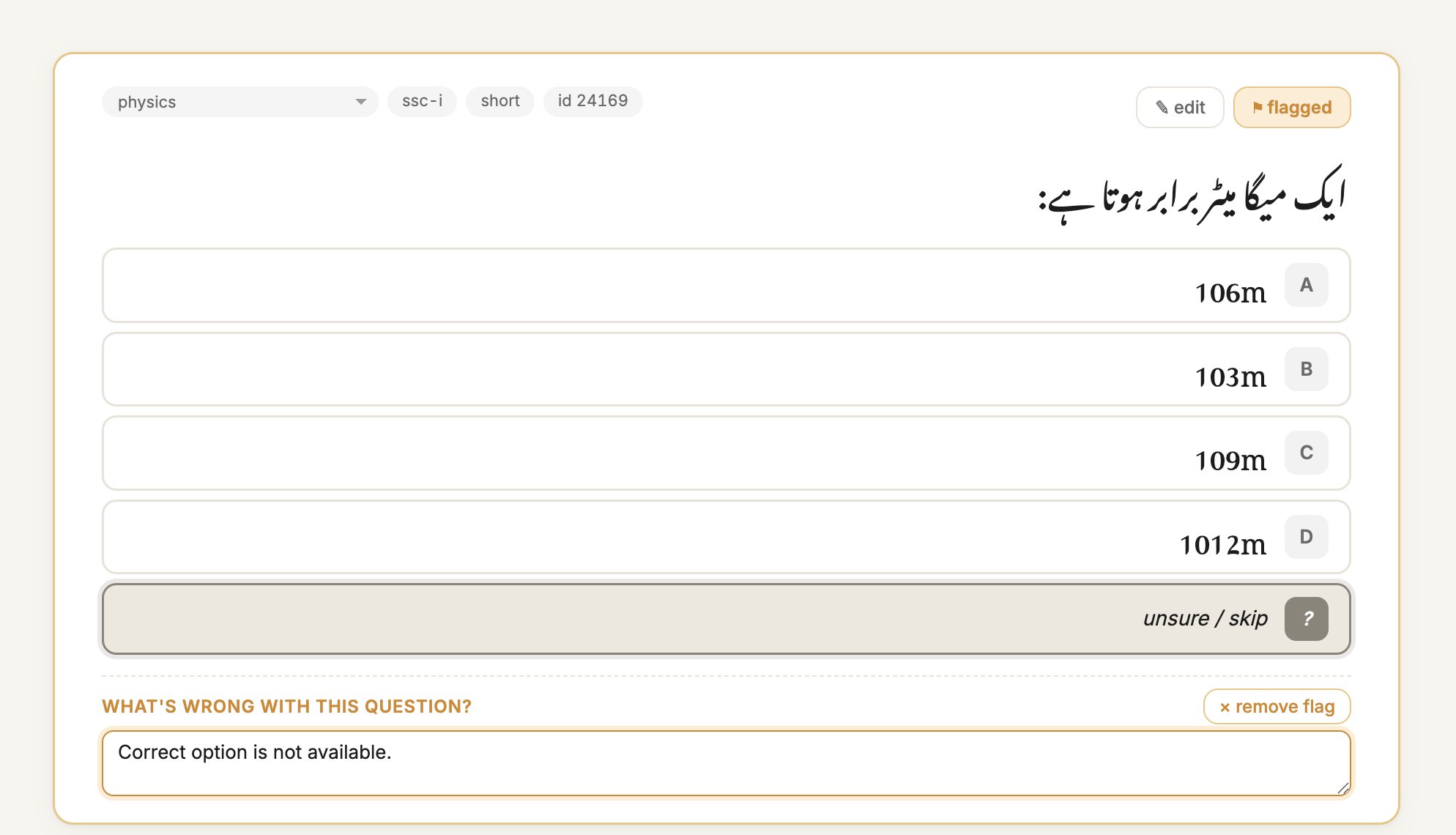}
        \caption{Flagged item with an attached review reason.}
        \label{fig:dash-flag-submitted}
    \end{subfigure}
    \caption{\textbf{Annotation dashboard.} Workflow for flagging problematic items.}
    \label{fig:dash-flag}
\end{figure*}

Section~\ref{app:annotation-guidelines} described the annotation policies; here we illustrate the dashboard used to apply them. For each item, annotators could select an answer, mark it as \emph{unsure / skip}, edit question or option text, or flag it for review with a free-text explanation. This design separated answer selection from quality-control feedback and text correction. It also allowed recoverable text errors to be corrected without automatically excluding the item. Figures~\ref{fig:dash-pick},~\ref{fig:dash-edit}, and~\ref{fig:dash-flag} illustrate the workflow.

\paragraph{Picking an answer:} Figure~\ref{fig:dash-pick} shows the annotation workflow. Annotators view the Urdu question stem, four answer options, and metadata describing the subdomain, academic level, length tier, and item identifier (Figure~\ref{fig:dash-pick}a). Selecting an option highlights the choice but does not advance to the next item (Figure~\ref{fig:dash-pick}b); annotators must confirm the selection before proceeding, which reduces accidental submissions. Keyboard shortcuts (\texttt{1}--\texttt{5} for option selection, arrow keys for navigation) support batch traversal.

\paragraph{Editing an item:} Figure~\ref{fig:dash-edit} illustrates the in-place editing UI. The example contains OCR and formatting artifacts in the answer options (Figure~\ref{fig:dash-edit}a). After entering edit mode, annotators modify the question text and options through inline editable fields (Figure~\ref{fig:dash-edit}b). The interface records all changes and attaches revision tags to each edited field for later review, which can be reverted with a single click (Figure~\ref{fig:dash-edit}c).

\paragraph{Flagging an item:} Figure~\ref{fig:dash-flag} shows the flagging UI. In the illustrated case, OCR corruption removes superscript formatting from a physics question, making all answer options invalid (Figure~\ref{fig:dash-flag}a). The annotator marks the item as \emph{unsure / skip} and submits a flag with a free-text explanation (Figure~\ref{fig:dash-flag}b). The dashboard visually highlights flagged items so admins can review them, and the inclusion rules in Section~\ref{app:annotation-inclusion} automatically remove flagged items from the consensus pool.

\subsection{Annotation Outcomes}
\label{app:annotation-outcomes}

\begin{figure*}[t]
\centering
\includegraphics[width=0.9\linewidth]{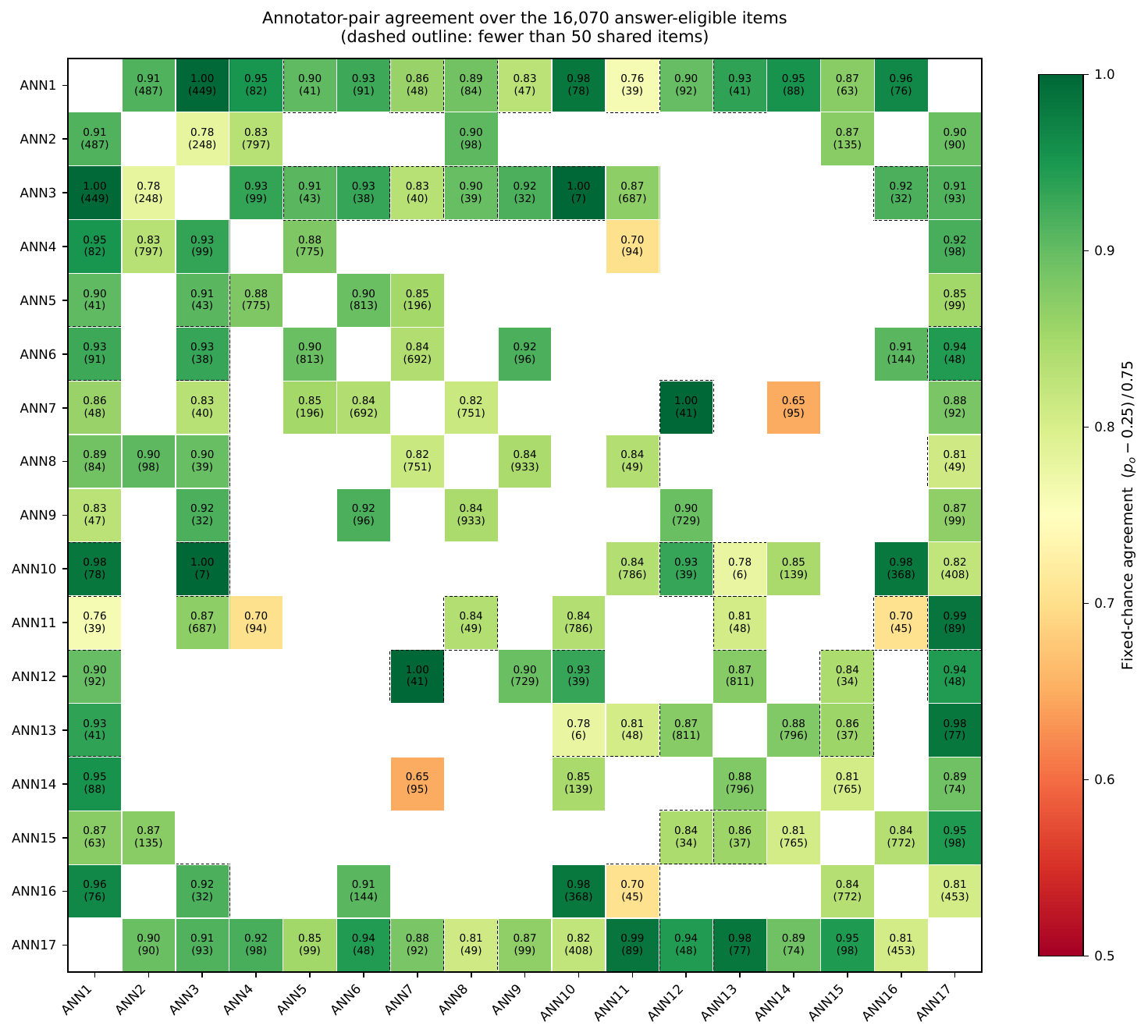}
\caption{\textbf{Pairwise annotator agreement on answer-eligible MCQs.} Each populated cell reports fixed-chance-adjusted agreement, $(p_o-0.25)/0.75$, with the number of shared answer-eligible items in parentheses. Dashed outlines mark pairs with fewer than 50 shared items, and blank cells indicate no overlap. Agreement is calculated over 16{,}070 answer-eligible items.}
\label{fig:annotator-pair-agreement}
\end{figure*}

A total of 17{,}565 exam-derived MCQs entered annotation, of which 14{,}459 were retained after applying the inclusion and edit-resolution rules from Tables~\ref{tab:annotation-inclusion-rules} and~\ref{tab:annotation-edit-rules}, giving an 82.3\% yield. Table~\ref{tab:annotation-drop-breakdown} summarizes the 3{,}106 exclusions. Answer disagreement accounts for 51.9\% of exclusions and identifies items unsuitable under a strict consensus policy. Flagged items form the second-largest category and mainly contain OCR corruption or malformed options, as in Figure~\ref{fig:dash-flag-before}.

Agreement is measured on the 16{,}070 answer-eligible items remaining after flagged, unsure, and single-annotation items are excluded. Annotators agreed on 14{,}459 items and disagreed on 1{,}611, giving observed agreement of $p_o=14{,}459/16{,}070=0.8998$. Against the fixed four-option chance baseline, fixed-chance-adjusted agreement is $(p_o-0.25)/(1-0.25)=0.8663$. Figure~\ref{fig:annotator-pair-agreement} reports the same measure by annotator pair. Each cell shows fixed-chance-adjusted agreement, where $p_o$ is the pair's observed agreement, together with the number of shared answer-eligible items. Blank cells indicate pairs with no shared answer-eligible items. Most populated cells exceed 0.85, indicating strong agreement across most annotator pairs. Lower-agreement pairs generally share fewer items and have limited influence on the aggregate result.

\section{Dataset Format}
\label{sec:dataset-format}

\begin{figure*}[t]
  \centering

  \begin{subfigure}[t]{0.58\textwidth}
    \centering
    \includegraphics[width=\linewidth]{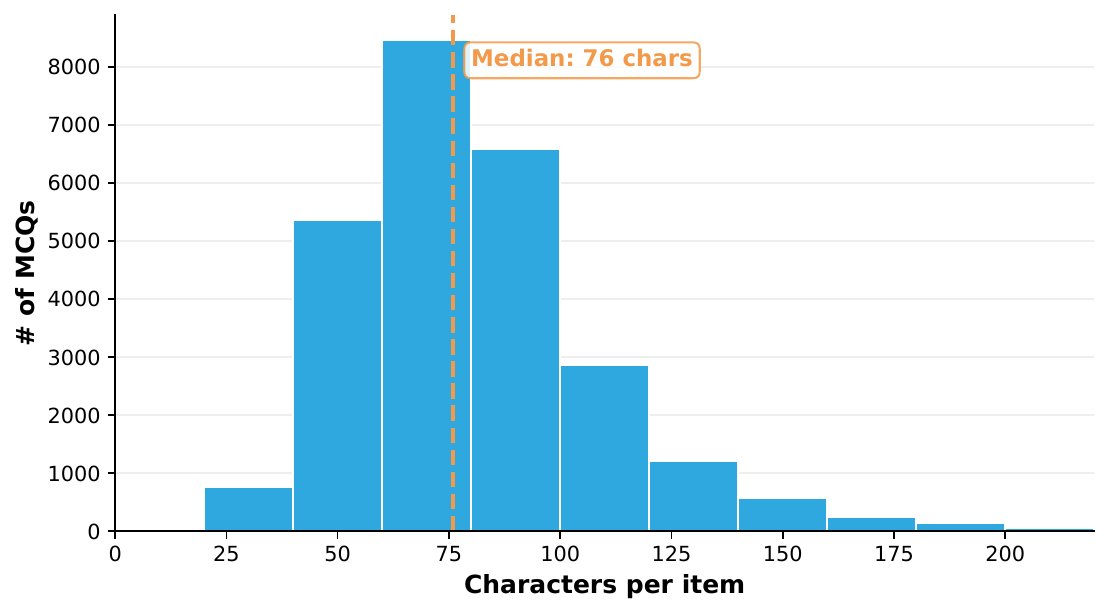}
    \caption{Character-length distribution per item.}
    \label{fig:item_length_distribution}
  \end{subfigure}
  \hfill
  \begin{subfigure}[t]{0.36\textwidth}
    \centering
    \includegraphics[width=\linewidth]{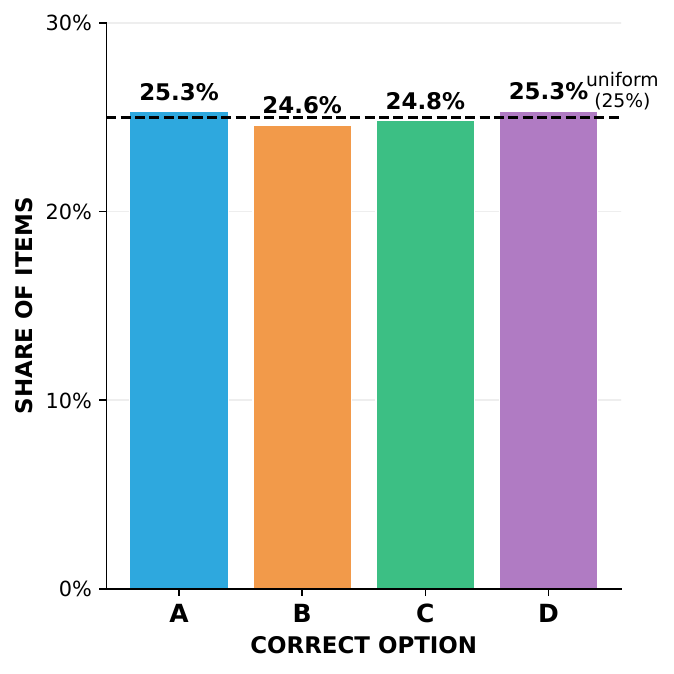}
    \caption{Gold answer-key distribution.}
    \label{fig:answer_key_distribution}
  \end{subfigure}

  \caption{
  \textbf{Dataset-level sanity checks for \ds{}.} Most questions remain compact enough for standard MCQ prompting, while the gold answer keys remain close to uniformly distributed across A--D.
  }
  \label{fig:dataset_sanity_checks}
\end{figure*}

Each \ds{} example is stored as a multiple-choice item containing a question, exactly four answer options, a gold answer label, domain and subdomain labels, academic level information, a question-length tier, and source metadata. The evaluation pipeline uses the question, options, and gold answer fields during inference, while the remaining metadata supports analysis, filtering, and reproducibility.

Figure~\ref{fig:dataset_sanity_checks} reports two dataset-level sanity checks: item-length distribution and answer-key distribution. The median serialized item length, counting the question stem together with all four options, is 76 characters; the median question stem alone is 39 characters. Gold labels are close to uniform across A--D (6{,}673 / 6{,}486 / 6{,}549 / 6{,}681), which limits option-position bias. Figure~\ref{fig:urdummlu_json_schema} shows the released JSON schema.

\begin{figure}[!h]
\centering
\begin{tcolorbox}[title={UrduMMLU JSON Schema}, width=\linewidth]
\footnotesize
\begin{verbatim}
{
  "id": "...",
  "question": "...",
  "options": {
    "A": "...",
    "B": "...",
    "C": "...",
    "D": "..."
  },
  "correct_key": "...",
  "domain": "...",
  "subdomain": "...",
  "level": "...",
  "length_tier": "...",
  "source": [
    {
      "name": "...",
      "url": "..."
    }
  ]
}
\end{verbatim}
\end{tcolorbox}
\caption{\textbf{Structure of an individual \ds{} record, as released.} \texttt{length\_tier} is \texttt{short}, \texttt{long}, or \texttt{null}; it is populated only for exam-derived items.}
\label{fig:urdummlu_json_schema}
\end{figure}

\subsection{Subject Acronyms and Education Levels}
\label{app:acronyms}

Table~\ref{tab:urdummlu-domains-levels} lists the full names, acronyms, and education levels for all 26 \ds{} subdomains, grouped by domain. We use these acronyms in the per-subdomain results tables (Tables~\ref{tab:subdomain_performance_en} and~\ref{tab:subdomain_performance_ur}).

\begin{table*}[!t]
\centering
\small
\renewcommand{\arraystretch}{0.94}

\begin{tabular*}{\textwidth}{@{\extracolsep{\fill}}lll}
\toprule
\textbf{Subdomain}
& \textbf{Acronym}
& \textbf{Education Levels} \\
\midrule

\rowcolor{softgray}
\multicolumn{3}{c}{\textbf{Humanities}} \\
\midrule

Ethics
& ETH
& SSC-I, SSC-II, HSSC-I \\

Fine Arts
& FNA
& SSC-I, SSC-II, HSSC-I \\

Islamic Studies
& ISL
& SSC-I, SSC-II, HSSC-I, HSSC-II \\

Urdu Grammar
& UGR
& SSC-I, SSC-II, HSSC-I, HSSC-II \\

Urdu Language
& ULG
& SSC-I, SSC-II, HSSC-I, HSSC-II \\

Urdu Literature
& ULT
& SSC-I, SSC-II, HSSC-I, HSSC-II \\

\midrule
\rowcolor{softgray}
\multicolumn{3}{c}{\textbf{Other}} \\
\midrule

General Knowledge
& GKN
& SSC-I, SSC-II, HSSC-I, HSSC-II \\

\midrule
\rowcolor{softgray}
\multicolumn{3}{c}{\textbf{Profession}} \\
\midrule

Professional Development
& PRD
& SSC-I, SSC-II, HSSC-I, HSSC-II \\

Professional Studies
& PRS
& SSC-I, SSC-II \\

\midrule
\rowcolor{softgray}
\multicolumn{3}{c}{\textbf{STEM}} \\
\midrule

Biology
& BIO
& SSC-I, SSC-II, HSSC-I, HSSC-II \\

Chemistry
& CHM
& SSC-I, SSC-II, HSSC-I \\

Computer Science
& CSC
& SSC-I, SSC-II, HSSC-I \\

General Science
& GSC
& SSC-I, SSC-II, HSSC-I \\

Mathematics
& MTH
& SSC-I, SSC-II, HSSC-I \\

Physics
& PHY
& SSC-I, SSC-II, HSSC-II \\

\midrule
\rowcolor{softgray}
\multicolumn{3}{c}{\textbf{Social Sciences}} \\
\midrule

Civics
& CIV
& SSC-I, SSC-II, HSSC-I, HSSC-II \\

Commerce
& COM
& SSC-I, SSC-II, HSSC-I, HSSC-II \\

Current \& International Affairs
& CIA
& SSC-I, SSC-II, HSSC-I, HSSC-II \\

Economics
& ECO
& SSC-I, SSC-II, HSSC-I, HSSC-II \\

Education
& EDU
& SSC-I, SSC-II, HSSC-I, HSSC-II \\

Geography
& GEO
& SSC-I, SSC-II, HSSC-I, HSSC-II \\

Health \& Physical Education
& HPE
& SSC-I, SSC-II, HSSC-II \\

Pakistan Studies
& PKS
& SSC-I, SSC-II, HSSC-I, HSSC-II \\

Psychology
& PSY
& HSSC-I, HSSC-II \\

Psychometrics
& PMT
& SSC-I, SSC-II, HSSC-I, HSSC-II \\

Sociology
& SOC
& HSSC-I, HSSC-II \\

\bottomrule
\end{tabular*}

\caption{\textbf{Domain coverage of \ds{}.} Subdomains, acronyms, and represented education levels within each domain.}
\label{tab:urdummlu-domains-levels}
\end{table*}

\section{Evaluation Details}
\label{app:evaluation-details}

This appendix provides the full model roster and prompt templates used in the \ds{} experiments.

\subsection{Model Roster}
\label{app:model-roster}

Table~\ref{tab:language-models} lists the 30 models evaluated in this work. We group models by family for readability, while the main paper discusses them using broader categories such as proprietary API models, open-weight multilingual models, compact models, mixture-of-experts models, reasoning-oriented variants, and Urdu- or regionally specialized models.

\subsection{Prompt Templates}
\label{app:prompt-templates}

We use separate English and Urdu prompt templates for zero-shot evaluation. Both templates present the same Urdu question and answer options while changing only the instruction language and field labels. The output format remains identical in both settings to support automatic parsing and consistent evaluation.

\paragraph{English prompt:} Figure~\ref{fig:eng_mcq_prompt} shows the English template combining fixed system and item-specific user prompts. The system prompt requires a strict two-line response with an \texttt{Answer key} and \texttt{Answer text}, without explanation or extra formatting. This structure supports deterministic answer extraction and consistent measurement of invalid outputs across models. The user prompt fills the placeholders \texttt{{domain}}, \texttt{{subdomain}}, \texttt{{level}}, \texttt{{question}}, and \texttt{{A}}--\texttt{{D}} directly from the dataset schema in Appendix~\ref{sec:dataset-format}, preserving the Urdu question content across both prompt-language settings.

\begin{figure}[!h]
\centering
\begin{tcolorbox}[title={Zero-shot English Prompt}, width=\linewidth]
\small
\textbf{System Prompt:}
You are an expert multiple-choice question answering assistant. Read the question carefully and select the single best answer. Respond in EXACTLY this two-line format, with no extra text:
\begin{itemize}
[noitemsep, topsep=2pt, leftmargin=*]
      \item Answer key: <one of A, B, C, D>
      \item Answer text: <verbatim text of the chosen option, copied character-for-character>
\end{itemize}
Do not add explanations, preambles, markdown, or punctuation
outside of the format. The Answer text must match the option text exactly so the response can be parsed programmatically.
      
\vspace{2pt}
\textbf{User Prompt:}
\begin{verbatim}
Subject: {domain} – {subdomain}
Level: {level}
Question: {question}
A) {A}
B) {B}
C) {C}
D) {D}
Answer key:
Answer text:
\end{verbatim}
\end{tcolorbox}
\caption{\textbf{Zero-shot English Prompt.} Used for multiple-choice question answering with strict answer formatting requirements.}
\label{fig:eng_mcq_prompt}
\end{figure}

\paragraph{Urdu prompt:} The Urdu prompt template mirrors the English template while translating the system instructions, user-field labels (\foreignlanguage{urdu}{مضمون}, \foreignlanguage{urdu}{سطح}, \foreignlanguage{urdu}{سوال}), and surrounding instructional text into Urdu. The Urdu question stem and answer options remain unchanged across both settings. 

We also preserve the same two-line response structure using the English fields \texttt{Answer key:} and \texttt{Answer text:}, which allows a single parser to process outputs under both prompt languages. For matched model configurations, this design ensures that prompt language is the only difference between the two evaluation settings. Figure~\ref{fig:urdu_mcq_prompt} shows the full Urdu prompt template. This setup makes the prompt-language comparison in Section~\ref{sec:results} interpretable for models with matched inference settings, since performance differences reflect instruction language rather than question content or evaluation logic.

\begin{figure}[!h]
\centering
\begin{tcolorbox}[title={Zero-shot Urdu Prompt}, width=\linewidth]
\small
\textbf{System Prompt:}

\begin{flushright}
\foreignlanguage{urdu}{جوابدہ معاون ہیں۔} MCQ \foreignlanguage{urdu}{آپ ایک ماہر} 
\foreignlanguage{urdu}{سوال کو غور سے پڑھیں اور صحیح آپشن منتخب کریں۔}
\vspace{2pt}
\foreignlanguage{urdu}{جواب بالکل اس دو سطری شکل میں دیں، کوئی اضافی متن شامل نہ کریں:}

<\foreignlanguage{urdu}{میں سے ایک } D \foreignlanguage{urdu}{یا} C, B,  A > :Answer key -

\foreignlanguage{urdu}{منتخب آپشن کا متن من و عن، حرف بہ حرف }> : Answer text -

<\foreignlanguage{urdu}{نقل کریں}

\foreignlanguage{urdu}{کوئی وضاحت، تمہید، مارک ڈاؤن، یا اضافی الفاظ نہ لکھیں۔}

\foreignlanguage{urdu}{کو آپشن کے متن سے بالکل مماثل ہونا چاہیے } Answer text

\foreignlanguage{urdu}{تاکہ پروگرام اسے درست طور پر پارس کر سکے۔}

\end{flushright}

\vspace{5pt}
\textbf{User Prompt:}

\begin{flushright}
\{subdomain\} -- \{domain\} :\foreignlanguage{urdu}{مضمون}\\
\{level\} :\foreignlanguage{urdu}{سطح}\\
\{question\} :\foreignlanguage{urdu}{سوال}\\
\{A\} : (A \\
\{B\} : (B \\
\{C\} : (C \\
\{D\} : (D \\
Answer key:\\
Answer text:
\end{flushright}

\end{tcolorbox}
\caption{\textbf{Zero-shot Urdu prompt.} Used for MCQ answering with strict answer formatting requirements.}
\label{fig:urdu_mcq_prompt}
\end{figure}

\subsection{Few-Shot Evaluation Setup}
\label{app:few-shot-setup}

We evaluate using the \texttt{lm-evaluation-harness} framework~\cite{eval-harness}. Each item is formatted as a four-option multiple-choice question, with the answer choices labeled \texttt{A} through \texttt{D}. 

The model generates the answer key and answer text, and the harness scores the generated continuation against the gold label with the \texttt{exact\_match} metric; no likelihood-based scoring is used. 

Demonstrations are drawn from a held-out 130-item pool, and those items are excluded from scoring, so all shot settings are compared on the same 26{,}259 of the 26{,}389 benchmark items.

\subsection{Implementation Details}
\label{app:implementation}

All evaluations use seed 42. Local open-weight models use \texttt{bfloat16} precision, greedy decoding, batch size 10, automatic device placement, and their chat templates.

For API-based systems, including OpenAI, Anthropic, Google Gemini, and the Hugging Face Inference API, we keep the prompt format and inference configuration fixed across English and Urdu prompts for 29 models. Qwen3.6-35B-A3B is the sole exception. Reasoning was disabled for its English-prompt run and enabled for its Urdu-prompt run. We report both results but exclude their difference from prompt-language conclusions because prompt language is not the only configuration change. We retry failed API requests up to five times and terminate evaluation after five consecutive failures to prevent silent errors.

\begin{table*}[!t]
\centering
\scriptsize
\renewcommand{\arraystretch}{1.17}

\begin{tabularx}{\textwidth}{@{}
>{\raggedright\arraybackslash}p{5.7cm}
>{\raggedright\arraybackslash}p{1.0cm}
>{\raggedright\arraybackslash}p{1.6cm}
>{\raggedright\arraybackslash}p{2.7cm}
>{\raggedright\arraybackslash}X
@{}}
\toprule
\multirow{2}{*}{\textbf{Model}} &
\multicolumn{4}{c}{\textbf{Model Metadata}} \\
\cmidrule(lr){2-5}
& \textbf{Size}
& \textbf{Family}
& \textbf{License}
& \textbf{Reference} \\
\midrule

\rowcolor{softgray}
\multicolumn{5}{c}{\textbf{Open-weight Models}} \\
\midrule

large-traversaal/Alif-1.0-8B-Instruct
& 8B
& Alif
& Apache 2.0
& \cite{shafique-etal-2025-alif} \\

\midrule

mistralai/Ministral-3-3B-Instruct-2512
& 3B
& Ministral
& Apache 2.0
& \cite{liu2026ministral3} \\

mistralai/Ministral-3-8B-Instruct-2512
& 8B
& Ministral
& Apache 2.0
& \cite{liu2026ministral3} \\

\midrule

enstazao/Qalb-1.0-8B-Instruct
& 8B
& Qalb
& Apache 2.0
& \cite{hassan2026qalb} \\

\midrule

Qwen/Qwen3-4B-Instruct-2507
& 4B
& Qwen 3
& Apache 2.0
& \cite{qwen2025qwen3} \\

Qwen/Qwen3-8B
& 8B
& Qwen 3
& Apache 2.0
& \cite{qwen2025qwen3} \\

Qwen/Qwen3.6-27B
& 27B
& Qwen 3.6
& Apache 2.0
& \cite{qwen3.6-27b} \\

Qwen/Qwen3.6-35B-A3B
& 36B
& Qwen 3.6
& Apache 2.0
& \cite{qwen36_35b_a3b} \\

\midrule

bigscience/bloomz-1b1
& 1.1B
& BLOOMZ
& Bigscience Bloom Rail 1.0
& \cite{muennighoff-etal-2023-crosslingual} \\

bigscience/bloomz-1b7
& 1.7B
& BLOOMZ
& Bigscience Bloom Rail 1.0
& \cite{muennighoff-etal-2023-crosslingual} \\

bigscience/bloomz-3b
& 3B
& BLOOMZ
& Bigscience Bloom Rail 1.0
& \cite{muennighoff-etal-2023-crosslingual} \\

bigscience/bloomz-7b1-mt
& 7B
& BLOOMZ
& Bigscience Bloom Rail 1.0
& \cite{muennighoff-etal-2023-crosslingual} \\

\midrule

deepseek-ai/DeepSeek-V4-Flash
& 284B
& DeepSeek
& DeepSeek License
& \cite{deepseekai2026deepseekv4} \\

\midrule

google/gemma-3-4b-it
& 4B
& Gemma 3
& Gemma
& \cite{gemma2025gemma3} \\

google/gemma-2-9b-it
& 9B
& Gemma
& Gemma
& \cite{gemma2024gemma2} \\

google/gemma-4-26B-A4B-it
& 27B
& Gemma
& Gemma
& \cite{google2026gemma4} \\

google/gemma-4-31B-it
& 31B
& Gemma
& Gemma
& \cite{google2026gemma4} \\

\midrule

meta-llama/Llama-3.2-3B-Instruct
& 3B
& LLaMA 3.2
& LLaMA License
& \cite{meta2024llama32} \\

meta-llama/Llama-3.1-8B-Instruct
& 8B
& LLaMA 3.1
& LLaMA License
& \cite{dubey2024llama3} \\

meta-llama/Llama-4-Scout-17B-16E-Instruct
& 109B
& LLaMA 4
& LLaMA License
& \cite{meta2025llama4} \\

meta-llama/Llama-4-Maverick-17B-128E-Instruct
& 402B
& LLaMA 4
& LLaMA License
& \cite{meta2025llama4} \\

meta-llama/Llama-3.3-70B-Instruct
& 70B
& LLaMA 3.3
& LLaMA License
& \cite{meta2024llama33} \\

\midrule

microsoft/Phi-4-mini-instruct
& 3B
& Phi-4
& MIT
& \cite{microsoft2025phi4mini} \\

microsoft/Phi-3.5-mini-instruct
& 4B
& Phi-3.5
& MIT
& \cite{abdin2024phi3} \\

\midrule
\rowcolor{softgray}
\multicolumn{5}{c}{\textbf{Proprietary Models}} \\
\midrule

claude-haiku-4-5
& N/D
& Claude
& Proprietary
& \cite{anthropic2025haiku45} \\

claude-sonnet-4-6
& N/D
& Claude
& Proprietary
& \cite{anthropic2026sonnet46} \\

\midrule

gemini-3.1-flash-lite
& N/D
& Gemini
& Proprietary
& \cite{google2026gemini31flashlite} \\

gemini-3.5-flash
& N/D
& Gemini
& Proprietary
& \cite{google2026gemini35flash} \\

\midrule

gpt-5.4-mini
& N/D
& GPT
& Proprietary
& \cite{singh2026openaigpt5card} \\

gpt-5.4
& N/D
& GPT
& Proprietary
& \cite{singh2026openaigpt5card} \\

\bottomrule
\end{tabularx}

\caption{\textbf{Language models evaluated in this study.} Sizes are reported when publicly disclosed; N/D denotes not disclosed.}
\label{tab:language-models}
\end{table*}

\begin{table}[!t]
\centering
\small
\renewcommand{\arraystretch}{1.1}
\begin{tabular*}{\columnwidth}{@{\extracolsep{\fill}}lrrr}
\toprule
\textbf{Model} & \textbf{STEM} & \textbf{Humanities} & \textbf{Gap} \\
\midrule
\rowcolor{softgray}
\multicolumn{4}{c}{\textbf{Open-weight Models: $>$ 25B Parameters}} \\
\midrule
DeepSeek-V4-Flash
& 97.36
& 67.30
& 30.06 \\
Gemma-4-26B-A4B-IT
& 87.19
& 57.87
& 29.32 \\
Gemma-4-31B-IT
& 93.86
& 63.38
& 30.48 \\
LLaMA-3.3-70B
& 78.39
& 56.24
& 22.15 \\
Qwen3.6-27B
& 91.12
& 55.86
& 35.26 \\
Qwen3.6-35B-A3B
& 96.15
& 57.98
& 38.17 \\
LLaMA-4-Scout-17B
& 85.39
& 56.69
& 28.70 \\
LLaMA-4-Maverick-17B
& 92.24
& 63.41
& 28.83 \\
\midrule
\rowcolor{softgray}
\multicolumn{4}{c}{\textbf{Open-weight Models: $\leq$ 25B Parameters}} \\
\midrule
BLOOMZ-1.1B
& 24.53
& 25.89
& $-$1.36 \\
BLOOMZ-1.7B
& 17.15
& 20.78
& $-$3.63 \\
BLOOMZ-3B
& 9.74
& 3.60
& 6.14 \\
BLOOMZ-7B
& 22.75
& 17.75
& 5.00 \\
Gemma-2-9B-IT
& 69.02
& 48.21
& 20.81 \\
Gemma-3-4B-IT
& 51.79
& 38.37
& 13.42 \\
LLaMA-3.2-3B
& 37.24
& 29.39
& 7.85 \\
LLaMA-3.1-8B
& 46.49
& 37.71
& 8.78 \\
Ministral-3-3B
& 57.25
& 43.17
& 14.08 \\
Ministral-3-8B
& 71.37
& 45.86
& 25.51 \\
Phi-4-mini
& 37.08
& 28.77
& 8.31 \\
Phi-3.5-mini
& 33.54
& 27.22
& 6.32 \\
Qwen3-4B-Instruct
& 68.75
& 43.11
& 25.64 \\
Qwen3-8B
& 74.05
& 30.77
& 43.28 \\
\midrule
\rowcolor{softgray}
\multicolumn{4}{c}{\textbf{Proprietary Models}} \\
\midrule
Claude-Haiku-4.5
& 91.92
& 59.42
& 32.50 \\
Claude-Sonnet-4.6
& 96.26
& 72.87
& 23.39 \\
Gemini-3.1-Flash-Lite
& 97.09
& 74.57
& 22.52 \\
Gemini-3.5-Flash
& 97.75
& 85.49
& 12.26 \\
GPT-5.4-mini
& 88.25
& 62.51
& 25.74 \\
GPT-5.4
& 97.40
& 74.29
& 23.11 \\
\midrule
\rowcolor{softgray}
\multicolumn{4}{c}{\textbf{Urdu Models}} \\
\midrule
Qalb-1.0-8B
& 30.59
& 31.39
& $-$0.80 \\
Alif-1.0-8B
& 29.04
& 27.98
& 1.06 \\
\bottomrule
\end{tabular*}
\caption{\textbf{STEM--Humanities gap under the Urdu prompt.} All-item domain accuracy and gaps in percentage points, ordered as in Table~\ref{tab:language-models}. Small or negative gaps coincide with near-chance or low accuracy rather than balanced competence. Values come from Table~\ref{tab:modelperformance}.}
\label{tab:stem-hum-gap}
\end{table}

\section{Detailed Results}
\label{app:detailed-results}

This section provides a detailed view of the results summarized in Table~\ref{tab:modelperformance}. Table~\ref{tab:stem-hum-gap} reports each model's STEM--Humanities accuracy gap under the Urdu prompt. Nearly all models performing above chance score higher on STEM than on Humanities. The gaps are small, and negative in three cases, for the BLOOMZ family and the two Urdu-targeted models, but for different reasons.

BLOOMZ accuracy remains near or below the random baseline on both domains, so these small gaps do not indicate balanced domain knowledge. Invalid outputs contribute to the low scores for some checkpoints, especially BLOOMZ-3B, which has a $74.2\%$ invalid-output rate. However, it is the only BLOOMZ checkpoint for which most outputs are invalid. The corresponding rates are below $0.1\%$ for BLOOMZ-1.1B, $24.7\%$ for BLOOMZ-1.7B, and $34.5\%$ for BLOOMZ-7B. Unparseability therefore affects some BLOOMZ results but does not explain the pattern uniformly. The Urdu-targeted models show small gaps because their accuracy is low on both domains, particularly STEM, where they trail comparably sized general-purpose models. For Qalb-1.0-8B, the gap reverses to $-0.80$ points.

\subsection{Per-Subdomain Results}
\label{app:detailed-results}

We expand the domain-level results from Table~\ref{tab:modelperformance} to all 26 subdomains. Table~\ref{tab:subdomain_performance_en} reports accuracy under the English prompt, while Table~\ref{tab:subdomain_performance_ur} reports accuracy under the Urdu prompt. Both tables follow the same ordering, with subdomains grouped by domain and sorted by dataset size.

\subsubsection{Subject-Wise Behavior}
\label{app:subject-wise}

The subdomain results sharpen the main finding from Section~\ref{sec:results}: STEM subjects transfer more reliably than Urdu-centered Humanities subjects. The strongest models approach saturation on several STEM subdomains, while performance remains lower on Urdu-centered subjects. This contrast persists even among the highest-scoring models and is partly hidden by aggregate accuracy.

Under the English prompt, Gemini-3.5-Flash reaches $97.86\%$ on chemistry, $98.60\%$ on biology, and $98.74\%$ on mathematics, while DeepSeek-V4-Flash reaches $98.71\%$ on physics. These scores remain nearly unchanged under the Urdu prompt, and in some cases increase slightly. The consistency across prompt languages suggests that scientific and mathematical concepts transfer relatively cleanly once the model can process Urdu input. Humanities presents a much harder challenge. Islamic studies and Urdu grammar remain accessible, with Gemini-3.5-Flash reaching $94.95\%$ and $88.34\%$, respectively, under the English prompt.

In contrast, Urdu literature remains difficult across the entire model suite. Even the strongest model reaches only $80.34\%$ under the English prompt and $80.76\%$ under the Urdu prompt. Most other proprietary and open-weight models perform substantially worse, often trailing by another 10 to 20 points. Urdu language occupies an intermediate position, with top scores near $89\%$. Across nearly all capable models, the same ordering persists: Islamic studies $>$ Urdu grammar $>$ Urdu language $>$ Urdu literature. This consistency suggests that the differences reflect genuine variation in subject difficulty rather than isolated model behavior. Social Sciences contains both highly accessible and consistently difficult subdomains. Geography, civics, sociology, psychology, and commerce all exceed $93\%$ accuracy for the strongest models. Pakistan studies also remains relatively strong despite its large size. In contrast, current and international affairs and psychometrics stand out as the two hardest Social Sciences subdomains. 

Current and international affairs peaks at roughly $78\%$ under both prompts, likely because many questions depend on time-sensitive world knowledge beyond pretraining cutoffs. Psychometrics is even more difficult: no model in the evaluation exceeds $60\%$ accuracy under either prompt language. This suggests that both subdomains are challenging even for the strongest models. The smaller Profession and Other domains follow patterns similar to Social Sciences, with proprietary models reaching the low 90s and smaller open-weight models trailing behind. These domains do not introduce additional failure modes. The subdomain results further clarify the behavior of smaller open-weight models. 

Among models with fewer than 25B parameters, Gemma-2-9B-IT performs best on Humanities subjects, including Urdu language, Urdu grammar, ethics, and fine arts, while Qwen3-8B leads on STEM subjects such as chemistry, mathematics, computer science, and physics. This pattern mirrors the domain-level results: Qwen3-8B retains relatively strong scientific knowledge but struggles on Urdu-centered humanities content, whereas Gemma-2-9B-IT shows more balanced performance across subdomains. The Urdu-targeted Qalb-1.0-8B and Alif-1.0-8B lead no subdomain and trail similarly sized general-purpose models. BLOOMZ remains near random on most subdomains, consistent with its high invalid-output rates in Section~\ref{sec:output-validity}.

\subsubsection{The Psychometrics Gap}
\label{app:psychometrics}

Psychometrics is the most difficult subdomain in our evaluation. The best English-prompt accuracy reaches only $57.30\%$ (Gemini-3.5-Flash), while the best Urdu-prompt accuracy reaches $52.97\%$ (Claude-Sonnet-4.6). No model exceeds $60\%$ under either prompt setting, in contrast with the $90$--$98\%$ accuracies achieved on many STEM and Social Sciences subjects. The difficulty appears specific to the content rather than the prompt language. English- and Urdu-prompt results remain close, and model rankings on psychometrics largely mirror their overall rankings. Psychometrics questions in Urdu SSC/HSSC curricula frequently emphasize analogies, number series, logical patterns, and aptitude-style reasoning tasks that require abstract structure recognition rather than factual recall. These results suggest that current LLMs still struggle on reasoning-heavy Urdu educational content even when they perform strongly on factual subjects.

\begin{table*}[t]
\centering
\resizebox{\textwidth}{!}{%
\begin{tabular}{l *{6}{r} *{6}{r} *{11}{r} *{2}{r} *{1}{r}}
\toprule
\multirow{2}{*}{\textbf{Model}} &
\multicolumn{6}{c}{\textbf{STEM}} &
\multicolumn{6}{c}{\textbf{Humanities}} &
\multicolumn{11}{c}{\textbf{Social Sciences}} &
\multicolumn{2}{c}{\textbf{Profession}} &
\multicolumn{1}{c}{\textbf{Other}} \\
\cmidrule(lr){2-7}\cmidrule(lr){8-13}\cmidrule(lr){14-24}\cmidrule(lr){25-26}\cmidrule(lr){27-27}
& \textbf{CHM} & \textbf{BIO} & \textbf{GSC} & \textbf{MTH} & \textbf{CSC} & \textbf{PHY} & \textbf{ULT} & \textbf{ULG} & \textbf{ISL} & \textbf{UGR} & \textbf{ETH} & \textbf{FNA} & \textbf{PKS} & \textbf{ECO} & \textbf{EDU} & \textbf{SOC} & \textbf{HPE} & \textbf{CIV} & \textbf{GEO} & \textbf{PSY} & \textbf{CIA} & \textbf{COM} & \textbf{PMT} & \textbf{PRD} & \textbf{PRS} & \textbf{GKN} \\
\midrule
\rowcolor{softgray}
\multicolumn{27}{c}{\textbf{Open-weight Models: $>$ 25B Parameters}} \\
\midrule
DeepSeek-V4-Flash
  & \textbf{97.65} & \textbf{97.96} & \textbf{96.44} & \textbf{98.17} & \textbf{95.63} & \fbox{\textbf{98.71}}
  & \textbf{56.65} & \textbf{77.38} & \textbf{91.92} & \textbf{79.47} & \textbf{88.60} & \textbf{87.16}
  & \textbf{91.83} & \textbf{93.94} & \textbf{85.17} & \fbox{\textbf{95.12}} & \textbf{92.05} & \textbf{93.78} & \textbf{94.29} & \textbf{94.78} & \textbf{71.40} & \fbox{\textbf{95.04}} & 40.00
  & \textbf{89.30} & \textbf{90.85}
  & \textbf{86.56} \\
Gemma-4-26B-A4B-IT
  & 85.47 & 83.57 & 85.64 & 85.05 & 88.46 & 88.41
  & 45.74 & 66.49 & 78.65 & 71.26 & 71.93 & 71.62
  & 73.06 & 73.10 & 74.64 & 88.82 & 71.16 & 78.37 & 79.52 & 81.09 & 42.56 & 77.54 & 44.32
  & 71.60 & 80.17
  & 70.70 \\
Gemma-4-31B-IT
  & 91.88 & 93.45 & 92.33 & 93.49 & 93.44 & 96.46
  & 51.61 & 72.36 & 85.32 & 77.35 & 84.21 & 82.43
  & 80.43 & 84.34 & 82.06 & 90.75 & 81.13 & 87.26 & 87.62 & 90.00 & 54.92 & 86.05 & 40.54
  & 79.57 & 88.67
  & 79.66 \\
LLaMA-3.3-70B
  & 79.06 & 78.41 & 86.50 & 77.40 & 85.54 & 81.32
  & 46.25 & 62.86 & 76.16 & 64.77 & 79.82 & 75.00
  & 77.20 & 75.74 & 77.39 & 86.76 & 67.65 & 82.96 & 82.86 & 81.52 & 47.60 & 77.07 & 29.19
  & 71.79 & 82.14
  & 73.94 \\
Qwen3.6-27B
  & 91.88 & 86.57 & 89.31 & 89.61 & 92.95 & 93.72
  & 45.49 & 62.07 & 75.29 & 69.80 & 77.19 & 79.73
  & 66.60 & 78.72 & 76.32 & 88.82 & 69.54 & 77.93 & 81.90 & 83.70 & 39.13 & 78.25 & \textbf{47.57}
  & 65.18 & 85.84
  & 69.75 \\
Qwen3.6-35B-A3B
  & 88.14 & 86.25 & 89.42 & 86.64 & 92.47 & 90.66
  & 46.13 & 62.74 & 79.12 & 67.55 & 77.63 & 79.05
  & 68.50 & 78.50 & 75.12 & 88.05 & 68.73 & 80.00 & 80.63 & 83.70 & 43.02 & 78.96 & 41.08
  & 70.04 & 81.92
  & 69.97 \\
LLaMA-4-Scout-17B-16E
  & 84.19 & 83.24 & 86.29 & 78.08 & 85.78 & 88.57
  & 47.79 & 63.14 & 75.62 & 65.30 & 73.25 & 76.35
  & 68.13 & 73.98 & 72.85 & 86.25 & 65.63 & 77.93 & 75.08 & 77.83 & 48.51 & 72.81 & 39.46
  & 68.09 & 81.26
  & 69.68 \\
LLaMA-4-Maverick-17B-128E
  & 91.99 & 90.23 & 91.79 & 91.89 & 92.59 & 94.04
  & 52.47 & 70.68 & 84.04 & 74.17 & 83.77 & 81.76
  & 81.65 & 83.35 & 77.87 & 88.43 & 80.19 & 87.41 & 84.76 & 86.96 & 52.40 & 79.43 & 34.59
  & 78.79 & 85.19
  & 80.84 \\
\midrule
\rowcolor{softgray}
\multicolumn{27}{c}{\textbf{Open-weight Models: $\leq$ 25B Parameters}} \\
\midrule
BLOOMZ-1.1B
  & 24.47 & 24.06 & 19.01 & 24.09 & 25.64 & 24.32
  & 27.80 & 27.32 & 25.19 & 24.24 & 28.07 & 20.95
  & 24.97 & 24.26 & 21.65 & 24.42 & 24.53 & 24.00 & 23.65 & 25.22 & 25.63 & 26.48 & 24.86
  & 24.90 & 27.23
  & 25.55 \\
BLOOMZ-1.7B
  & 25.53 & 24.81 & 31.32 & 25.68 & 33.29 & 23.83
  & 26.69 & 27.72 & 29.63 & 23.31 & 31.58 & 22.30
  & 31.18 & 32.30 & 38.28 & 42.29 & 32.08 & 34.67 & 29.68 & 27.61 & 28.83 & 30.97 & 35.68
  & 27.04 & 44.23
  & 30.25 \\
BLOOMZ-3B
  & 25.11 & 30.40 & 29.27 & 27.74 & 25.52 & 25.93
  & 28.45 & 22.58 & 30.51 & 20.13 & 31.14 & 29.73
  & 30.33 & 34.95 & 32.30 & 38.82 & 29.78 & 38.67 & 29.21 & 32.17 & 28.60 & 31.21 & 14.59
  & 26.46 & 32.03
  & 27.97 \\
BLOOMZ-7B
  & 24.15 & 25.99 & 27.32 & 24.77 & 23.69 & 23.19
  & 23.26 & 21.66 & 27.68 & 27.02 & 30.26 & 25.68
  & 29.11 & 29.77 & 36.84 & 36.38 & 31.27 & 28.89 & 28.73 & 28.48 & 25.40 & 28.13 & 21.08
  & 24.71 & 33.99
  & 27.53 \\
Gemma-2-9B-IT
  & 65.28 & \textbf{64.02} & \textbf{71.06} & 64.16 & 71.57 & 66.67
  & 40.98 & \textbf{52.21} & 56.03 & \textbf{54.44} & \textbf{65.35} & \textbf{64.86}
  & \textbf{52.55} & 58.99 & \textbf{65.07} & \textbf{75.71} & 51.35 & 64.00 & 62.86 & 61.30 & \textbf{40.96} & \textbf{63.12} & 30.81
  & 49.81 & \textbf{71.68}
  & \textbf{54.70} \\
Gemma-3-4B-IT
  & 46.37 & 46.29 & 55.83 & 43.49 & 60.27 & 47.67
  & 33.96 & 40.89 & 43.57 & 41.19 & 50.44 & 54.73
  & 42.74 & 46.42 & 51.08 & 64.78 & 41.64 & 58.07 & 53.33 & 46.09 & 31.58 & 49.65 & 23.78
  & 41.25 & 55.34
  & 45.67 \\
LLaMA-3.2-3B
  & 32.48 & 34.05 & 40.28 & 38.47 & 42.41 & 35.10
  & 21.42 & 31.91 & 32.79 & 36.03 & 41.67 & 27.03
  & 33.99 & 36.38 & 43.42 & 48.33 & 35.85 & 48.89 & 42.70 & 32.17 & 21.74 & 36.88 & 31.89
  & 34.82 & 37.47
  & 35.68 \\
LLaMA-3.1-8B
  & 44.98 & 42.53 & 51.94 & 44.86 & 52.86 & 44.28
  & 28.21 & 44.40 & 46.80 & 46.36 & 53.07 & 49.32
  & 48.04 & 49.61 & 55.02 & 65.17 & 43.53 & 56.89 & 52.70 & 47.61 & 26.77 & 47.75 & 30.81
  & 41.63 & 56.43
  & 44.64 \\
Ministral-3-3B
  & 49.79 & 48.98 & 59.40 & 54.68 & 60.39 & 58.94
  & \textbf{41.25} & 44.36 & 49.09 & 46.36 & 50.44 & 58.78
  & 42.47 & 49.72 & 55.74 & 62.47 & 46.63 & 51.26 & 55.08 & 50.87 & 37.07 & 49.88 & 28.65
  & 42.41 & 56.64
  & 48.02 \\
Ministral-3-8B
  & 61.43 & 62.41 & 70.52 & 65.41 & \textbf{77.52} & 71.98
  & 39.38 & 47.51 & \textbf{57.78} & 52.58 & 64.47 & 56.76
  & 52.44 & \textbf{61.08} & 62.68 & 74.55 & \textbf{53.91} & \textbf{67.26} & \textbf{65.40} & \textbf{63.70} & 37.30 & 59.57 & \textbf{42.70}
  & \textbf{51.75} & 68.63
  & 54.55 \\
Phi-4-mini
  & 35.47 & 28.46 & 36.39 & 45.21 & 41.68 & 35.43
  & 25.57 & 33.07 & 32.73 & 30.33 & 41.67 & 26.35
  & 33.78 & 39.36 & 43.90 & 54.88 & 31.81 & 42.67 & 37.62 & 37.39 & 21.74 & 39.01 & 29.19
  & 34.63 & 47.06
  & 32.75 \\
Phi-3.5-mini
  & 33.44 & 28.57 & 29.70 & 43.72 & 34.51 & 33.01
  & 19.54 & 28.32 & 26.46 & 21.46 & 35.09 & 20.27
  & 27.15 & 32.52 & 34.69 & 38.82 & 31.27 & 33.48 & 31.75 & 30.22 & 18.76 & 34.75 & 20.00
  & 26.65 & 38.56
  & 28.34 \\
Qwen3-4B-Instruct-2507
  & \textbf{66.88} & 57.79 & 68.68 & 71.12 & 77.40 & 72.14
  & 38.11 & 43.80 & 49.83 & 51.66 & 48.68 & 60.14
  & 44.11 & 52.04 & 56.70 & 67.99 & 49.33 & 57.19 & 56.83 & 55.65 & 36.84 & 50.35 & 29.73
  & 43.19 & 63.40
  & 47.94 \\
Qwen3-8B
  & \textbf{66.88} & 60.47 & 70.63 & \textbf{75.57} & \textbf{77.52} & \textbf{76.01}
  & 30.77 & 46.35 & 50.91 & 51.66 & 60.09 & 46.62
  & 45.23 & 56.45 & 62.68 & 71.08 & 52.43 & 59.85 & 57.46 & 59.35 & 29.98 & 57.92 & 30.27
  & 45.91 & 68.85
  & 50.29 \\
\midrule
\rowcolor{softgray}
\multicolumn{27}{c}{\textbf{Proprietary Models}} \\
\midrule
Claude-Haiku-4.5
  & 90.17 & 89.26 & 87.69 & 91.67 & 91.98 & 92.91
  & 47.58 & 64.90 & 81.82 & 69.01 & 78.07 & 77.03
  & 71.90 & 80.26 & 75.72 & 87.66 & 74.80 & 81.78 & 82.06 & 85.43 & 46.91 & 79.67 & 40.54
  & 69.84 & 83.01
  & 74.30 \\
Claude-Sonnet-4.6
  & 96.26 & 96.78 & 94.38 & 97.83 & 95.26 & 98.07
  & 61.90 & 81.53 & 91.72 & 84.50 & 87.72 & \fbox{\textbf{89.19}}
  & 88.71 & 91.40 & 83.85 & 92.80 & 83.96 & 92.74 & 92.54 & 94.35 & 65.45 & 91.49 & 52.43
  & 85.60 & 89.32
  & 86.20 \\
Gemini-3.1-Flash-Lite
  & 97.76 & 97.10 & 96.00 & 96.46 & \fbox{\textbf{96.23}} & 97.75
  & 63.53 & 83.53 & 91.52 & 87.15 & \fbox{\textbf{92.11}} & 85.14
  & 92.42 & 94.16 & \fbox{\textbf{88.04}} & \textbf{93.70} & 90.70 & 94.37 & 94.60 & \fbox{\textbf{95.43}} & 67.05 & 94.56 & 38.92
  & 87.55 & \fbox{\textbf{93.68}}
  & 86.27 \\
Gemini-3.5-Flash
  & \fbox{\textbf{97.86}} & \fbox{\textbf{98.60}} & \fbox{\textbf{96.76}} & \fbox{\textbf{98.74}} & 94.90 & \textbf{98.55}
  & \fbox{\textbf{80.34}} & \fbox{\textbf{89.15}} & \fbox{\textbf{94.95}} & \fbox{\textbf{88.34}} & 89.47 & 87.16
  & \fbox{\textbf{95.02}} & \fbox{\textbf{94.27}} & 87.80 & \textbf{93.70} & \fbox{\textbf{93.40}} & \fbox{\textbf{94.96}} & \fbox{\textbf{95.87}} & 95.00 & \fbox{\textbf{78.49}} & \textbf{94.80} & \fbox{\textbf{57.30}}
  & \fbox{\textbf{92.22}} & 92.37
  & \fbox{\textbf{90.82}} \\
GPT-5.4-mini
  & 90.28 & 89.26 & 89.52 & 80.59 & 90.77 & 90.02
  & 51.87 & 71.84 & 82.76 & 72.32 & 80.70 & 79.05
  & 78.42 & 81.15 & 76.32 & 89.85 & 67.52 & 83.11 & 83.02 & 84.78 & 56.75 & 79.43 & 35.68
  & 73.74 & 85.40
  & 75.40 \\
GPT-5.4
  & 96.26 & 96.56 & 94.06 & 92.81 & 94.90 & 96.46
  & 57.91 & 78.62 & 91.11 & 78.28 & 84.65 & 86.49
  & 87.49 & 88.20 & 82.30 & 91.26 & 86.93 & 91.56 & 91.43 & 91.74 & 70.02 & 91.25 & 42.70
  & 83.66 & 88.67
  & 84.29 \\
\midrule
\rowcolor{softgray}
\multicolumn{27}{c}{\textbf{Urdu Models}} \\
\midrule
Qalb-1.0-8B
  & 35.26 & 34.26 & 45.68 & 33.68 & 43.38 & 36.71
  & \textbf{27.09} & \textbf{31.19} & \textbf{35.96} & \textbf{32.45} & \textbf{40.79} & \textbf{41.22}
  & 36.74 & 36.38 & 36.12 & 47.69 & 28.44 & \textbf{47.11} & 43.49 & 35.22 & \textbf{30.89} & 38.06 & 24.32
  & \textbf{37.74} & 43.36
  & 39.65 \\
Alif-1.0-8B
  & \textbf{35.58} & \textbf{37.27} & \textbf{47.95} & \textbf{36.53} & \textbf{47.14} & \textbf{39.45}
  & 20.48 & 31.11 & 34.14 & 26.09 & 40.35 & 35.14
  & \textbf{39.93} & \textbf{40.79} & \textbf{42.34} & \textbf{52.31} & \textbf{35.58} & 46.37 & \textbf{46.67} & \textbf{40.22} & 26.32 & \textbf{40.19} & \textbf{25.41}
  & 32.68 & \textbf{46.62}
  & \textbf{41.04} \\
\bottomrule
\end{tabular}%
}
\caption{\textbf{Subdomain-level model performance on \ds{} under the English prompt.} All-item accuracy (\%) across all 26 subdomains grouped by domain, with unparsable outputs scored as incorrect. Subdomains are ordered within each domain by dataset size (descending); see Table~\ref{tab:urdummlu-domains-levels} for acronym expansions. Boxed values mark the best score per column, while bold values indicate the best score within each model group.}
\label{tab:subdomain_performance_en}
\end{table*}

\begin{table*}[!h]
\centering
\resizebox{\textwidth}{!}{%
\begin{tabular}{l *{6}{r} *{6}{r} *{11}{r} *{2}{r} *{1}{r}}
\toprule
\multirow{2}{*}{\textbf{Model}} &
\multicolumn{6}{c}{\textbf{STEM}} &
\multicolumn{6}{c}{\textbf{Humanities}} &
\multicolumn{11}{c}{\textbf{Social Sciences}} &
\multicolumn{2}{c}{\textbf{Profession}} &
\multicolumn{1}{c}{\textbf{Other}} \\
\cmidrule(lr){2-7}\cmidrule(lr){8-13}\cmidrule(lr){14-24}\cmidrule(lr){25-26}\cmidrule(lr){27-27}
& \textbf{CHM} & \textbf{BIO} & \textbf{GSC} & \textbf{MTH} & \textbf{CSC} & \textbf{PHY} & \textbf{ULT} & \textbf{ULG} & \textbf{ISL} & \textbf{UGR} & \textbf{ETH} & \textbf{FNA} & \textbf{PKS} & \textbf{ECO} & \textbf{EDU} & \textbf{SOC} & \textbf{HPE} & \textbf{CIV} & \textbf{GEO} & \textbf{PSY} & \textbf{CIA} & \textbf{COM} & \textbf{PMT} & \textbf{PRD} & \textbf{PRS} & \textbf{GKN} \\
\midrule
\rowcolor{softgray}
\multicolumn{27}{c}{\textbf{Open-weight Models: $>$ 25B Parameters}} \\
\midrule
DeepSeek-V4-Flash
  & \textbf{97.65} & \textbf{97.42} & \textbf{96.44} & \textbf{97.83} & \fbox{\textbf{96.48}} & \fbox{\textbf{98.71}}
  & \textbf{54.62} & \textbf{76.51} & \textbf{91.58} & \textbf{78.28} & \textbf{85.53} & \textbf{85.81}
  & \textbf{89.93} & \textbf{92.83} & \textbf{85.53} & \fbox{\textbf{93.70}} & \textbf{90.03} & \textbf{94.22} & \textbf{93.17} & \textbf{93.70} & \textbf{66.82} & \textbf{94.09} & 38.92
  & \textbf{87.94} & \textbf{91.50}
  & \textbf{84.88} \\
Gemma-4-26B-A4B-IT
  & 87.82 & 84.32 & 86.93 & 85.27 & 89.67 & 90.34
  & 46.58 & 65.78 & 78.72 & 69.93 & 74.56 & 74.32
  & 71.53 & 78.50 & 76.91 & 89.59 & 73.72 & 81.33 & 80.95 & 84.35 & 43.94 & 81.09 & 41.08
  & 72.76 & 83.66
  & 71.95 \\
Gemma-4-31B-IT
  & 92.52 & 93.98 & 93.63 & 93.61 & 94.41 & 95.65
  & 51.41 & 71.68 & 84.92 & 76.42 & 82.46 & 84.46
  & 80.91 & 84.67 & 82.30 & 91.13 & 81.81 & 88.30 & 86.98 & 90.43 & 53.78 & 86.76 & 41.08
  & 79.57 & 89.54
  & 78.56 \\
LLaMA-3.3-70B
  & 74.04 & 77.12 & 83.69 & 72.26 & 83.72 & 80.52
  & 47.31 & 60.51 & 75.02 & 64.37 & 76.75 & 75.68
  & 74.07 & 75.08 & 74.28 & 85.48 & 65.63 & 80.15 & 78.89 & 78.26 & 48.51 & 76.83 & 35.14
  & 65.37 & 78.21
  & 71.81 \\
Qwen3.6-27B
  & 92.95 & 88.83 & 89.42 & 88.93 & 93.07 & 94.85
  & 45.47 & 61.39 & 75.62 & 67.42 & 80.26 & 78.38
  & 67.23 & 81.48 & 75.96 & 89.33 & 71.43 & 81.19 & 84.29 & 85.00 & 40.96 & 81.09 & \textbf{43.78}
  & 64.59 & 85.84
  & 70.70 \\
Qwen3.6-35B-A3B
  & 97.12 & 95.92 & 93.84 & 96.92 & \fbox{\textbf{96.48}} & 96.94
  & 46.05 & 64.30 & 81.82 & 70.86 & \textbf{85.53} & 75.68
  & 82.66 & 89.75 & 81.70 & 91.39 & 86.66 & 90.52 & 92.86 & 91.30 & 49.20 & 89.83 & 43.24
  & 80.54 & 88.02
  & 77.53 \\
LLaMA-4-Scout-17B-16E
  & 85.15 & 85.82 & 86.50 & 78.54 & 89.19 & 88.08
  & 47.26 & 61.55 & 76.03 & 65.70 & 77.63 & 75.68
  & 70.04 & 76.07 & 72.73 & 87.79 & 64.96 & 79.56 & 78.10 & 81.74 & 46.68 & 74.70 & 36.22
  & 67.32 & 78.87
  & 69.16 \\
LLaMA-4-Maverick-17B-128E
  & 93.16 & 91.19 & 92.44 & 90.75 & 91.86 & 94.69
  & 53.27 & 70.60 & 82.49 & 72.19 & 83.33 & 76.35
  & 82.03 & 84.79 & 77.87 & 89.72 & 82.48 & 88.59 & 85.56 & 88.70 & 54.69 & 82.27 & 36.76
  & 76.65 & 83.22
  & 80.98 \\
\midrule
\rowcolor{softgray}
\multicolumn{27}{c}{\textbf{Open-weight Models: $\leq$ 25B Parameters}} \\
\midrule
BLOOMZ-1.1B
  & 26.18 & 24.60 & 19.98 & 25.34 & 26.61 & 24.80
  & 25.94 & 25.61 & 25.72 & 26.62 & 28.95 & 21.62
  & 26.14 & 24.48 & 23.68 & 25.71 & 26.68 & 24.59 & 23.97 & 25.22 & 27.00 & 26.24 & 24.86
  & 22.37 & 24.84
  & 24.52 \\
BLOOMZ-1.7B
  & 17.41 & 15.04 & 19.55 & 16.67 & 20.41 & 12.72
  & 20.67 & 21.50 & 18.45 & 22.25 & 28.51 & 16.89
  & 27.68 & 30.10 & 35.77 & 36.50 & 23.05 & 27.56 & 24.92 & 24.35 & 26.32 & 26.95 & 36.76
  & 24.51 & 30.50
  & 25.92 \\
BLOOMZ-3B
  & 11.00 & 5.69 & 6.48 & 8.90 & 12.52 & 16.26
  & 3.67 & 2.59 & 4.04 & 3.44 & 7.89 & 7.43
  & 12.35 & 12.57 & 5.98 & 6.81 & 14.56 & 9.33 & 16.98 & 8.70 & 17.62 & 9.22 & 4.86
  & 5.25 & 7.84
  & 7.56 \\
BLOOMZ-7B
  & 22.33 & 23.20 & 23.97 & 22.15 & 20.41 & 24.80
  & 15.22 & 19.90 & 23.03 & 15.10 & 26.32 & 28.38
  & 18.24 & 23.48 & 25.60 & 25.96 & 26.55 & 22.22 & 24.60 & 21.96 & 21.28 & 25.53 & 17.30
  & 14.98 & 25.71
  & 22.54 \\
Gemma-2-9B-IT
  & 66.35 & 66.17 & 73.33 & 64.38 & 74.97 & 69.57
  & \textbf{41.92} & \textbf{52.33} & 57.04 & \textbf{56.56} & 65.79 & \textbf{68.92}
  & 52.55 & 63.73 & 64.35 & 80.08 & 52.83 & 69.04 & 68.25 & 64.13 & \textbf{42.33} & 62.65 & 32.43
  & 50.78 & \textbf{75.16}
  & 55.95 \\
Gemma-3-4B-IT
  & 46.79 & 49.52 & 58.32 & 44.86 & 62.21 & 48.95
  & 33.38 & 41.08 & 46.80 & 44.90 & 50.00 & 54.05
  & 43.05 & 48.07 & 51.91 & 67.74 & 44.47 & 56.74 & 53.81 & 47.61 & 29.75 & 51.06 & 29.19
  & 43.39 & 57.95
  & 46.33 \\
LLaMA-3.2-3B
  & 32.80 & 35.98 & 39.52 & 37.56 & 41.31 & 36.55
  & 24.75 & 33.83 & 34.41 & 36.29 & 40.79 & 35.14
  & 37.06 & 36.49 & 45.69 & 53.34 & 35.71 & 49.63 & 42.54 & 34.57 & 23.11 & 39.24 & 34.59
  & 33.27 & 44.23
  & 38.11 \\
LLaMA-3.1-8B
  & 42.20 & 45.97 & 51.51 & 43.38 & 51.76 & 43.64
  & 32.09 & 42.60 & 44.11 & 44.90 & 57.02 & 46.62
  & 47.24 & 49.28 & 58.25 & 68.89 & 42.45 & 56.30 & 53.33 & 44.35 & 27.00 & 47.75 & 30.27
  & 41.05 & 59.26
  & 45.08 \\
Ministral-3-3B
  & 52.35 & 50.38 & 62.85 & 55.02 & 64.64 & 59.90
  & 39.84 & 43.68 & 50.10 & 48.61 & 54.82 & 51.35
  & 43.27 & 53.36 & 59.69 & 71.98 & 45.82 & 56.00 & 58.57 & 53.91 & 37.99 & 53.90 & 32.97
  & 44.94 & 60.57
  & 48.97 \\
Ministral-3-8B
  & 66.67 & \textbf{69.60} & 73.43 & 66.21 & \textbf{79.10} & 75.04
  & 39.41 & 47.87 & \textbf{58.79} & 53.77 & \textbf{71.93} & 56.76
  & \textbf{54.88} & \textbf{66.48} & \textbf{66.15} & \textbf{80.33} & \textbf{58.09} & \textbf{69.63} & \textbf{70.48} & \textbf{67.17} & 38.67 & \textbf{65.72} & \textbf{44.32}
  & \textbf{54.28} & 69.93
  & \textbf{57.12} \\
Phi-4-mini
  & 36.54 & 32.44 & 32.72 & 45.32 & 41.92 & 33.33
  & 23.84 & 35.90 & 32.73 & 32.58 & 39.04 & 28.38
  & 34.04 & 40.02 & 45.57 & 55.01 & 35.71 & 45.33 & 39.21 & 34.13 & 21.28 & 40.43 & 28.11
  & 35.21 & 42.70
  & 35.90 \\
Phi-3.5-mini
  & 33.01 & 28.79 & 30.02 & 41.67 & 34.39 & 34.14
  & 25.29 & 30.91 & 28.08 & 26.75 & 32.46 & 26.35
  & 29.16 & 31.75 & 36.48 & 37.79 & 31.40 & 32.30 & 31.27 & 33.04 & 21.74 & 37.83 & 16.76
  & 30.54 & 38.34
  & 30.62 \\
Qwen3-4B-Instruct-2507
  & 66.45 & 56.82 & 69.55 & 71.69 & 76.31 & 74.72
  & 39.00 & 44.87 & 49.90 & 51.39 & 46.05 & 60.81
  & 43.64 & 57.22 & 58.49 & 72.11 & 50.54 & 60.74 & 56.83 & 56.30 & 37.07 & 55.32 & 30.81
  & 45.14 & 62.53
  & 47.80 \\
Qwen3-8B
  & \textbf{70.83} & 64.02 & \textbf{75.16} & \textbf{77.74} & 78.98 & \textbf{80.52}
  & 17.36 & 43.12 & 49.43 & 44.90 & 64.04 & 41.89
  & 46.50 & 58.43 & 65.19 & 74.94 & 54.18 & 63.41 & 61.43 & 62.61 & 28.38 & 61.47 & 34.59
  & 42.61 & 72.77
  & 49.49 \\
\midrule
\rowcolor{softgray}
\multicolumn{27}{c}{\textbf{Proprietary Models}} \\
\midrule
Claude-Haiku-4.5
  & 91.77 & 92.05 & 89.42 & 92.81 & 92.95 & 93.08
  & 47.79 & 66.49 & 82.76 & 71.26 & 77.19 & 78.38
  & 73.22 & 81.15 & 76.20 & 87.15 & 76.82 & 83.70 & 84.76 & 87.83 & 47.83 & 82.27 & 38.38
  & 70.82 & 86.93
  & 74.45 \\
Claude-Sonnet-4.6
  & 96.69 & 96.56 & 94.38 & 97.26 & 95.50 & 97.58
  & 62.13 & 81.13 & 91.65 & 85.17 & 87.72 & 84.46
  & 88.49 & 91.29 & 84.33 & \fbox{\textbf{93.70}} & 85.04 & 91.70 & 92.70 & 93.91 & 66.82 & 91.49 & \fbox{\textbf{52.97}}
  & 84.63 & 90.41
  & 86.05 \\
Gemini-3.1-Flash-Lite
  & 97.65 & 98.07 & 96.76 & 96.35 & \textbf{96.23} & 97.42
  & 64.00 & 83.73 & 91.72 & 85.43 & \fbox{\textbf{90.35}} & 85.81
  & 92.52 & 94.16 & 87.92 & 93.44 & 91.24 & 94.81 & 95.08 & 94.13 & 68.42 & 94.80 & 37.30
  & 87.55 & \fbox{\textbf{93.90}}
  & 85.76 \\
Gemini-3.5-Flash
  & \fbox{\textbf{97.76}} & \fbox{\textbf{98.60}} & \fbox{\textbf{97.08}} & 98.97 & 95.87 & 98.23
  & \fbox{\textbf{80.76}} & \fbox{\textbf{89.35}} & \fbox{\textbf{94.95}} & \fbox{\textbf{88.87}} & 89.04 & 89.19
  & \fbox{\textbf{95.02}} & \fbox{\textbf{94.82}} & \fbox{\textbf{88.76}} & \fbox{\textbf{93.70}} & \fbox{\textbf{93.67}} & \fbox{\textbf{94.96}} & \fbox{\textbf{95.71}} & \fbox{\textbf{95.65}} & \fbox{\textbf{77.80}} & \fbox{\textbf{95.27}} & 52.43
  & \fbox{\textbf{91.44}} & 91.72
  & \fbox{\textbf{91.85}} \\
GPT-5.4-mini
  & 89.64 & 91.19 & 91.04 & 78.20 & 90.16 & 89.21
  & 51.03 & 71.24 & 83.30 & 72.45 & 82.02 & 79.73
  & 77.84 & 82.14 & 77.15 & 89.72 & 69.81 & 83.11 & 81.75 & 89.57 & 59.73 & 85.11 & 36.76
  & 75.29 & 84.75
  & 75.26 \\
GPT-5.4
  & 97.44 & 97.85 & 95.79 & \fbox{\textbf{99.20}} & 95.87 & \textbf{98.55}
  & 65.86 & 79.46 & 91.45 & 81.06 & 89.47 & \fbox{\textbf{89.86}}
  & 90.67 & 93.94 & 86.96 & 92.42 & 92.32 & 94.22 & 94.29 & 93.48 & 69.34 & 92.43 & 40.00
  & 84.05 & 90.85
  & 83.92 \\
\midrule
\rowcolor{softgray}
\multicolumn{27}{c}{\textbf{Urdu Models}} \\
\midrule
Qalb-1.0-8B
  & \textbf{30.77} & \textbf{33.08} & 27.65 & \textbf{30.59} & 32.69 & \textbf{28.18}
  & \textbf{31.22} & 29.00 & \textbf{35.02} & \textbf{30.33} & \textbf{37.28} & \textbf{38.51}
  & \textbf{35.37} & \textbf{30.65} & \textbf{27.63} & 32.78 & \textbf{29.11} & \textbf{33.93} & \textbf{32.22} & 26.96 & \textbf{29.98} & \textbf{28.37} & 12.43
  & \textbf{35.60} & \textbf{35.29}
  & \textbf{36.05} \\
Alif-1.0-8B
  & 28.21 & 27.07 & \textbf{30.99} & 27.17 & \textbf{33.54} & 27.05
  & 27.17 & \textbf{29.24} & 28.96 & 28.08 & 25.44 & 32.43
  & 31.39 & 27.45 & 24.40 & \textbf{34.70} & 24.12 & 32.30 & 29.21 & \textbf{28.70} & 26.54 & 27.90 & \textbf{14.59}
  & 33.66 & 27.89
  & 29.66 \\
\bottomrule
\end{tabular}%
}
\caption{\textbf{Subdomain-level model performance on \ds{} under the Urdu prompt.} All-item accuracy (\%) across all 26 subdomains grouped by domain, with unparsable outputs scored as incorrect. Subdomains are ordered within each domain by dataset size (descending); see Table~\ref{tab:urdummlu-domains-levels} for acronym expansions. Boxed values mark the best score per column, while bold values indicate the best score within each model group.}
\label{tab:subdomain_performance_ur}
\end{table*}

\subsubsection{Urdu-Tuned Models on Urdu Literature}
\label{app:urdu-literature-deep-dive}

\begin{figure}[!h]
    \centering
    \includegraphics[width=\linewidth]{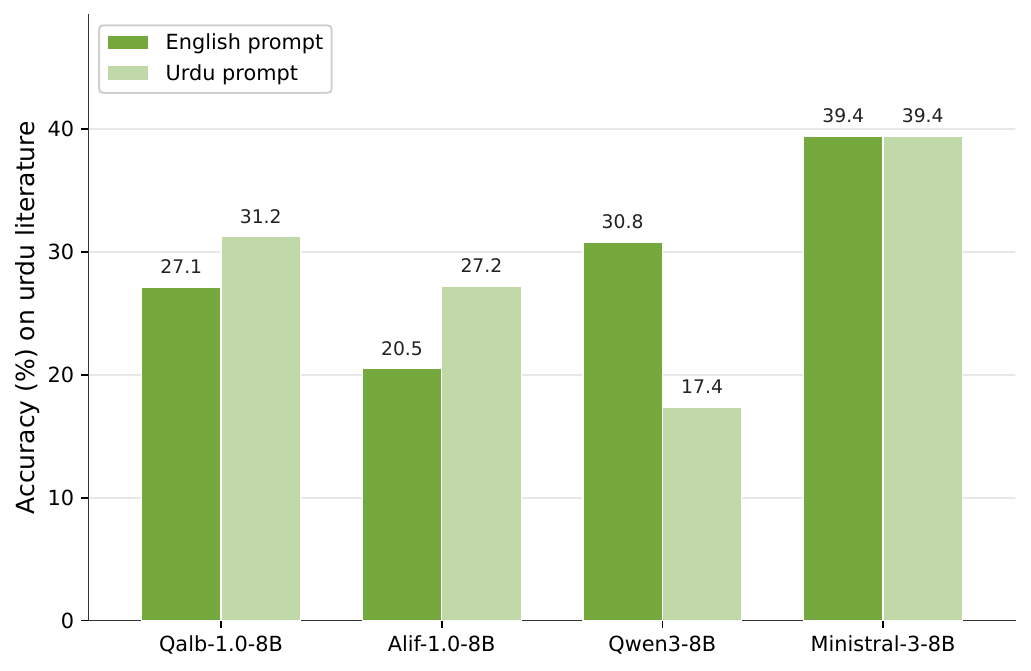}
    \caption{\textbf{Urdu literature accuracy across 8B-class models.} Ministral-3-8B leads under both English and Urdu prompts, while Qwen3-8B shows the largest decline.}
    \label{fig:urdu-literature-8b}
\end{figure}

Urdu literature is the largest subdomain in \ds{}, with 5{,}859 items, and contains content with limited overlap with English-dominated pretraining corpora, including classical poetry, prosody, and literary history. It therefore provides a useful test of Urdu-focused training on culturally grounded knowledge. Figure~\ref{fig:urdu-literature-8b} compares two Urdu-targeted 8B models, Qalb-1.0-8B and Alif-1.0-8B, with two general-purpose 8B instruction-tuned models, Qwen3-8B and Ministral-3-8B to test whether Urdu-focused training provides an advantage at a matched scale.

The Urdu-targeted models do not outperform the general-purpose baselines on this subdomain. Ministral-3-8B achieves the highest accuracy under both prompts at $39.4\%$, while Qalb-1.0-8B and Alif-1.0-8B remain below $32\%$. Qwen3-8B performs competitively under the English prompt ($30.8\%$) but drops to $17.4\%$ under the Urdu prompt. In contrast, both Urdu-targeted models achieve higher accuracy under the Urdu prompt. However, these prompt-level differences do not establish whether the gains arise from Urdu-specific tuning, instruction following, or other model-specific factors. Overall, Urdu literature remains challenging even for Urdu-targeted LLMs.

\subsubsection{English-Prompt Subdomain Accuracy}
\label{app:subdomain-en}

Table~\ref{tab:subdomain_performance_en} reports per-subdomain accuracy for all 30 models under the English prompt. The table groups subdomains by domain and orders them by dataset size within each group, so earlier columns contribute more strongly to the corresponding domain-level scores in Table~\ref{tab:modelperformance}. Acronym expansions appear directly in the table header. Results show model behavior across subjects: Gemini-3.5-Flash remains consistently strong across nearly all subdomains, DeepSeek-V4-Flash approaches proprietary performance on STEM but drops on Urdu language and literature, and BLOOMZ models perform near or below the random baseline across most subjects.

\subsubsection{Urdu-Prompt Subdomain Accuracy}
\label{app:subdomain-ur}

Table~\ref{tab:subdomain_performance_ur} reports the per-subdomain breakdown under the Urdu prompt. It follows the structure and ordering of Table~\ref{tab:subdomain_performance_en}, allowing direct comparison between prompt settings. Most differences remain small, reinforcing the main finding from Section~\ref{sec:results} that the difficulty of \ds{} comes primarily from the question content rather than the instruction language. For most proprietary models and the Gemma family, English- and Urdu-prompt accuracies remain nearly identical across the majority of subdomains. Model-specific shifts are clearer by subdomain. Qwen3.6-35B-A3B records higher Urdu-setting scores across domains, led by Social Sciences ($+10.59$ points), Profession ($+8.43$), Other ($+7.56$), and STEM ($+7.38$). However, reasoning was enabled only under the Urdu prompt, so these differences cannot be attributed to prompt language. Qwen3-8B declines mainly in Humanities, especially Urdu language and literature, driving its overall Humanities drop.

The Urdu-targeted models also show modest gains on several Humanities subdomains under the Urdu prompt, although these improvements do not substantially change their overall ranking. These patterns further support the conclusion that prompt language plays a secondary role compared with the educational and cultural knowledge required by the benchmark. Urdu instructions may help models access knowledge acquired during training, but they cannot compensate for knowledge that was not learned.

\section{Invalid-Output Examples}
\label{app:invalid-outputs}

Section~\ref{sec:output-validity} reports invalid-output rates across the model suite. This appendix presents representative failure cases drawn from actual predictions under the Urdu prompt. We group them by failure type to show recurring decoding behaviors and how invalid generations appear across models. Each response lacks an extractable answer key and is scored as incorrect without being removed from the denominator, illustrating why strict parsing is necessary.

\paragraph{Repetition collapse:} In some cases, the model enters a degenerate decoding loop and repeatedly emits the same token sequence without producing a meaningful or valid answer. Example~\ref{ex:invalid-repetition} illustrates this behavior for BLOOMZ-7B, which repeatedly generates the token ``Question:'' dozens of times instead of producing a task-relevant response.

\refstepcounter{invex}\label{ex:invalid-repetition}
\begin{tcolorbox}[title={Example~\theinvex: repetition collapse (BLOOMZ-7B, Urdu literature, gold = B)},
                  width=\linewidth, label={ex:invalid-repetition}, breakable]
\small
\textbf{Question:} \foreignlanguage{urdu}{خواجہ دل محمد کس مضمون کے استاد تھے}\\
\textbf{Options:}
\begin{itemize}[noitemsep, topsep=2pt, leftmargin=*]
  \item A. \foreignlanguage{urdu}{ان میں سے کوئی نہیں}
  \item B. \foreignlanguage{urdu}{ریاضی} \quad\textit{(gold)}
  \item C. \foreignlanguage{urdu}{اردو}
  \item D. \foreignlanguage{urdu}{پنجابی}
\end{itemize}
\textbf{Model output:}\\
\texttt{Question: Question: Question: Question: \dots}
\end{tcolorbox}

\paragraph{Prompt echo:} The model copies part of the user prompt instead of answering the question. Example~\ref{ex:invalid-prompt-echo} shows BLOOMZ-1.7B reproducing the beginning of the question prompt and terminating before generating a valid answer.

\refstepcounter{invex}\label{ex:invalid-prompt-echo}
\begin{tcolorbox}[title={Example~\theinvex: prompt echo (BLOOMZ-1.7B, Islamic studies, gold = B)},
                  width=\linewidth, label={ex:invalid-prompt-echo}, breakable]
\small
\textbf{Question:} \foreignlanguage{urdu}{مومنو! اگر تم میری راہ میں لڑنے اور میری خوشنودی طلب کرنے کے لیے نکلے ہو تو میرے اور اپنے دشمنوں کو \rule{1cm}{0.4pt} مت بناؤ۔}\\
\textbf{Options:}
\begin{itemize}[noitemsep, topsep=2pt, leftmargin=*]
  \item A. \foreignlanguage{urdu}{بھائی}
  \item B. \foreignlanguage{urdu}{دوست} \quad\textit{(gold)}
  \item C. \foreignlanguage{urdu}{دشمن}
  \item D. \foreignlanguage{urdu}{راز دار}
\end{itemize}
\textbf{Model output:}\\
\foreignlanguage{urdu}{مضمون:} Humanities -- islamic studies | \foreignlanguage{urdu}{سطح:} SSC-II\\
\foreignlanguage{urdu}{سوال: مومنو! اگر تم میری راہ میں لڑنے اور میری خوشنودی طلب کر}
\end{tcolorbox}

\paragraph{Refusal or clarification request:} Instead of selecting an answer, the model returns a conversational clarification request. Example~\ref{ex:invalid-refusal} shows Qalb-1.0-8B treating the MCQ as an ambiguous user query. Although four answer options are provided, the model attempts to continue the dialogue rather than producing the required answer key.

\refstepcounter{invex}\label{ex:invalid-refusal}
\begin{tcolorbox}[title={Example~\theinvex: refusal (Qalb-1.0-8B, mathematics, gold = A)},
                  width=\linewidth, label={ex:invalid-refusal}, breakable]
\small
\textbf{Question:} \foreignlanguage{urdu}{قطاروں اور کالموں کی تعداد کسی قالب میں \rule{1cm}{0.4pt} کو ظاہر کرتی ہے۔}\\
\textbf{Options:}
\begin{itemize}[noitemsep, topsep=2pt, leftmargin=*]
  \item A. \foreignlanguage{urdu}{مرتبہ} \quad\textit{(gold)}
  \item B. \foreignlanguage{urdu}{مقطع}
  \item C. \foreignlanguage{urdu}{قطاریں}
  \item D. \foreignlanguage{urdu}{کالم}
\end{itemize}
\textbf{Model output:}\\
\foreignlanguage{urdu}{آپ کا سوال واضح نہیں ہے۔ کیا آپ اسے دوبارہ بیان کر سکتے ہیں؟}
\end{tcolorbox}

\paragraph{System-prompt echo:} The model reproduces the system prompt instead of answering the question. Example~\ref{ex:invalid-system-echo} shows Alif-1.0-8B repeating the assistant role description without generating an answer. The resulting output contains no answer key that can be scored.

\refstepcounter{invex}\label{ex:invalid-system-echo}
\begin{tcolorbox}[title={Example~\theinvex: system-prompt echo (Alif-1.0-8B, chemistry, gold = B)},
                  width=\linewidth, label={ex:invalid-system-echo}, breakable]
\small
\textbf{Question:} \foreignlanguage{urdu}{الکحلز کی ڈی ہائیڈریشن \rule{1cm}{0.4pt} کے ساتھ کی جا سکتی ہے۔}\\
\textbf{Options:}
\begin{itemize}[noitemsep, topsep=2pt, leftmargin=*]
  \item A. NaOH
  \item B. H$_{2}$SO$_{4}$ \quad\textit{(gold)}
  \item C. HCl
  \item D. KOH
\end{itemize}
\textbf{Model output:}\\
\foreignlanguage{urdu}{جوابدہ معاون ہیں۔} MCQ \foreignlanguage{urdu}{آپ ایک ماہر} 
\end{tcolorbox}

\paragraph{Empty or placeholder output:} The model emits a nearly empty response, often copied directly from a blank marker in the question. Example~\ref{ex:invalid-empty} shows BLOOMZ-3B returning only a placeholder token. Because it contains neither an answer key nor option text, the response cannot be mapped to any candidate answer.

\refstepcounter{invex}\label{ex:invalid-empty}
\begin{tcolorbox}[title={Example~\theinvex: empty output (BLOOMZ-3B, chemistry, gold = A)},
                  width=\linewidth, label={ex:invalid-empty}, breakable]
\small
\textbf{Question:} \foreignlanguage{urdu}{\rule{1cm}{0.4pt} فریکشن بطور جیٹ فیول استعمال ہوتی ہے۔}\\
\textbf{Options:}
\begin{itemize}[noitemsep, topsep=2pt, leftmargin=*]
  \item A. \foreignlanguage{urdu}{کیروسین آئل} \quad\textit{(gold)}
  \item B. \foreignlanguage{urdu}{لبریکیٹنگ آئل}
  \item C. \foreignlanguage{urdu}{فیول آئل}
  \item D. \foreignlanguage{urdu}{ڈیزل آئل}
\end{itemize}
\textbf{Model output:}\\
\texttt{\_\_\_}
\end{tcolorbox}

\paragraph{Discussion:} All five examples fail at the response-interface level because none contains an extractable answer key from \texttt{A} through \texttt{D}. They therefore cannot be treated as ordinary answer-selection errors. Under all-item accuracy, however, each response remains in the denominator and is scored as incorrect. This distinction matters because accuracy alone does not show whether a model selected the wrong option or failed to produce a scorable answer. The failure types also suggest different behaviors. Repetition and placeholder outputs are consistent with decoding instability, prompt echoes show failure to separate instructions from question content, and clarification requests reflect a conversational response that does not follow the MCQ task. These failures occur most often among weaker models under the Urdu prompt. Reporting accuracy only over parseable responses would exclude such cases and overstate practical performance. We therefore report all-item accuracy together with invalid-output rates to distinguish subject-level errors from failures to follow the required response format.

\end{document}